\documentclass[10pt,twocolumn,letterpaper]{article}

\usepackage[pagenumbers]{cvpr}      
\usepackage{times}
\usepackage{epsfig}
\usepackage{graphicx}
\usepackage{amsmath}
\usepackage{amssymb}
\usepackage{color, colortbl}
\usepackage[dvipsnames]{xcolor}
\usepackage{pifont}
\usepackage{booktabs}
\usepackage{multirow}
\usepackage[accsupp]{axessibility}

\usepackage[utf8]{inputenc}

\newcommand\blfootnote[1]{%
  \begingroup
  \renewcommand\thefootnote{}\footnote{#1}%
  \addtocounter{footnote}{-1}%
  \endgroup
}

\definecolor{cvprblue}{rgb}{0.21,0.49,0.74}
\usepackage[breaklinks,colorlinks,citecolor=cvprblue]{hyperref}

\definecolor{gold}{rgb}{1.0, 0.874, 0}
\definecolor{silver}{rgb}{0.77,0.77,0.77}
\definecolor{brown}{rgb}{0.95, 0.678, 0.4}

\newcommand{\gold}[1]{\colorbox{gold}{\textbf{#1}}}
\newcommand{\silver}[1]{\colorbox{silver}{\textbf{#1}}}
\newcommand{\bronze}[1]{\colorbox{brown}{\textbf{#1}}}

\begin{document}
\title{NTIRE 2025 Challenge on \\ HR Depth from Images of Specular and Transparent Surfaces}

\author{Pierluigi Zama Ramirez  \and  Fabio Tosi \and  Luigi Di Stefano \and  Radu Timofte \and 
Alex Costanzino \and  Matteo Poggi \and  Samuele Salti \and  Stefano Mattoccia  \and
Zhe Zhang \and Yang Yang \and
Wu Chen \and Anlong Ming \and Mingshuai Zhao \and Mengying Yu \and
Shida Gao \and Xiangfeng Wang \and Feng Xue \and
Jun Shi \and  Yong Yang \and Yong A \and  Yixiang Jin \and  Dingzhe Li \and
Aryan Shukla \and Liam Frija-Altarac \and Matthew Toews \and
Hui Geng \and Tianjiao Wan \and Zijian Gao \and Qisheng Xu \and Kele Xu \and Zijian Zang \and
Jameer Babu Pinjari \and Kuldeep Purohit \and
Mykola Lavreniuk \and
Jing Cao \and Shenyi Li \and Kui Jiang \and Junjun Jiang \and Yong Huang
}

\maketitle

\begin{abstract}
This\blfootnote{$^*$Pierluigi Zama Ramirez (pierluigi.zama@unibo.it), Alex Costanzino, Fabio Tosi, Matteo Poggi, Samuele Salti, Stefano Mattoccia, Luigi Di Stefano and Radu Timofte are the NTIRE 2025 HR Depth from Images of Specular and Transparent Surfaces challenge organizers. The other authors participated in the challenge.\\ Sections \ref{sec:teams_stereo} and \ref{sec:teams_mono} contains the authors’ team names and affiliations.\\ The NTIRE
website: \url{https://cvlai.net/ntire/2025/}} paper reports on the NTIRE 2025 challenge on HR Depth From images of Specular and Transparent surfaces, held in conjunction with the New Trends in Image Restoration and Enhancement (NTIRE) workshop at CVPR 2025.
This challenge aims to advance the research on depth estimation, specifically to address two of the main open issues in the field: high-resolution and non-Lambertian surfaces. 
The challenge proposes two tracks on stereo and single-image depth estimation, attracting about 177 registered participants. 
In the final testing stage, 4 and 4 participating teams submitted their models and fact sheets for the two tracks. 
\end{abstract}

\section{Introduction}

Reversing the image formation process to model the 3D structure of the world represents one of the quintessential tasks studied by computer vision. 
For this purpose, estimating depth from images often lays the foundation of this process, as well as the entry point to higher-level applications such as augmented reality, autonomous or assisted driving, robotics, and more.
Recovering this information from images represents a cheaper and more viable alternative to the use of \textit{active} depth sensors -- such as Radars, LiDARs, Time-of-Flight (ToF), and others -- which are known for their higher cost and multiple limitations preventing their unconstrained deployment in any environment. Furthermore, the disruptive advent of deep learning in computer vision has made the former strategy more and more preferable over active sensors, also thanks to the recent development of the first \textit{foundational} models for depth estimation and, in general, 3D vision.
Although this brought a rapid evolution of the depth estimation models observed in the last decade, this task remains far from being solved in the presence of some particularly challenging conditions. 
Among the many, we argue that two matters of interest are common to the different approaches devoted to estimating depth from images -- and even to active sensors.

The first is a longstanding challenge, common to any computer vision task: spatial resolution. Indeed, although nowadays we have color cameras capable of capturing frames up to dozens of Megapixels (Mpx) whereas active sensors fall far behind, processing high-resolution images poses several challenges, in terms of computational requirements as well as data and training methodologies to deploy deep models capable of exploiting such rich information.

The second consists of the ambiguity that may occur in some images: this may assume different forms, such as the lack of texture or the absence of perspective cues, and affect different kinds of depth estimation strategies. Specifically, the presence of \textit{non-Lambertian} surfaces represents a challenge to any of these approaches, as well as to active sensors -- since materials featuring this property often violate the basic assumptions upon which active sensors are built, e.g., with LiDARs beams being refracted or surpassing transparent surfaces. 
This makes ground-truth depth annotations, often sourced through active sensors, very hard to collect and, therefore, very rare in the training data leveraged by most state-of-the-art image-based depth estimation models, making these latter  failing to estimate the distance of a transparent surface in favor of the distance of objects behind it, or the surface of a mirror in place of the depth of the reflected objects.
Although these latter examples might not represent real failure cases, since the definition of depth itself becomes ambiguous in such circumstances, we argue it is for some popular applications, for instance when properly perceiving the real depth for transparent objects may be crucial to accomplishing a higher level task, like grasping some glassy objects or navigating in an indoor environment where glass doors may be common.

This NTIRE 2025 Challenge on HR Depth from Images of Specular and Transparent Surfaces aims at pushing the development of 
state-of-the-art methodologies answering to the aforementioned challenges. Following the previous, successful editions \cite{ramirez2023ntire,ramirez2024ntire}, we build our challenge over the Booster dataset \cite{booster,booster2}, a benchmark peculiarly encompassing both high-resolution and non-Lambertian surfaces, thanks to its 12Mpx images and the abundant presence of transparent and reflective objects. 
Following our tradition, the challenge is organized into two tracks: one devoted to \textit{Stereo} approaches, where depth is measured through triangulation from the \textit{disparity} estimated between pixels into two rectified frames, and the other focusing on single-image frameworks (\textit{Mono}).
The challenge attracted up to 177 registered participants. Among them, 5 and 4 teams, respectively, for the monocular and stereo tracks, submitted their models and fact sheets during the final phase. Some adopt the most recent foundational models in the field as off-the-shelf solutions, whereas others develop their own custom frameworks. The outcome of this edition of our challenge is reported and discussed in detail in Section~\ref{sec:results}.

\section{Related Work}

\textbf{Deep Stereo Matching.} Ten years ago already, the community started facing stereo depth estimation with deep neural networks \cite{zbontar2016stereo}, becoming the standard approach to this task over the years \cite{poggi2021synergies,poggi2021confidence}. 
At first, two main families of end-to-end models were developed, respectively, 2D \cite{mayer2016large,Pang_2017_ICCV_Workshops, Liang_2018_CVPR, tonioni2019real, saikia2019autodispnet, song2018edgestereo, yang2018segstereo, yin2019hierarchical, Tankovich_2021_CVPR, Poggi_2024_CVPR_FedStereo} and 3D \cite{Kendall_2017_ICCV, chang2018psmnet, khamis2018stereonet, zhang2019ga, cheng2019learning, cheng2020hierarchical, duggal2019deeppruner, yang2019hierarchical, wang2019anytime, guo2019group,Tosi2021CVPR, Shen_2021_CVPR} architectures. 
In the 2020s \cite{tosi2025survey}, the advent of new paradigms to deal with dense matching tasks, such as the use of optimization-based frameworks \cite{teed2020raft} or transformers \cite{li2021revisiting, guo2022context, lou2023elfnet} ignited the development of two new lines of research.
The former in particular, starting with RAFT-Stereo \cite{teed2020raft}, rapidly become the most popular approach \cite{li2022practical,Jing_2023_ICCV_CREStereo++,xu2023iterative_IGEV-Stereo,Zeng_2023_ICCV,xu2023unifying_GMStereo,Tosi_2023_CVPR}.
This translated into a steady saturation of the most popular benchmarks, from KITTI 2012 \cite{geiger2010efficient} and 2015 \cite{Menze2015CVPR}, to ETH3D \cite{schops2017multi} and Middlebury 2014 \cite{scharstein2014high}. Other approaches try to refine disparity maps for high-resolution predictions\cite{aleotti2021neural,ndr}. 
Lately, the first foundational models for stereo depth estimation have appeared \cite{cheng2025monster,Bartolomei2025StereoAnywhere,jiang2025defom,wen2025stereo}, achieving a consistent step forward in terms of zero-shot generalization and robustness in handling non-Lambertian surfaces. 
Indeed, as we will notice in the remainder of this paper, some of these solutions were successfully deployed on the Booster dataset \cite{booster} as well.

\textbf{Monocular Depth Estimation.}  
Deep learning allowed facing highly ill-posed tasks, such as estimating depth out of a single image \cite{chen2016single, eigen2014depth,laina2016deeper, Ramamonjisoa_2020_CVPR, wang2020cliffnet}, thanks to the increasing availability of large-scale, annotated datasets \cite{chen2016single, eigen2014depth,laina2016deeper, Ramamonjisoa_2020_CVPR, wang2020cliffnet} or to the emergence of self-supervised paradigms  \cite{godard2017unsupervised,gonzales2020forgetlidar,godard2019monodepth2,Zhou_2017_ICCV,watson2019depthhints,guo2018learning,zama2019geometry,jiang2018self,johnston2020self,Tosi_2019_CVPR,Tosi_2020_CVPR,Poggi_2020_CVPR,zhao2023gasmono} replacing the need for explicit depth annotation with principles of multi-view geometry, for instance by casting the depth estimation process into an image reconstruction problem during training thanks to the availability of either paired stereo images or monocular videos.
A major trend emerging in the twenties consists of the development of affine-invariant depth estimation models \cite{Ranftl2022,Ranftl2021}, capable of generalizing beyond the single-dataset domain. MiDaS \cite{Ranftl2022} took this direction first, training a deep network on a mixture of multiple datasets to achieve cross-domain generalization, then followed by DPT \cite{Ranftl2021}, and others focusing on recovering the real shapes from the deformed point cloud obtained from monocular depth maps \cite{yin2021learning} or restoring high-frequency details \cite{miangoleh2021boosting,li2023patchfusion} at a higher resolution. Affine-invariant models recently converged into the first foundational models for single-image depth estimation, such as the Depth Anything series \cite{depthanything,yang2024depth}, or the newest diffusion-derived frameworks such as Marigold \cite{ke2023repurposing}, GeoWizard \cite{fu2024geowizard}, Lotus \cite{he2024lotus} and others, then extended to deal with video depth estimation \cite{shao2025chronodepth,hu2025DepthCrafter,ke2024rollingdepth}.

Lately, the ability of single-image depth estimation to effectively handle transparent and reflective surfaces has gained relevance, also thanks to the advent of benchmarks dedicated to this purpose \cite{booster2}. On this track, some approaches developed an annotation pipeline to obtain reliable pseud-labels for non-Lambertian objects, by using pre-trained monocular depth estimation models jointly with material segmentation masks \cite{costanzino2023iccv} or diffusion models \cite{tosi2024diffusion}, whereas others employed depth completion approaches \cite{choi2021selfdeco, sajjan2020clear} to fill the holes in the depth maps occurring in correspondence of transparent surfaces.
Furthermore, some of the latest foundational models such as Depth Anything v2 \cite{yang2024depth} expose surprising effectiveness at perceiving transparent or mirroring surfaces, as shown in the remainder.

\textbf{Competitions/Challenges on Depth Estimation.} The depth estimation task, both from stereo and monocular images, as been the object of several challenges taking place in the previous years, or even concurrently with ours. Among them, the Robust Vision Challenge (ROB) \cite{rob_challenge} covering both, the Dense Depth for Autonomous Driving challenge (DDAD)\cite{ddad_challenge}, the Fast and Accurate Single-Image Depth Estimation on Mobile Devices Challenge (MAI) \cite{mai_challenge}, the Argoverse Stereo Challenge \cite{argoverse_challenge} and the Monocular Depth Estimation Challenge (MDEC) \cite{Spencer_2023_WACV,Spencer2023MDEC2,Spencer2024MDEC,Obukhov2025MDEC}. 
Finally, we recall the previous editions of this challenge \cite{ramirez2023ntire,ramirez2024ntire,zamaramirez2024tricky}, part of the NTIRE workshop at CVPR 2023 and 2024 and the TRICKY workshop at ECCV 2024.

\textbf{NTIRE 2025 Challenges.}
This challenge is one of the NTIRE 2025~\footnote{\url{https://www.cvlai.net/ntire/2025/}} Workshop associated challenges on: ambient lighting normalization~\cite{ntire2025ambient}, reflection removal in the wild~\cite{ntire2025reflection}, shadow removal~\cite{ntire2025shadow}, event-based image deblurring~\cite{ntire2025event}, image denoising~\cite{ntire2025denoising}, XGC quality assessment~\cite{ntire2025xgc}, UGC video enhancement~\cite{ntire2025ugc}, night photography rendering~\cite{ntire2025night}, image super-resolution (x4)~\cite{ntire2025srx4}, real-world face restoration~\cite{ntire2025face}, efficient super-resolution~\cite{ntire2025esr}, HR depth estimation~\cite{ntire2025hrdepth}, efficient burst HDR and restoration~\cite{ntire2025ebhdr}, cross-domain few-shot object detection~\cite{ntire2025cross}, short-form UGC video quality assessment and enhancement~\cite{ntire2025shortugc,ntire2025shortugc_data}, text to image generation model quality assessment~\cite{ntire2025text}, day and night raindrop removal for dual-focused images~\cite{ntire2025day}, video quality assessment for video conferencing~\cite{ntire2025vqe}, low light image enhancement~\cite{ntire2025lowlight}, light field super-resolution~\cite{ntire2025lightfield}, restore any image model (RAIM) in the wild~\cite{ntire2025raim}, raw restoration and super-resolution~\cite{ntire2025raw} and raw reconstruction from RGB on smartphones~\cite{ntire2025rawrgb}.

\begin{table*}[t]
    \centering
    \resizebox{\linewidth}{!}{
    \begin{tabular}{l||c|rrrrrr||c|rrrrrr||rrrrrr}
    
     & \multicolumn{7}{c||}{\cellcolor{blue!25}ToM} & \multicolumn{7}{c||}{\cellcolor{pink}All} & \multicolumn{6}{c}{\cellcolor{YellowOrange}Other} \\
    \hline
    Team & Rank & {\textcolor{red}{\textbf{bad-2}}} & bad-4 & bad-6 & bad-8 & MAE & RMSE & Rank & {\textcolor{red}{\textbf{bad-2}}} & bad-4 & bad-6 & bad-8 & MAE & RMSE & bad-2 & bad-4 & bad-6 & bad-8 & MAE & RMSE \\
    
    \hline
    \textbf{SRC-B} &  \gold{\#1} &
    \gold{42.47} & \gold{25.39} & \gold{18.91} & \gold{14.21} & \gold{4.66} & \gold{7.55} &  \bronze{\#3} &
    \bronze{25.50} & \bronze{12.27} & \bronze{7.81} & \silver{5.42} & \gold{2.34} & \gold{5.46} &
    22.54 & \bronze{9.19} & \bronze{5.21} & \bronze{3.31} & \silver{1.81} & \silver{4.43} \\

    \textbf{Robot01-vRobotit} &  \silver{\#2} &
    \silver{45.20} & \bronze{28.30} & \bronze{20.47} & \bronze{16.37} & \silver{5.14} & \bronze{9.39} & \#4 &
    30.02 & 13.34 & 8.59 & 6.25 & \bronze{2.86} & \bronze{7.23} &
    28.75 & 11.22 & 6.82 & 4.67 & 2.54 & 6.61 \\

    \textbf{NJUST-KMG} &  \bronze{\#3} &
    \bronze{50.25} & \silver{27.97} & \silver{19.97} & \silver{16.28} & \bronze{5.79} & \silver{8.77} & \silver{\#2} &
    \silver{22.64} & \silver{9.76} & \gold{6.29} & \gold{4.77} & \silver{2.44} & \silver{5.68} &
    \bronze{18.75} & \silver{6.36} & \silver{3.71} & \silver{2.62} & \bronze{1.82} & \bronze{4.45} \\
    
    \textbf{weouibaguette} & \#4 &
    52.30 & 32.57 & 26.49 & 23.63 & 18.54 & 25.40 & \gold{\#1} &
    \gold{19.27} & \gold{8.74} & \silver{6.44} & \bronze{5.43} & 5.06 & 12.22 &
    \silver{13.64} & \gold{3.51} & \gold{1.76} & \gold{1.04} & \gold{1.27} & \gold{3.48} \\
    
    \hline
    
    \textbf{CREStereo [baseline]} & \#5 &
    59.64 & 47.26 & 40.27 & 35.41 & 24.69 & 42.28 & \#5 &
    35.75 & 23.51 & 18.98 & 16.42 & 12.13 & 28.46 &
    \gold{8.34} & 13.93 & 7.91 & 4.64 & 2.95 & 7.59 \\
    \hline
    
\end{tabular}
    }
    \vspace{-0.3cm}
    \caption{\textbf{Stereo Track: Evaluation on the Challenge Test Set.} Predictions evaluated at full resolution (4112$\times$3008) on All pixels and pixels belonging to ToM (Transparent or Mirror) or Other materials. 
    In \gold{gold}, \silver{silver}, and \bronze{bronze}, we show first, second, and third-rank approaches, respectively.
    We rank methods on two metrics, \textcolor{red}{\textbf{$\delta < {1.05}$}} computed on either ToM or All pixels.
    }\vspace{-0.3cm}
    \label{tab:stereo}
\end{table*}


\begin{figure*}[t]
\setlength{\tabcolsep}{1pt}
    \centering
    \begin{tabular}{ccccccc}
\scriptsize\textit{\textbf{RGB}} & \scriptsize\textbf{\textit{GT}} & \scriptsize\textbf{\textit{CREStereo}}\cite{bhat2023zoedepth} & \scriptsize\textbf{\textit{SRC-B}} & \scriptsize\textbf{\textit{Robot01-vRobotit}} & \scriptsize\textbf{\textit{NJUST-KMG}} & \scriptsize\textbf{\textit{weouibaguette}} \\    

\includegraphics[width=0.125\linewidth]{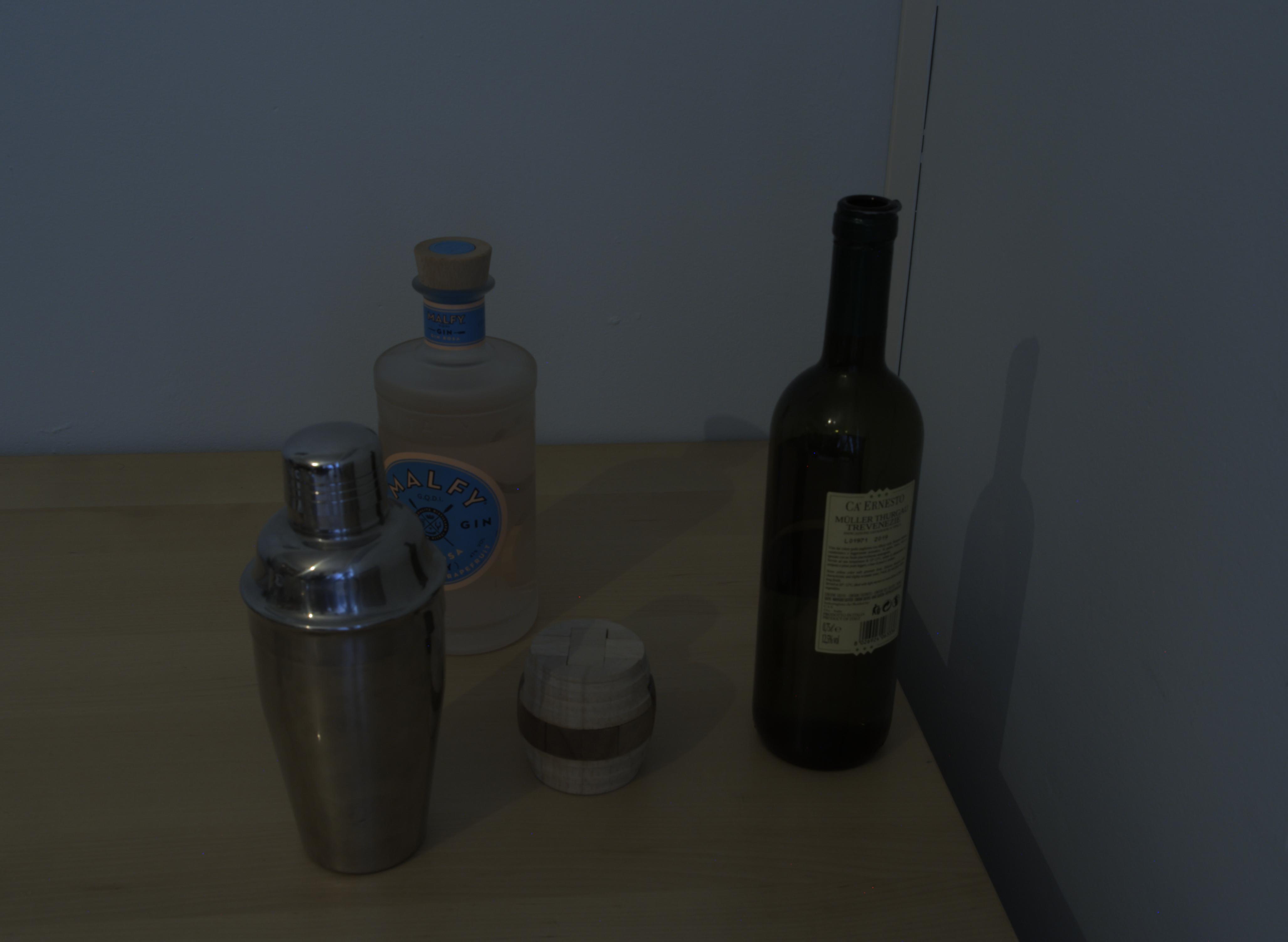} & \includegraphics[width=0.125\linewidth]{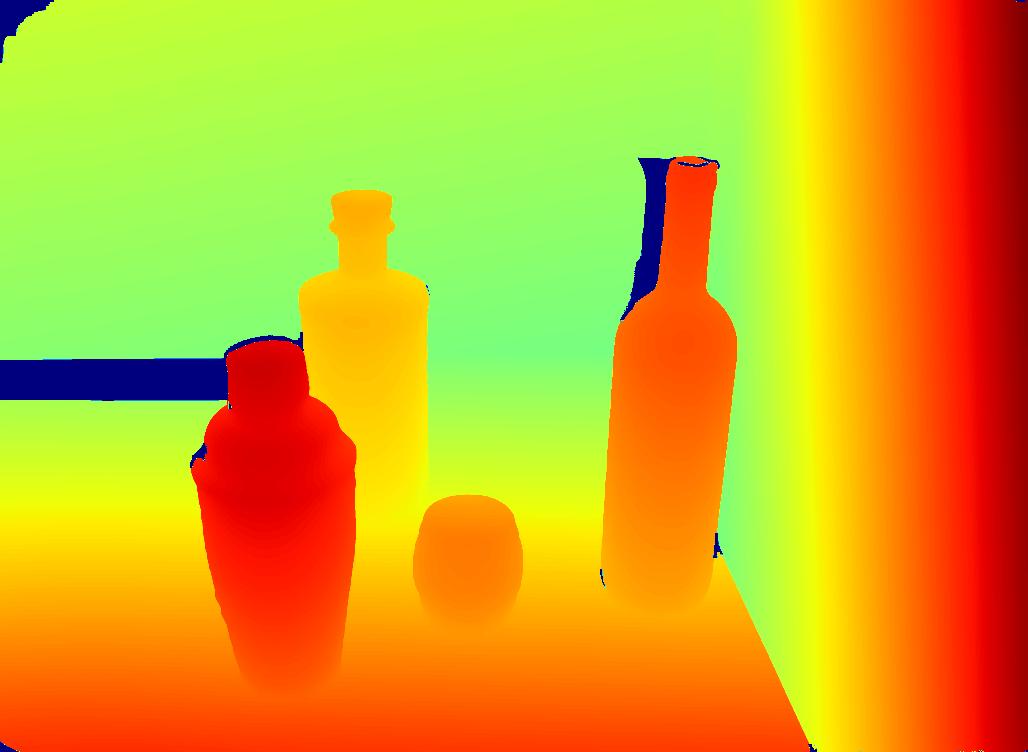 } & 
\includegraphics[width=0.125\linewidth]{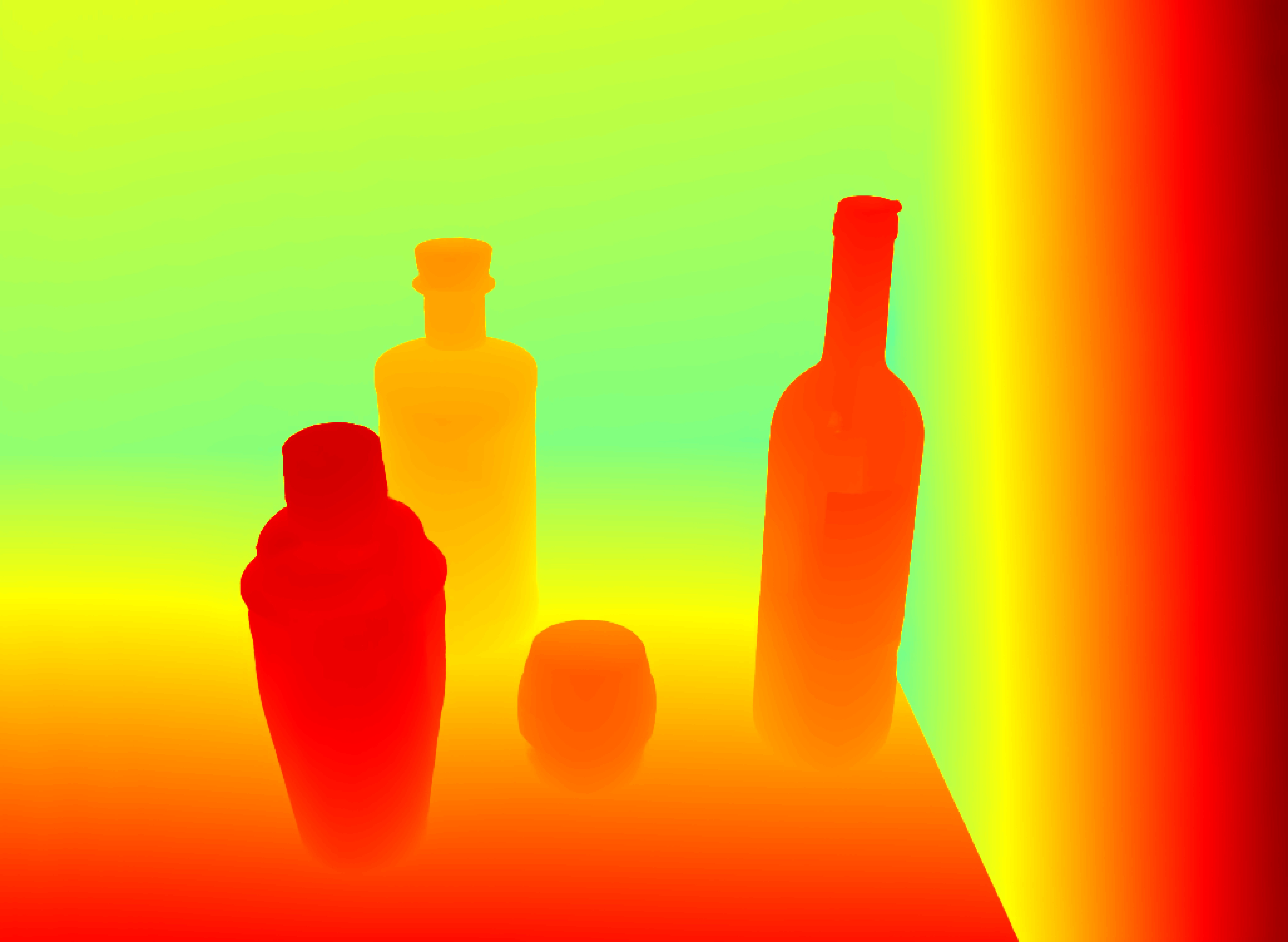 } & 
\includegraphics[width=0.125\linewidth]{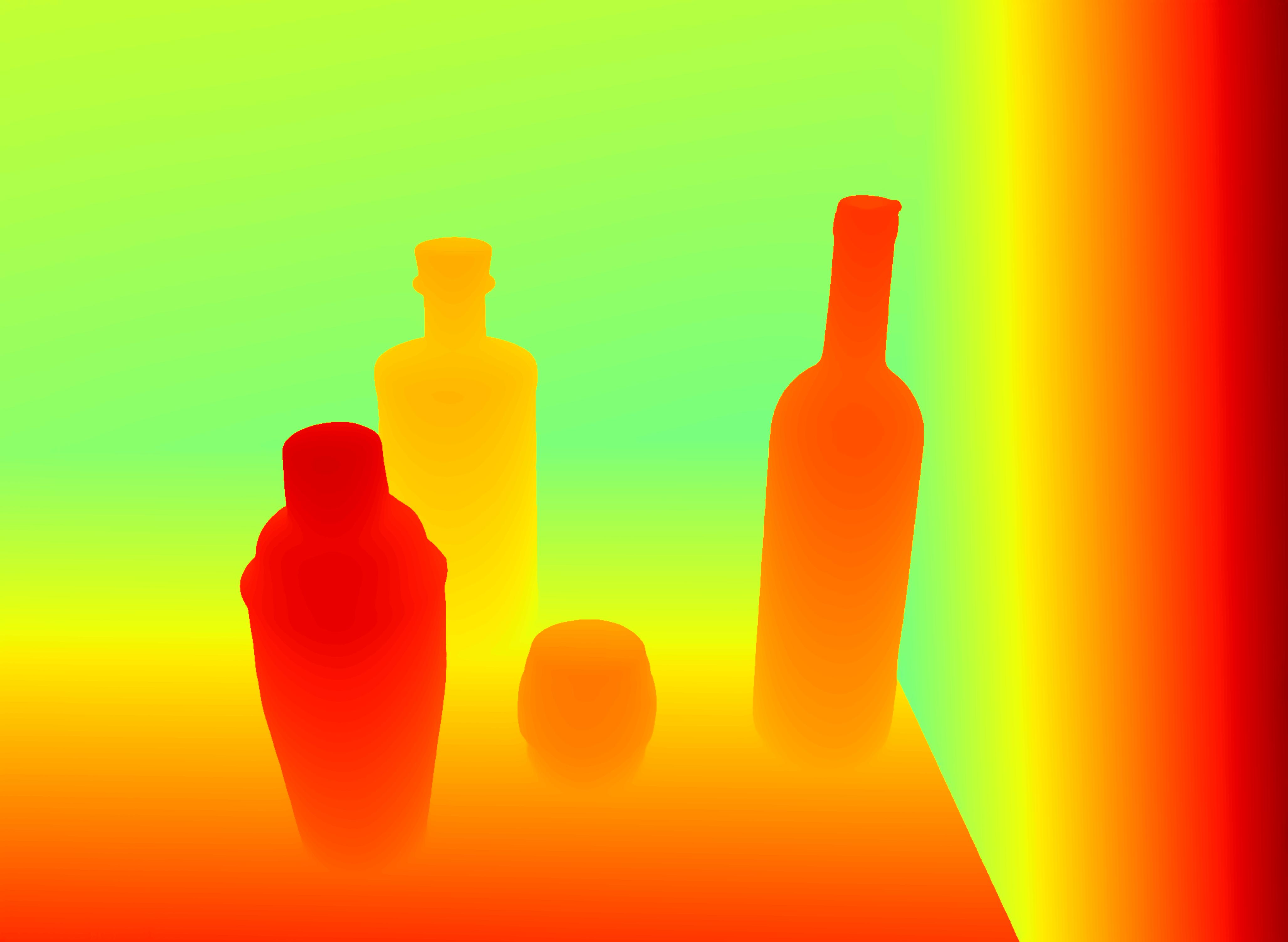} & 
\includegraphics[width=0.125\linewidth]{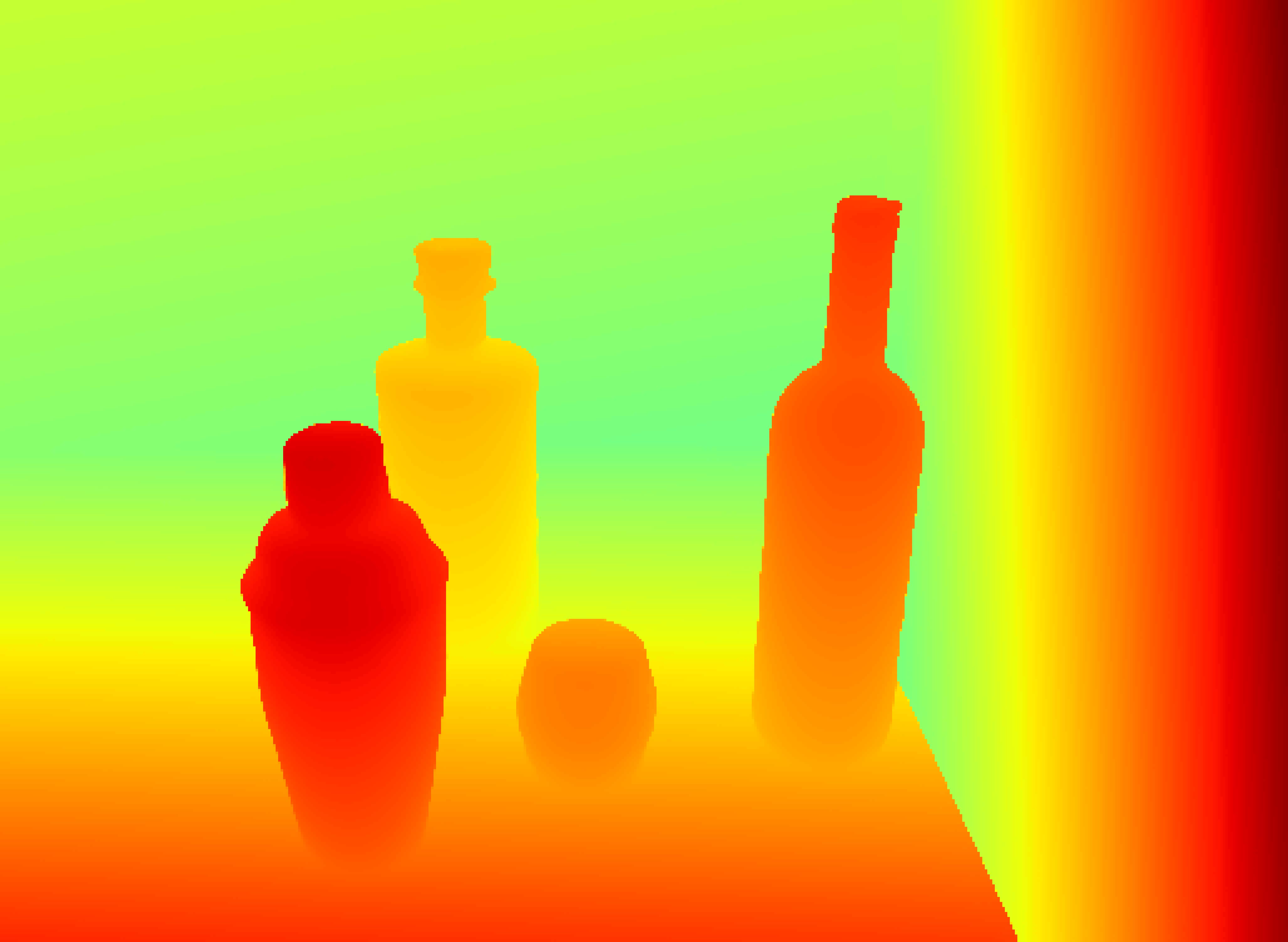} &
\includegraphics[width=0.125\linewidth]{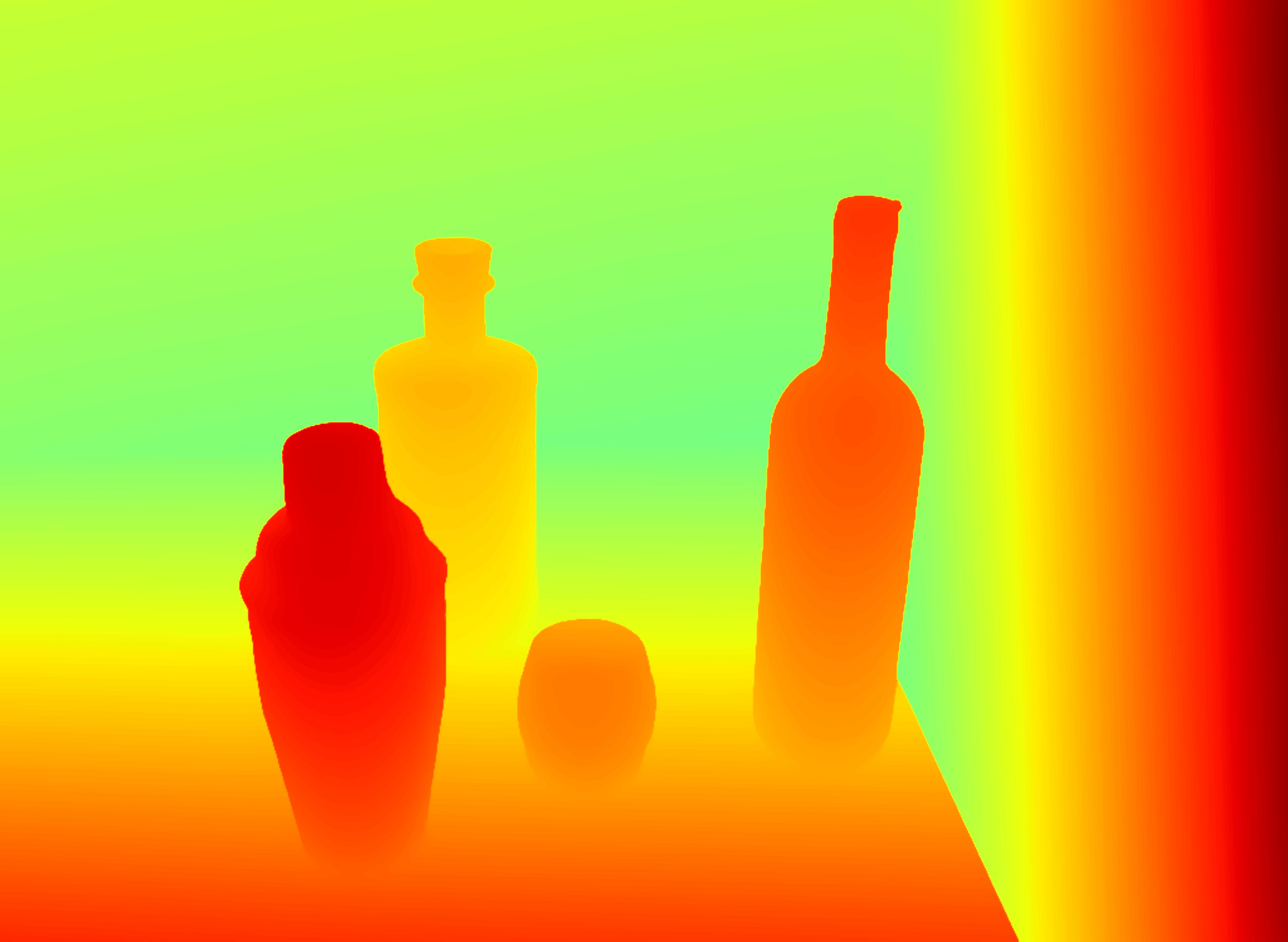 } & 
 \includegraphics[width=0.125\linewidth]{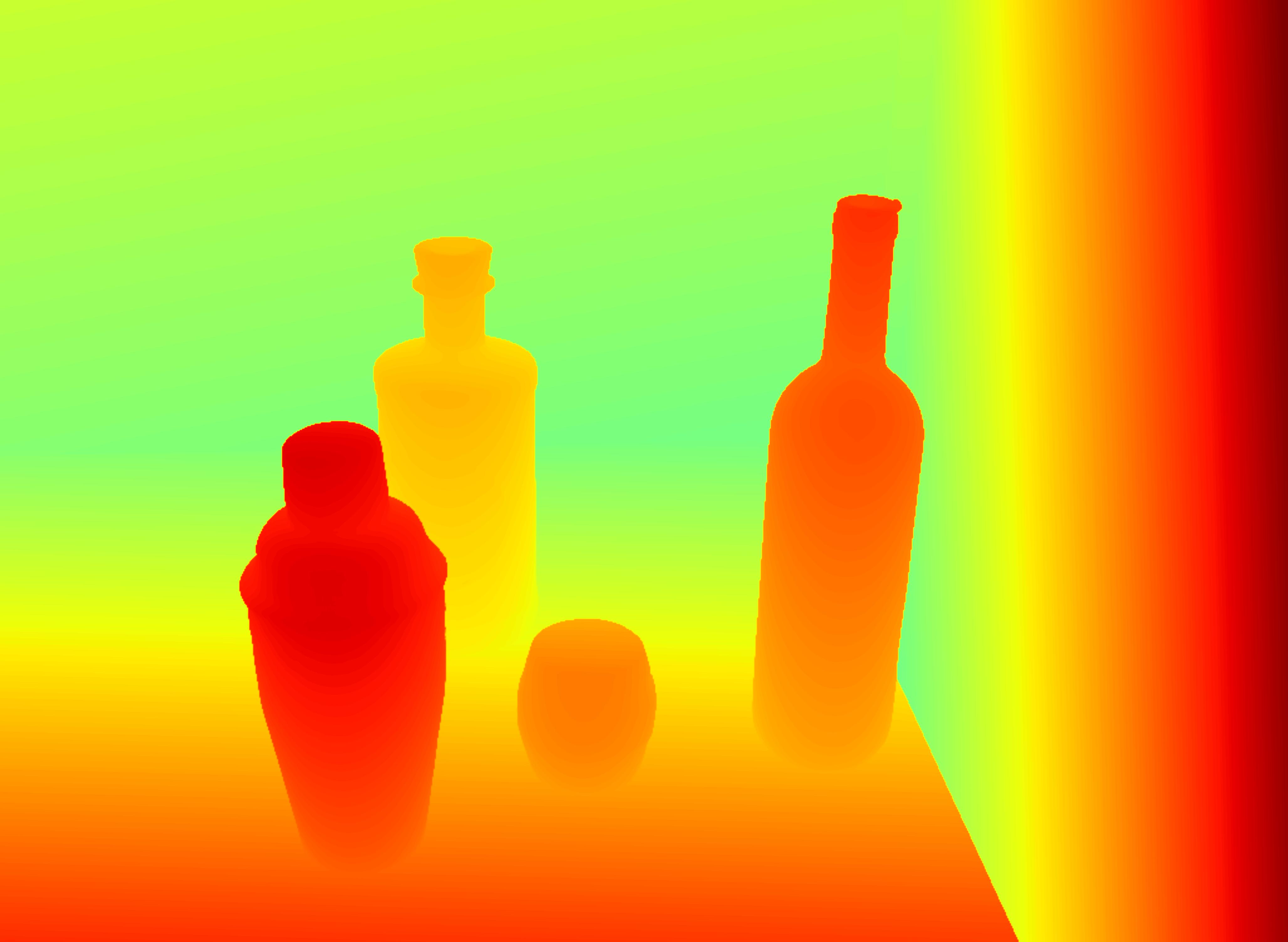} \\

\includegraphics[width=0.125\linewidth]{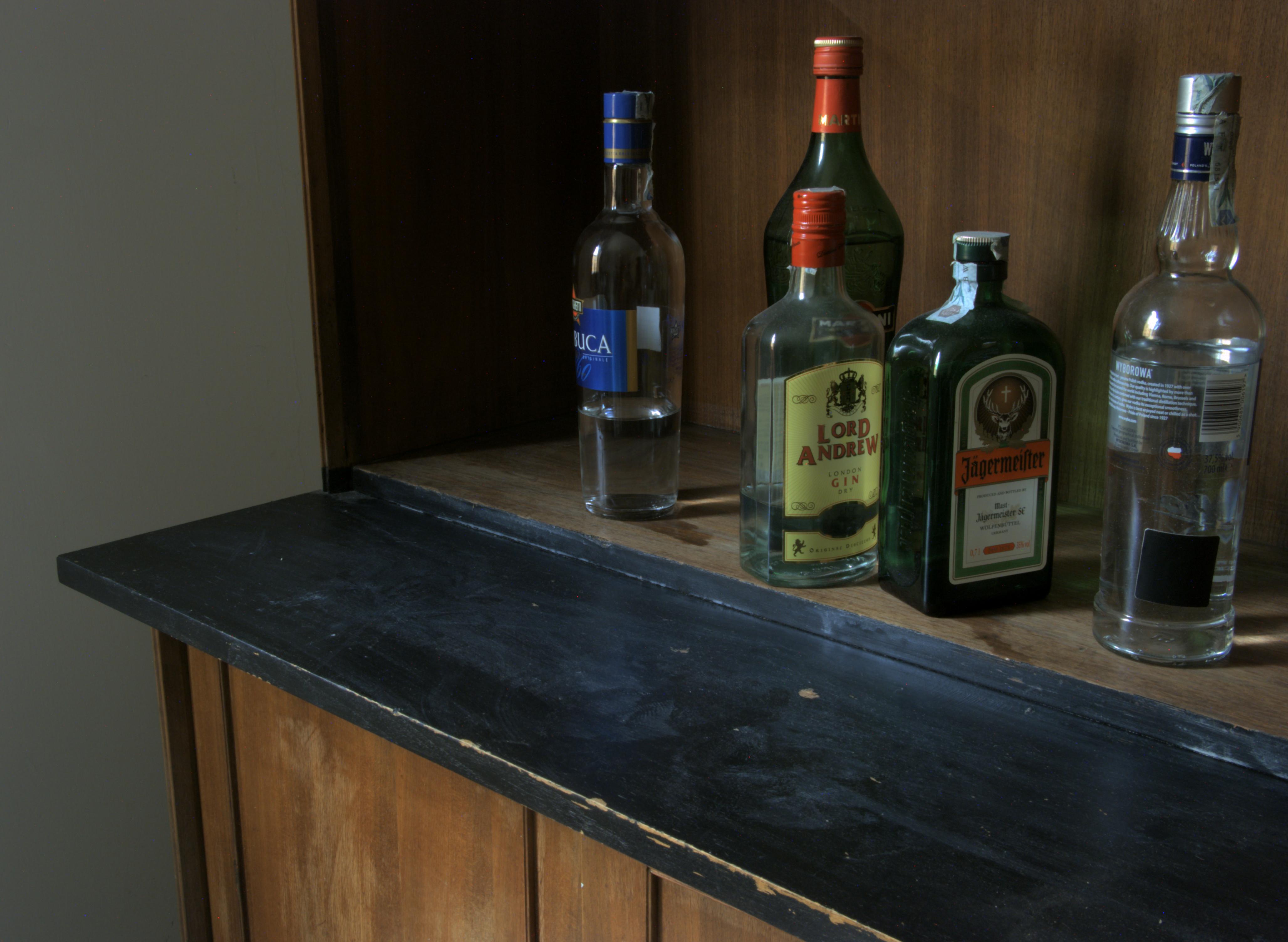} & \includegraphics[width=0.125\linewidth]{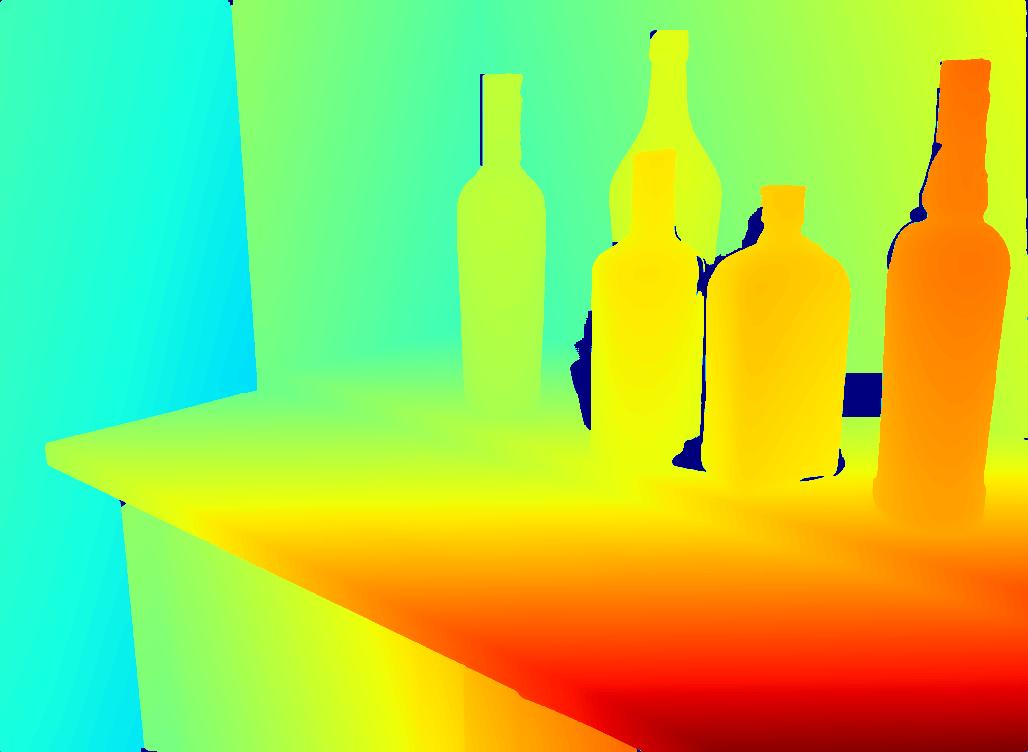 } & 
\includegraphics[width=0.125\linewidth]{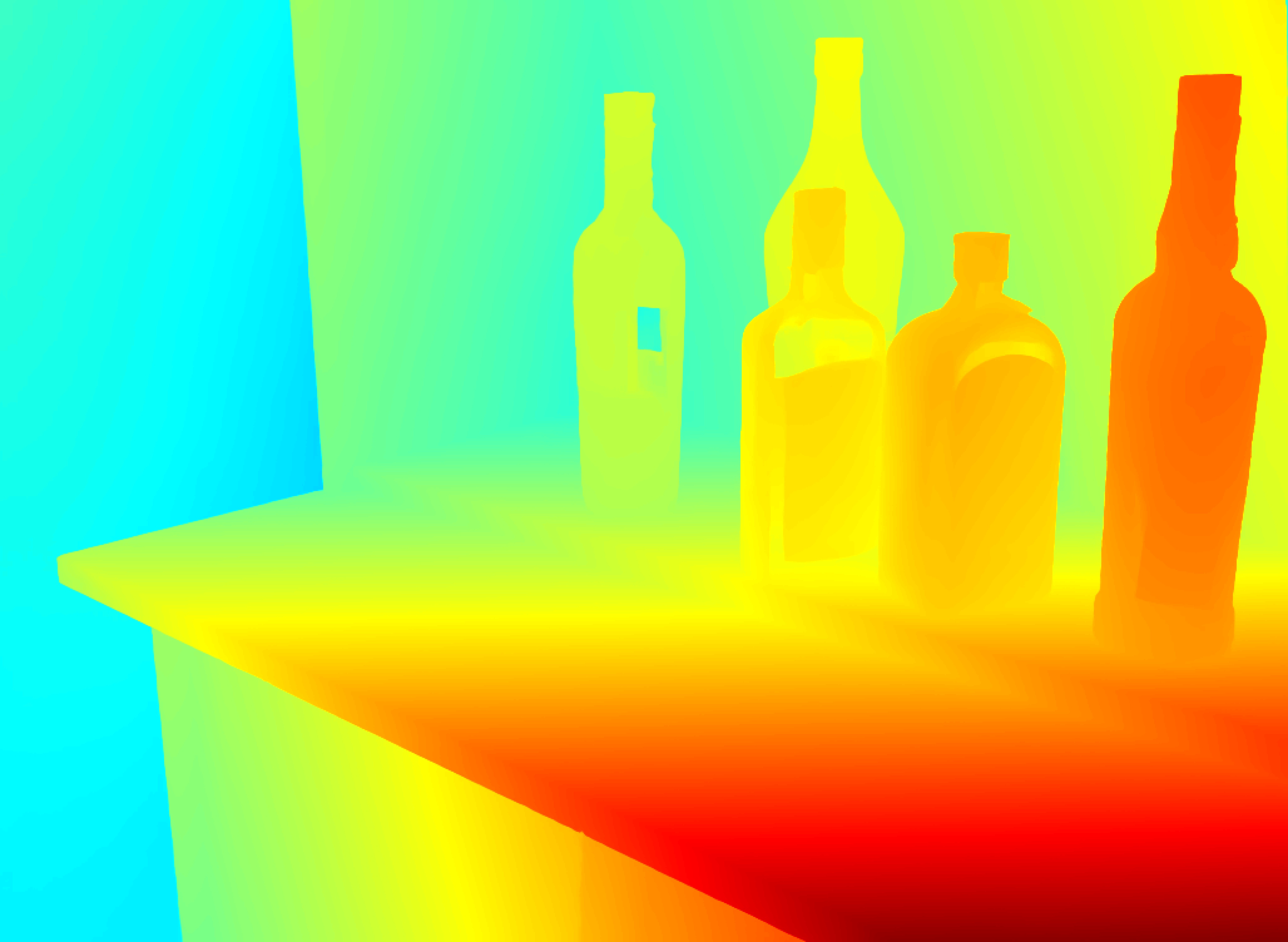 } & 
\includegraphics[width=0.125\linewidth]{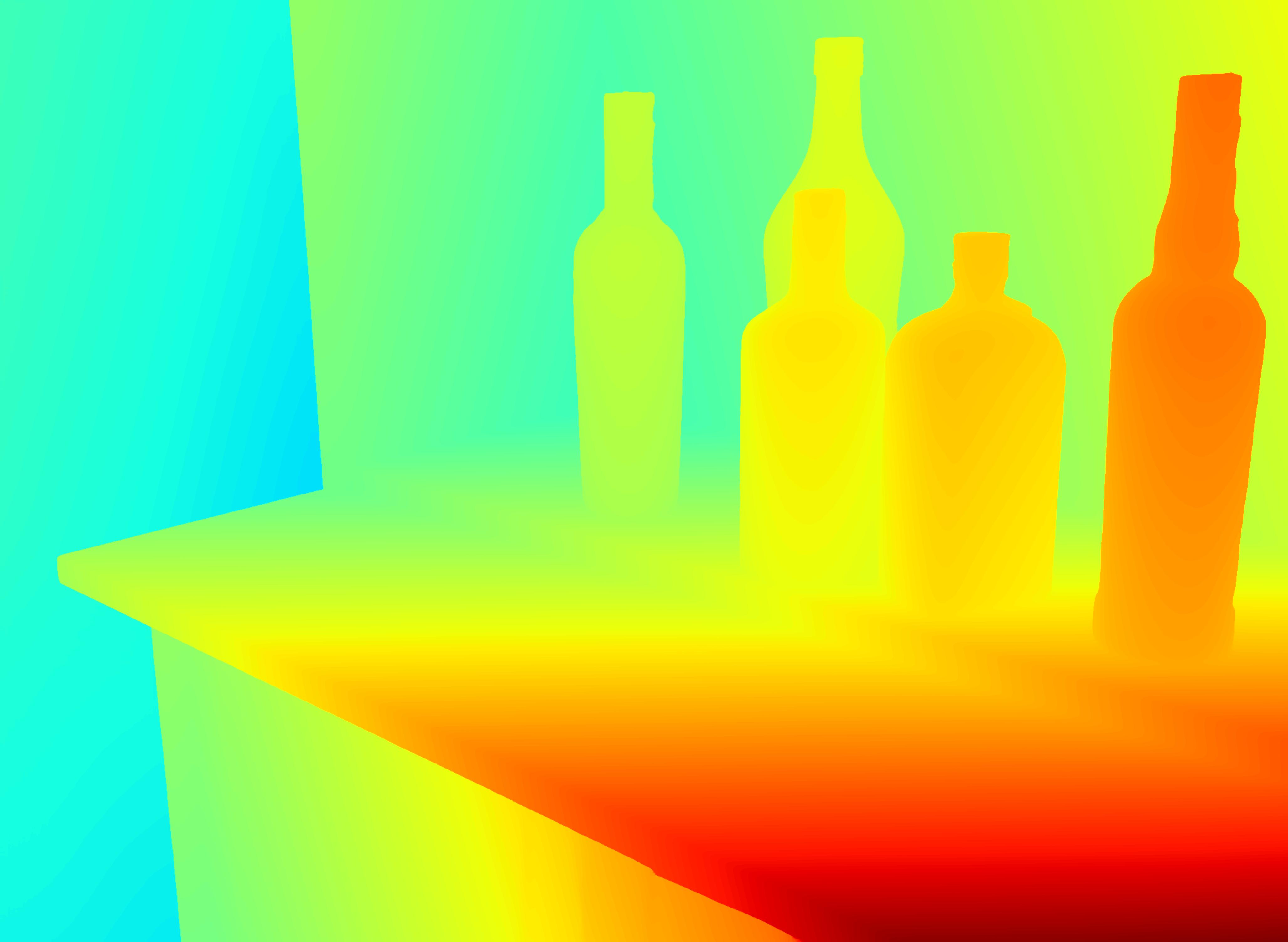} & 
\includegraphics[width=0.125\linewidth]{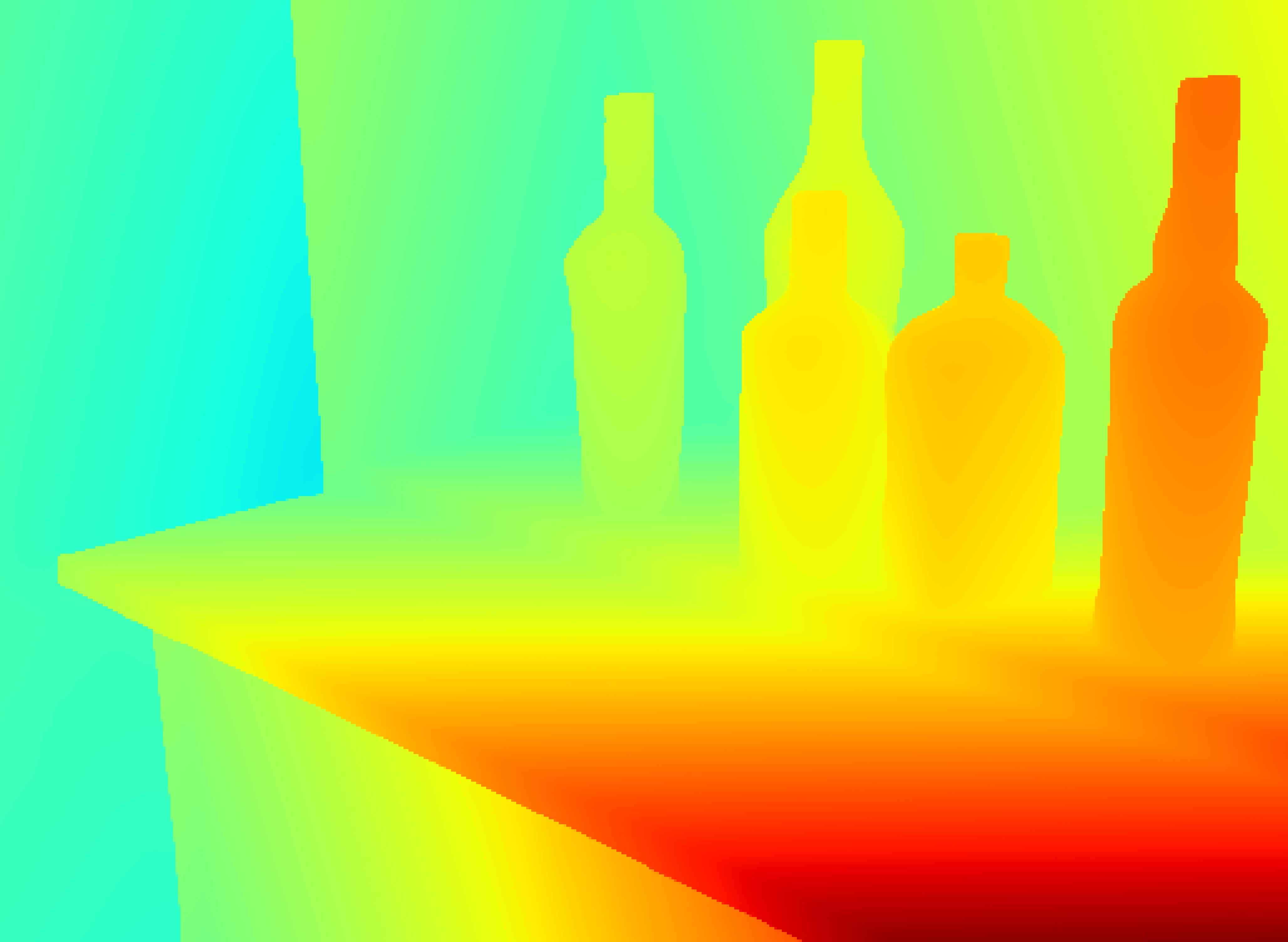} &
\includegraphics[width=0.125\linewidth]{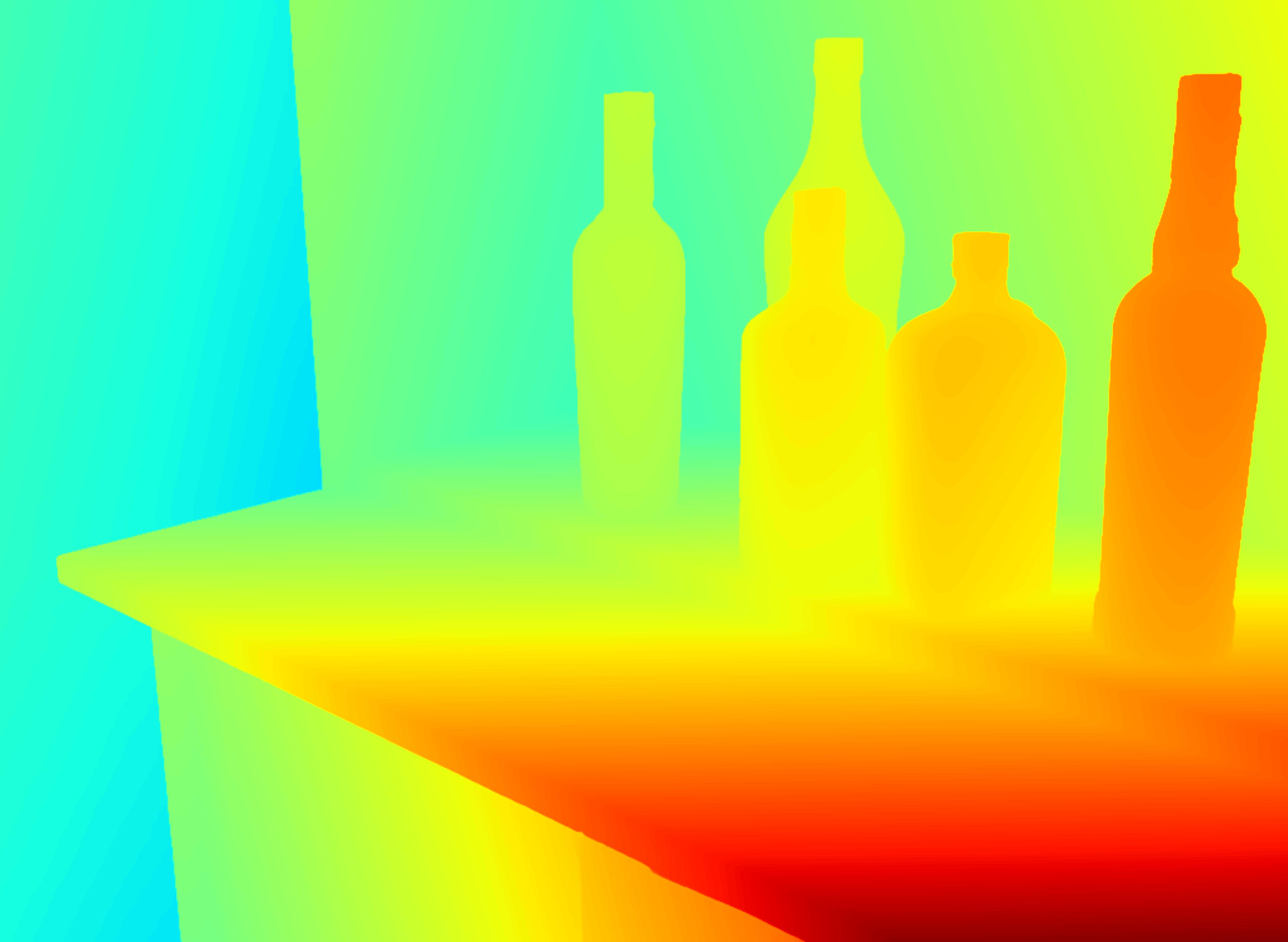 } & 
 \includegraphics[width=0.125\linewidth]{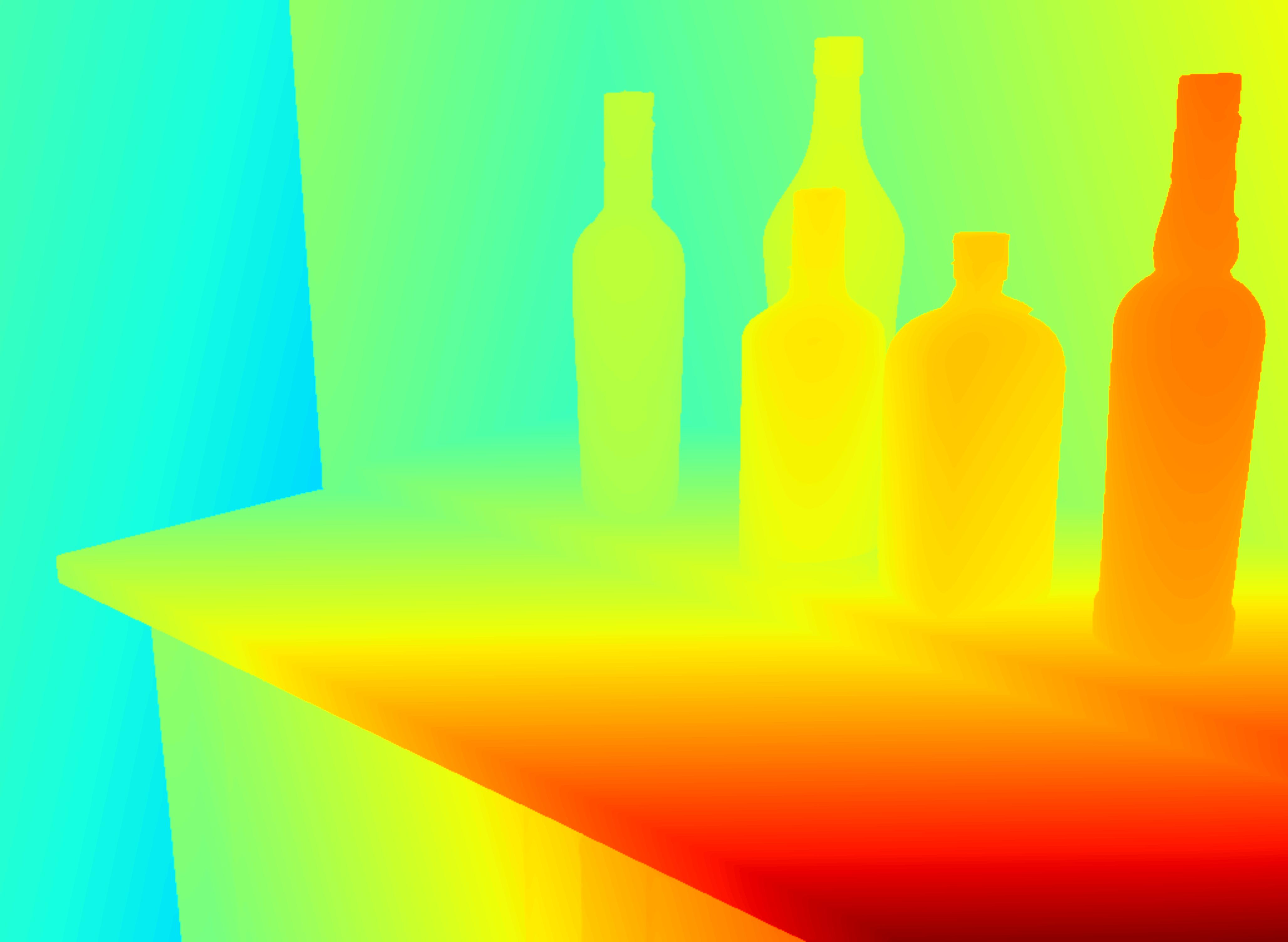} \\

\includegraphics[width=0.125\linewidth]{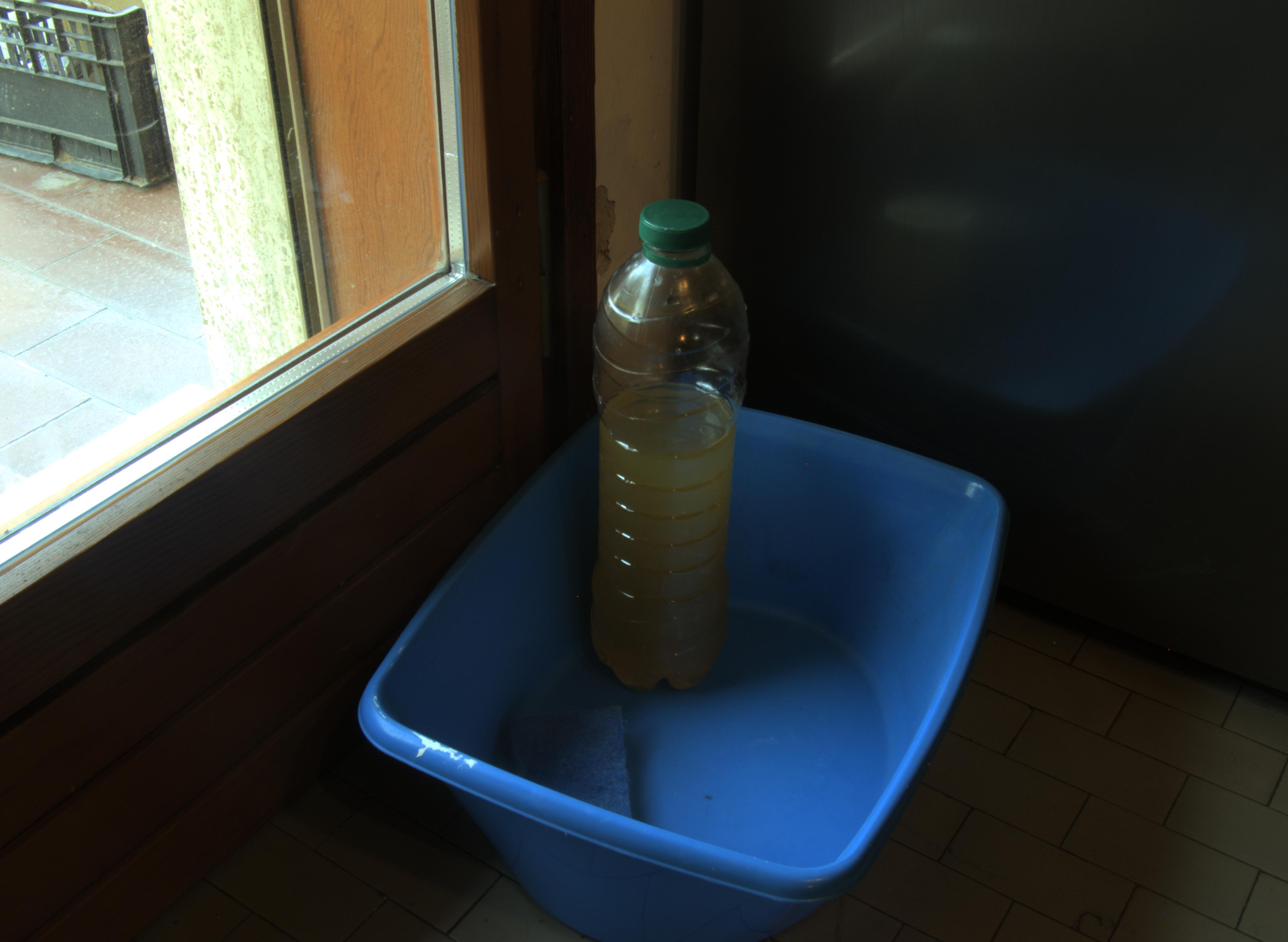} & \includegraphics[width=0.125\linewidth]{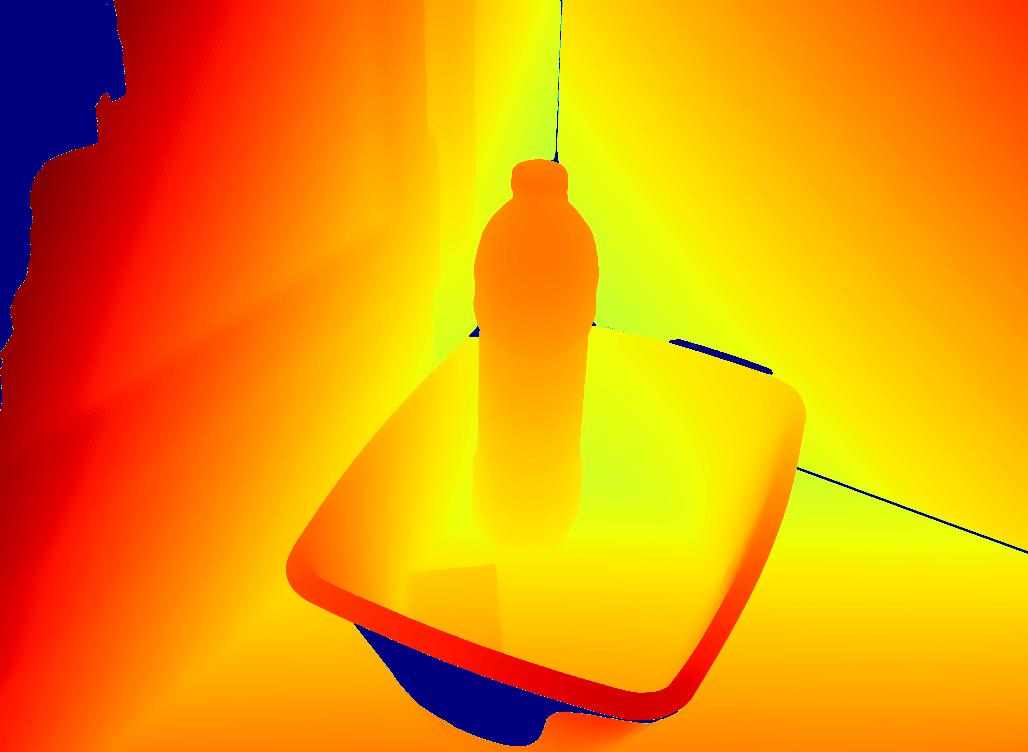 } & 
\includegraphics[width=0.125\linewidth]{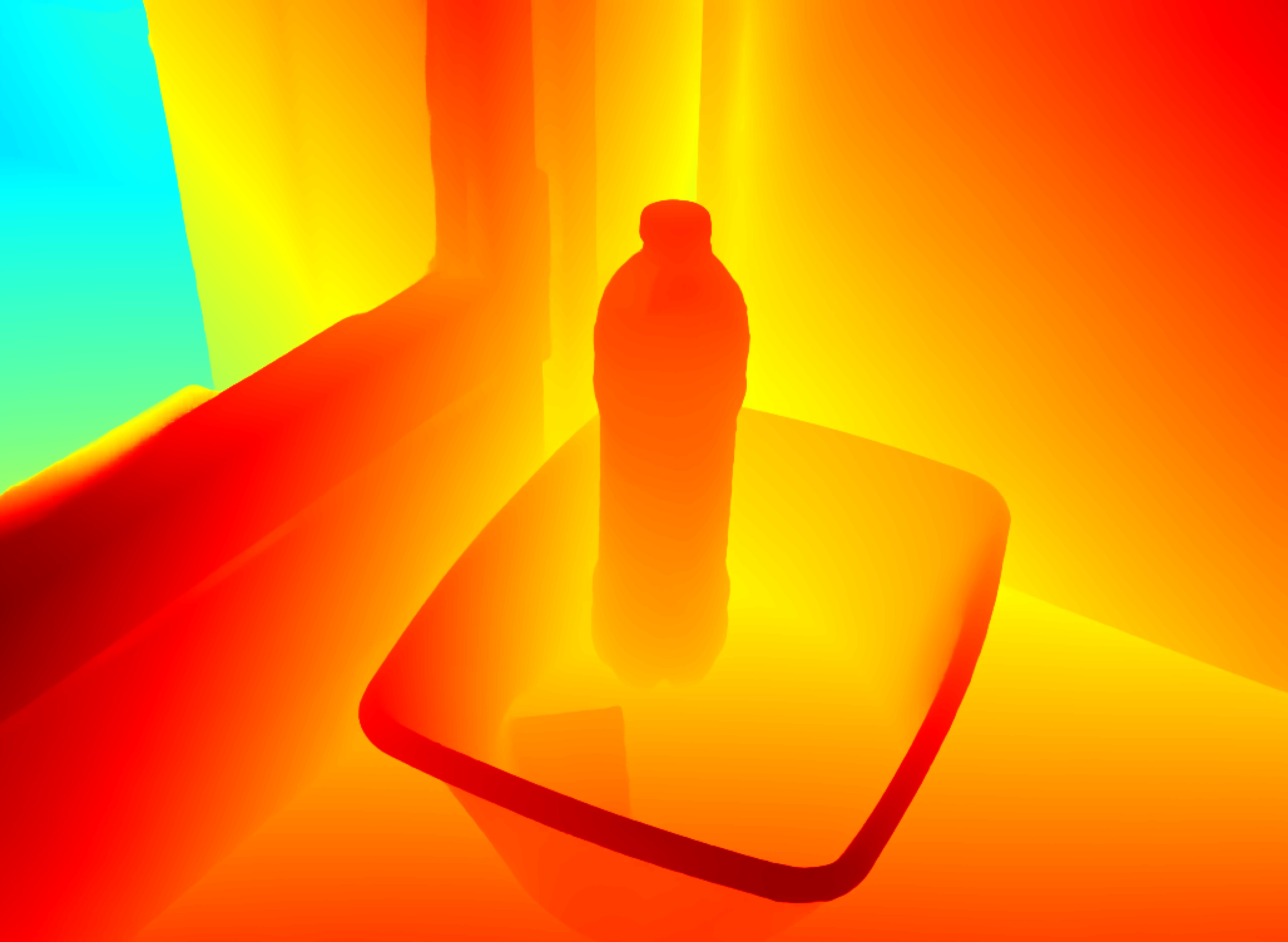 } & 
\includegraphics[width=0.125\linewidth]{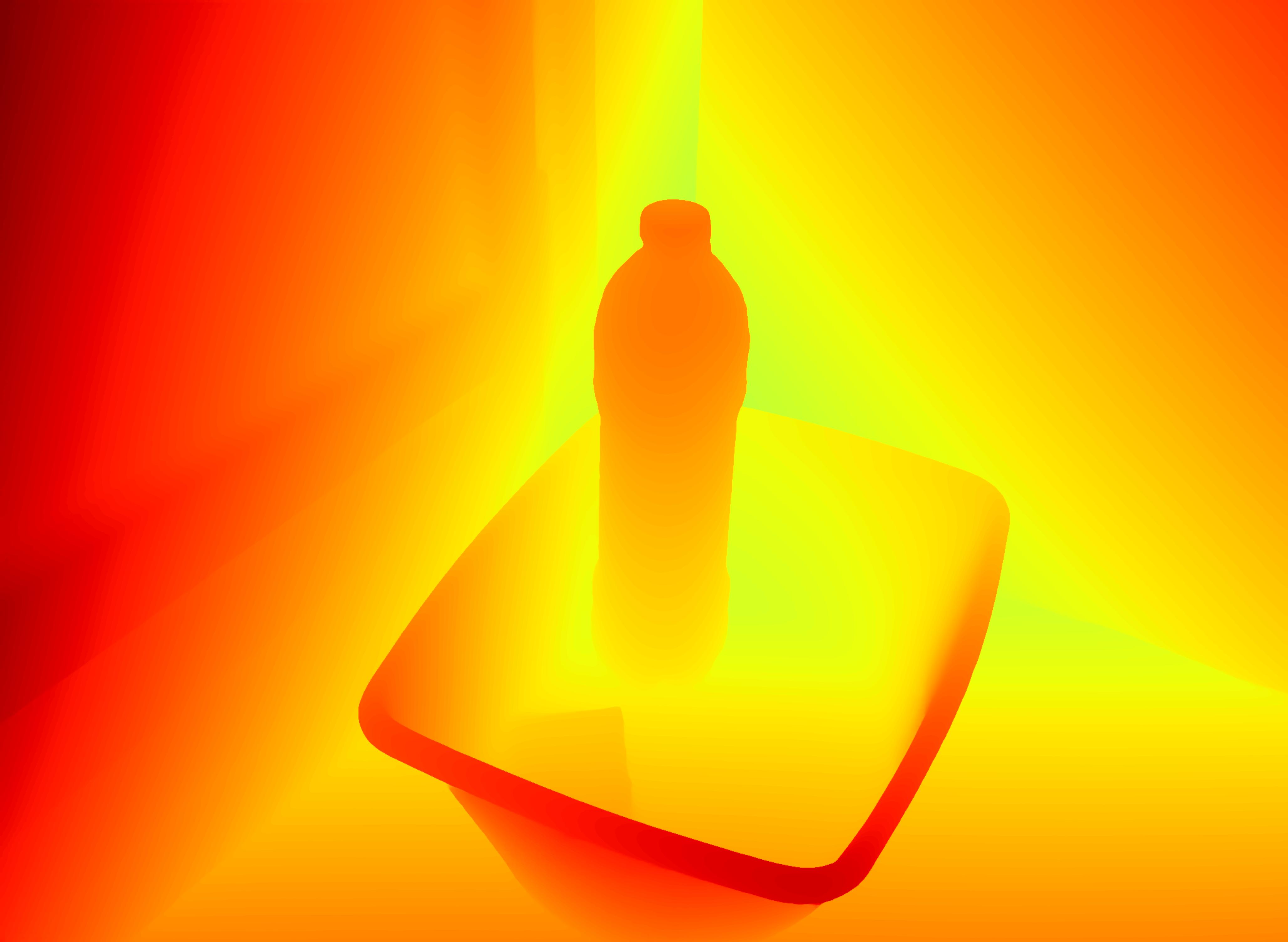} & 
\includegraphics[width=0.125\linewidth]{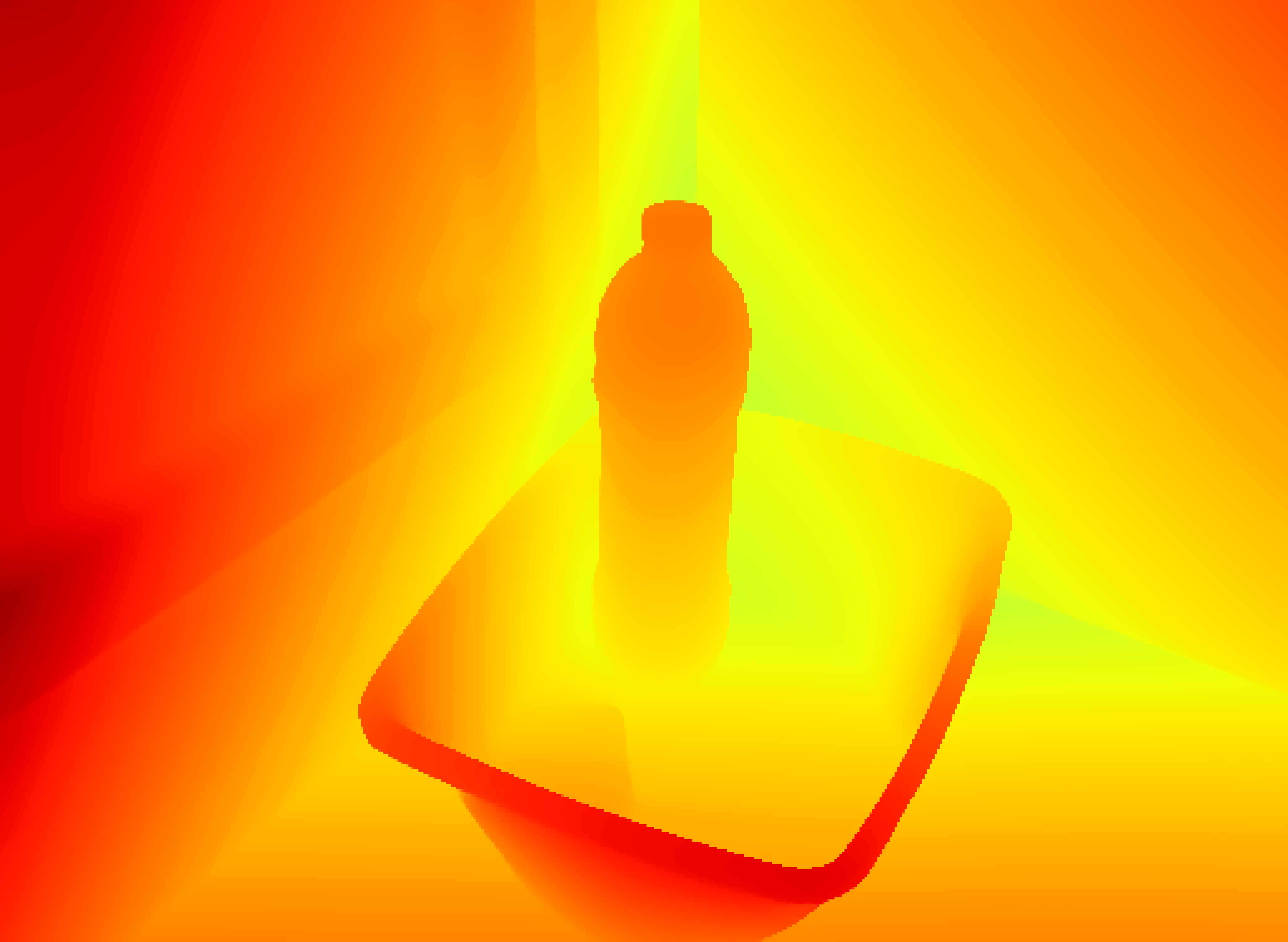} &
\includegraphics[width=0.125\linewidth]{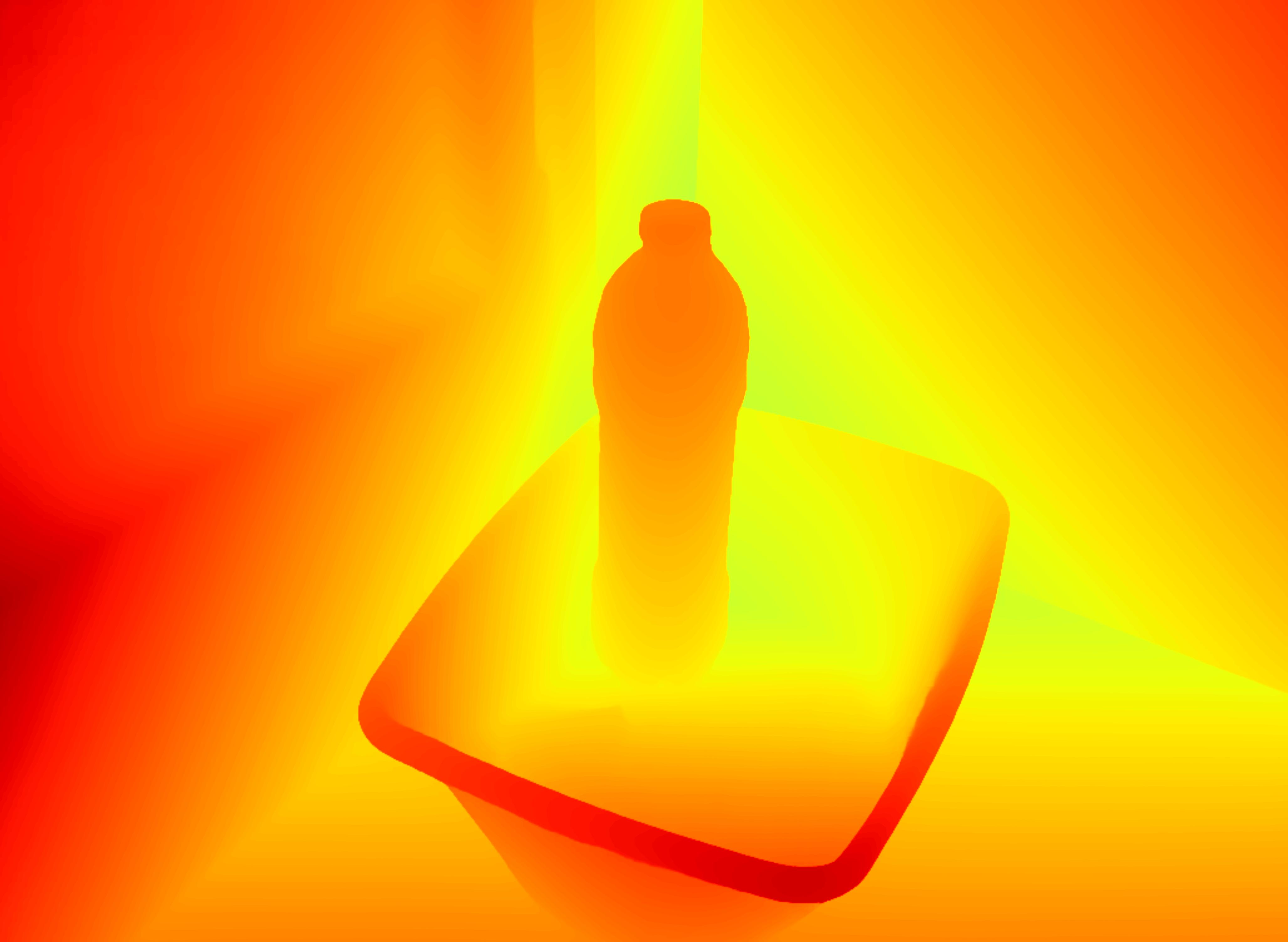 } & 
 \includegraphics[width=0.125\linewidth]{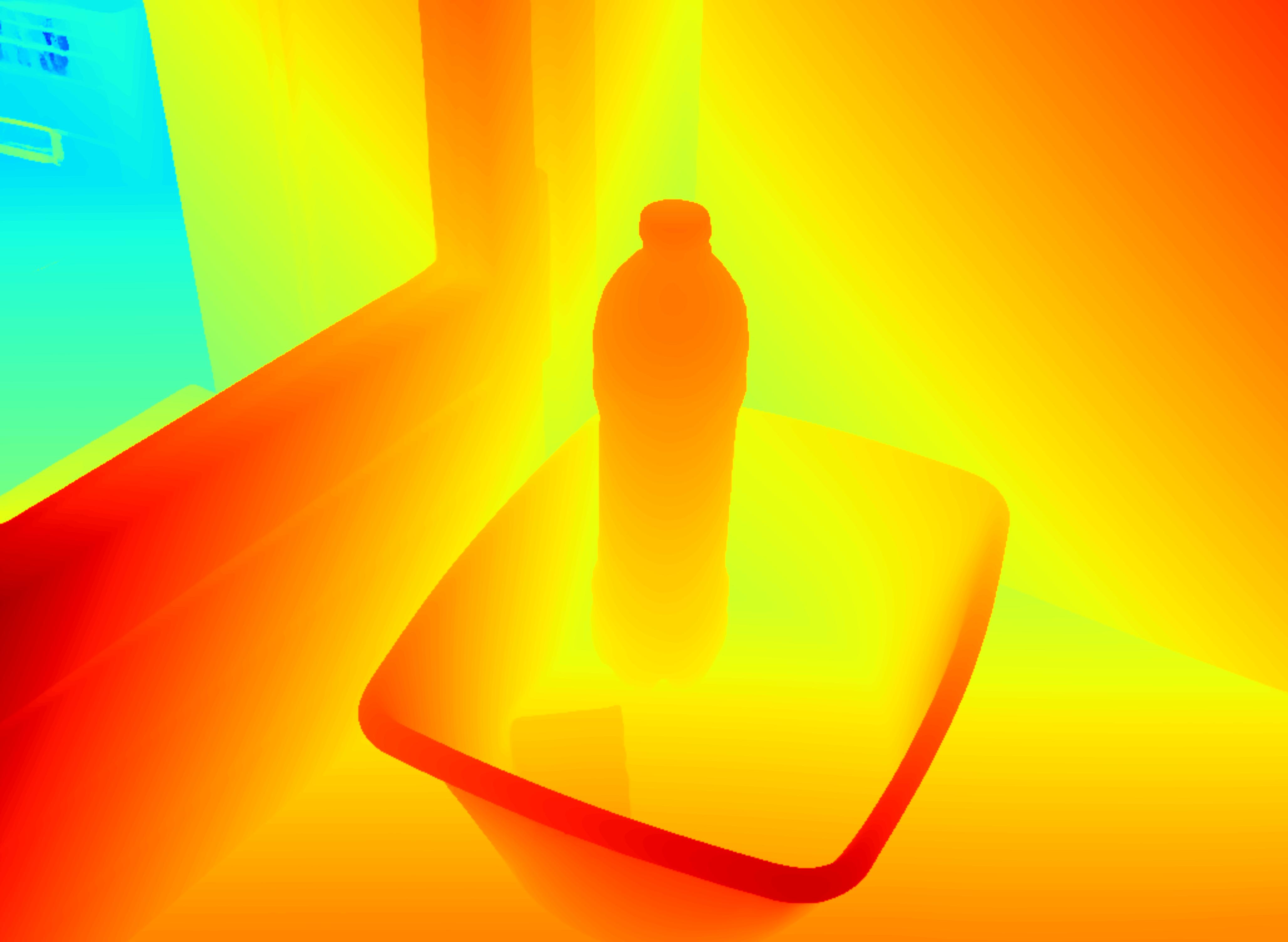} \\

\includegraphics[width=0.125\linewidth]{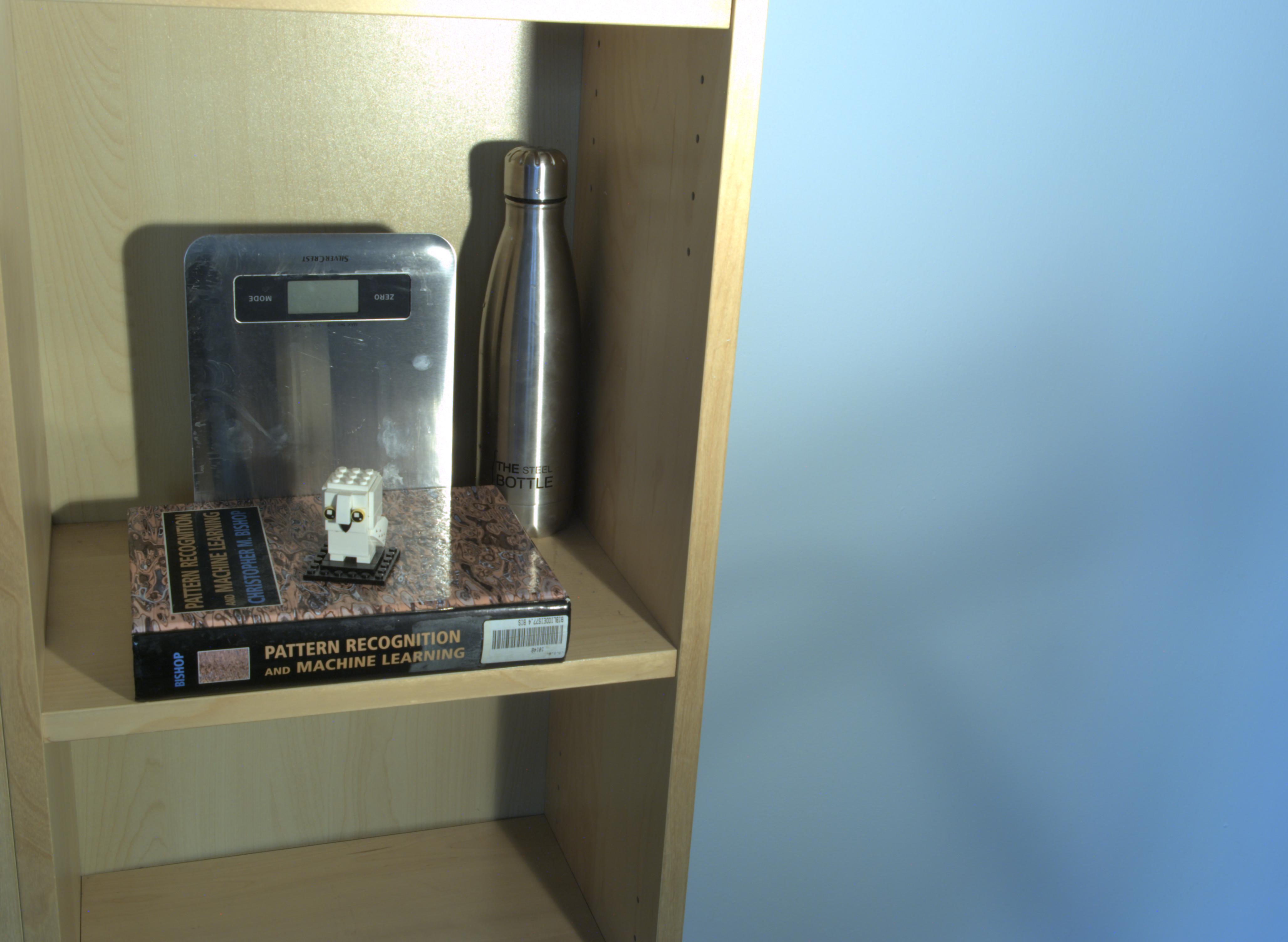} & \includegraphics[width=0.125\linewidth]{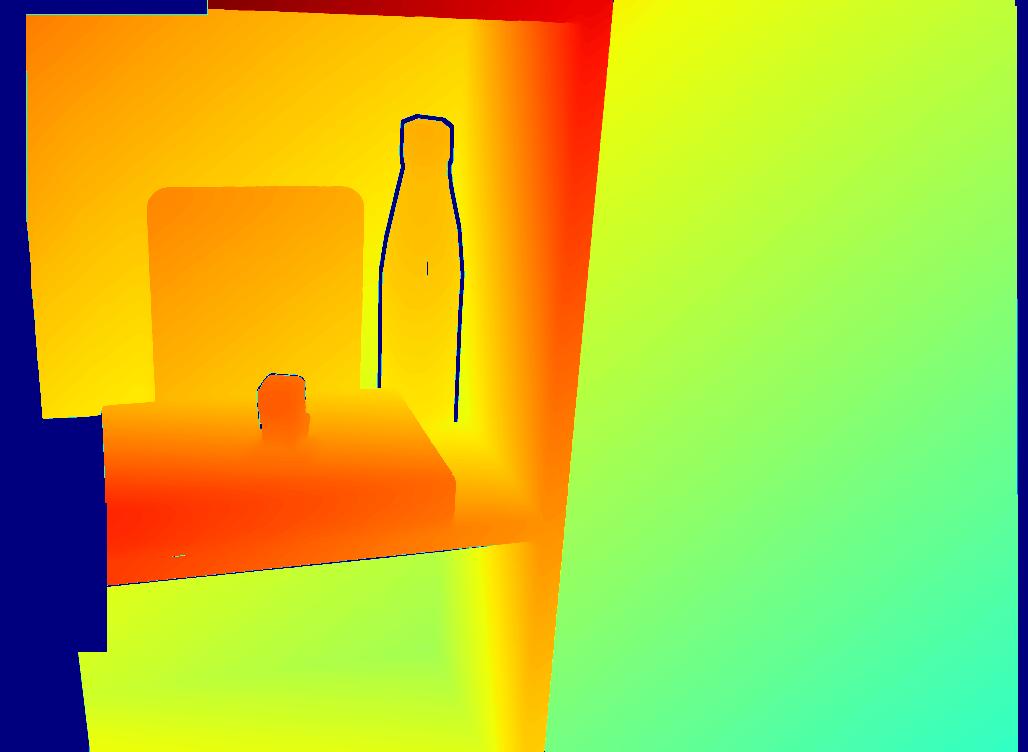 } & 
\includegraphics[width=0.125\linewidth]{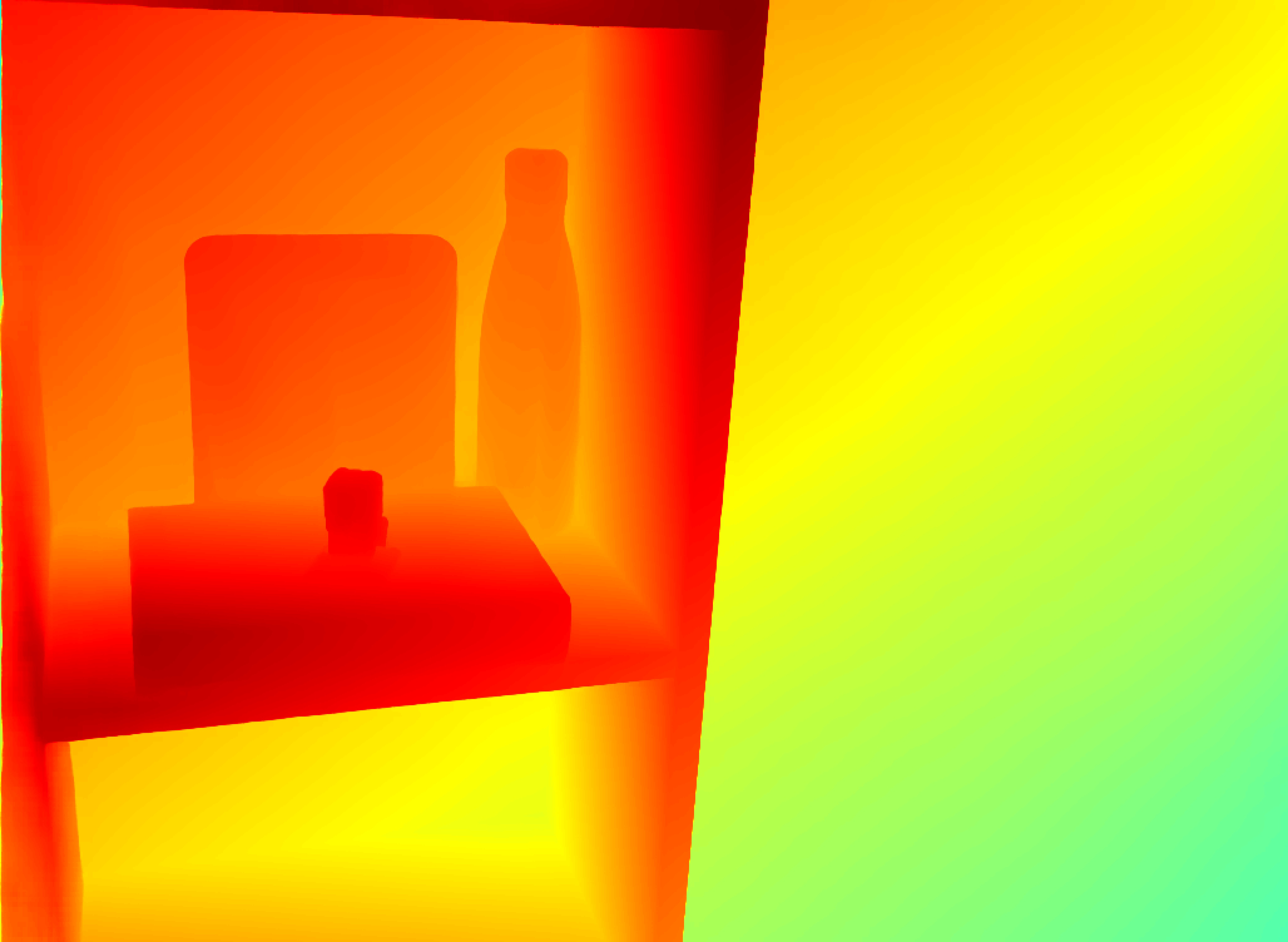 } & 
\includegraphics[width=0.125\linewidth]{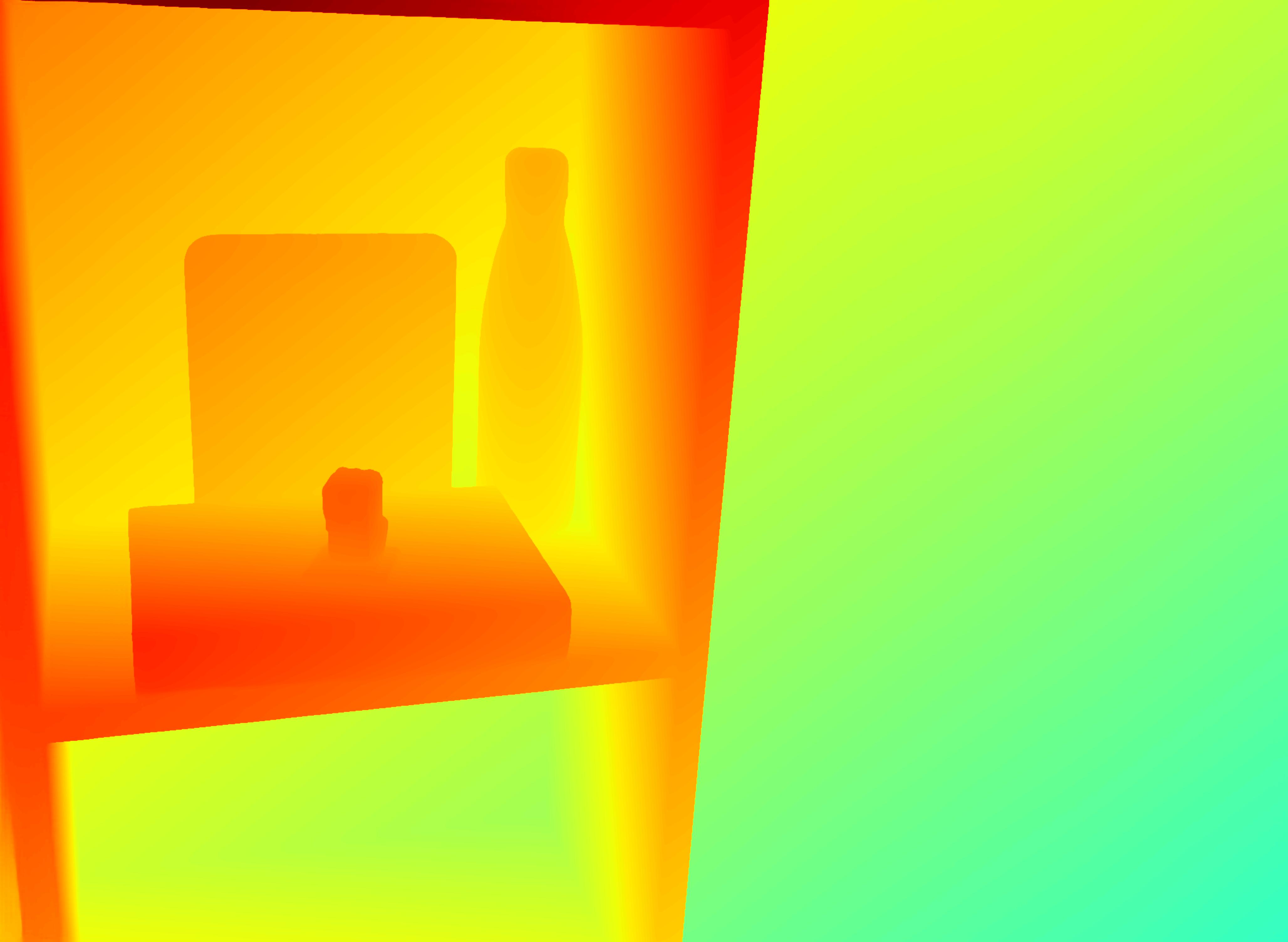} & 
\includegraphics[width=0.125\linewidth]{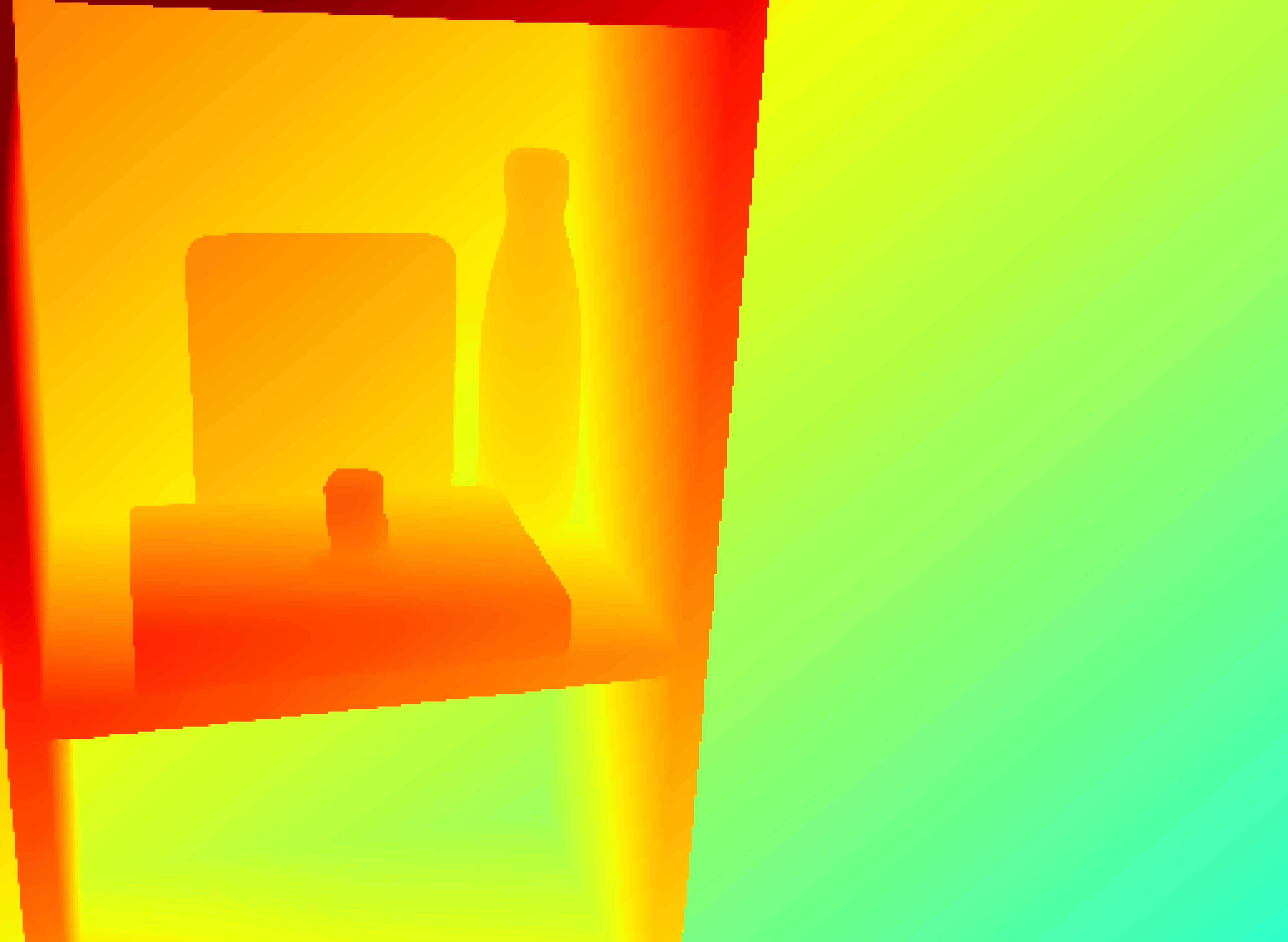} &
\includegraphics[width=0.125\linewidth]{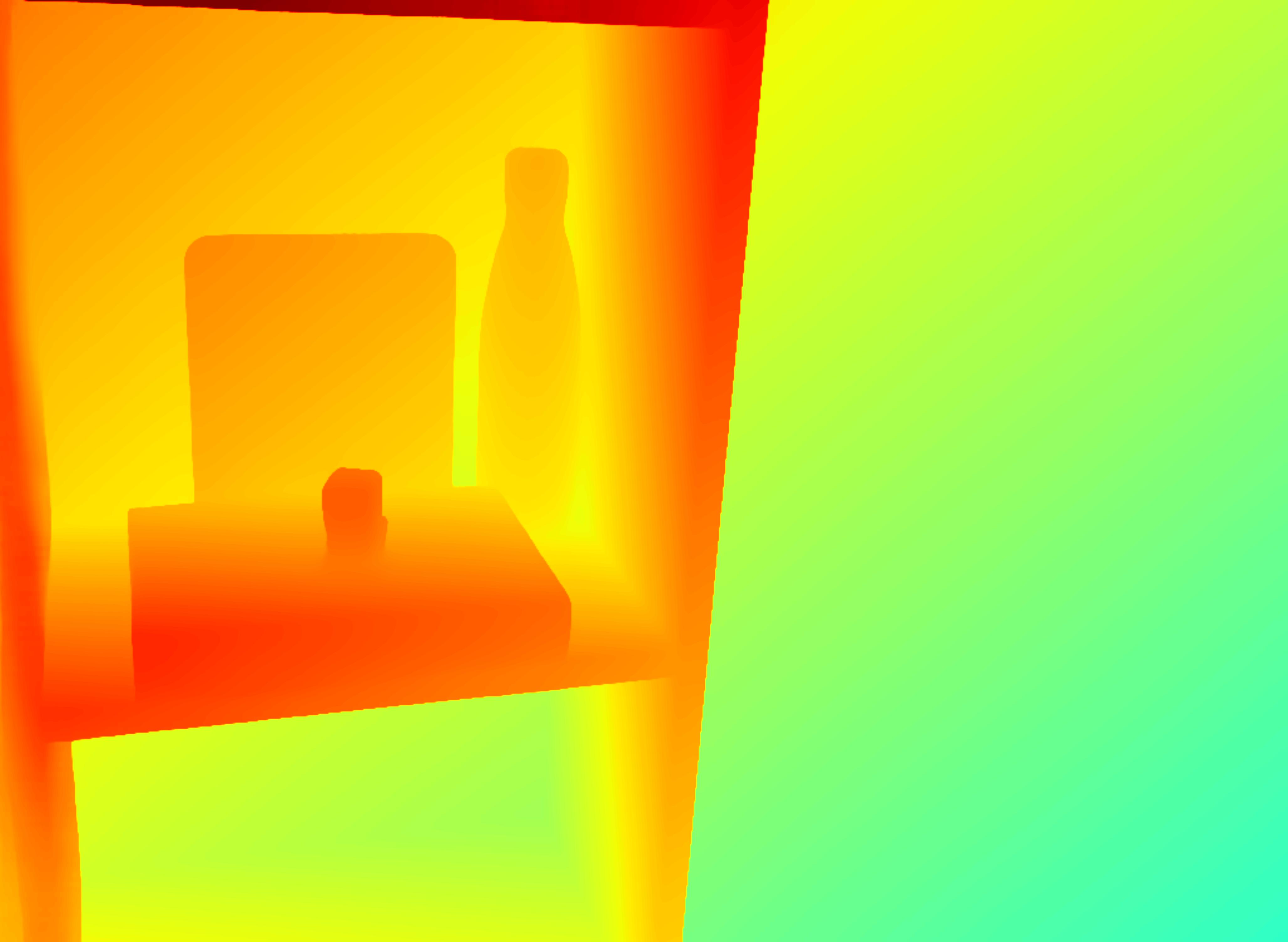 } & 
 \includegraphics[width=0.125\linewidth]{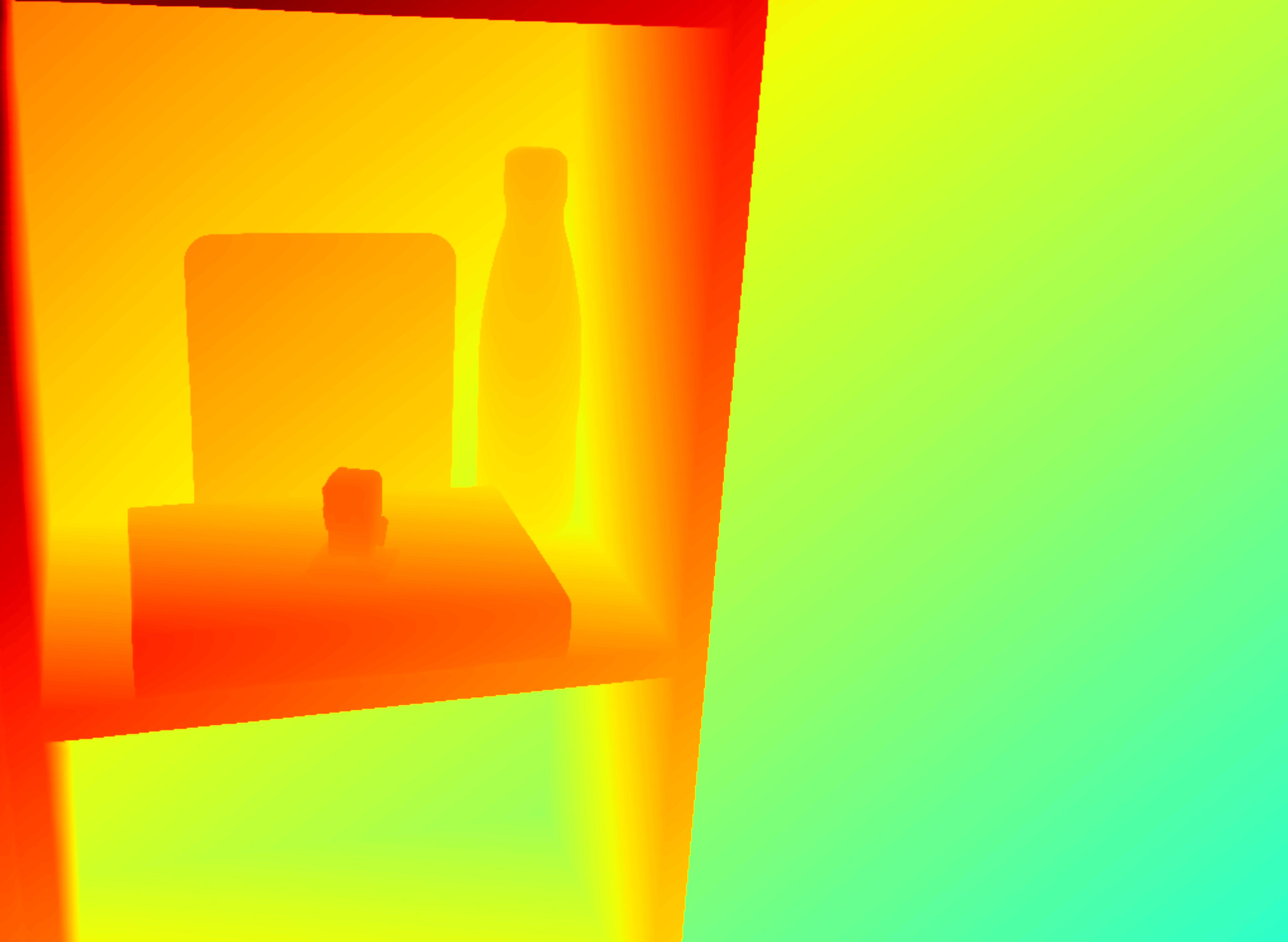} \\

\end{tabular}\vspace{-0.3cm}
\caption{\textbf{Qualitative results -- Stereo track.} From left to right: RGB reference image, ground-truth disparity, predictions by CREStereo \cite{li2022practical}, SRC-B, Robot01-vRobotit, NJUST-KMG, and weouibaguette.}\vspace{-0.4cm}
\label{fig:stereo}
\end{figure*}

\section{NTIRE Challenge on HR Depth from Images of Specular and Transparent Surfaces}

We organize the NTIRE 2025 Challenge on HR Depth from Images of Specular and Transparent Surfaces to further push the community toward developing newer, state-of-the-art solutions to properly deal with high-resolution images and non-Lambertian surfaces -- such as mirrors and glasses.
We now outline the main characteristics of our challenge.

\textbf{Tracks.} This challenge is composed of two tracks: \textit{Stereo}, devoted to methods estimating the disparity between pairs of rectified images, and \textit{Mono}, which instead allows estimating depth from a single input image only.

\begin{itemize}
    \item \textbf{Track 1: Stereo.} This tracks demands the participants to obtain high-quality, high-resolution dense disparity maps from 12 Mpx stereo frames. The resolution itself represents one of the main challenges, as it is prohibitive for most state-of-the-art existing stereo networks. Furthermore, the presence of non-Lambertian objects violating the common assumptions made in stereo matching makes this track even harder.

    \item \textbf{Track 2: Mono.} In parallel, this track requires to estimate depth out of a single 12Mpx frame. In this case, the inherent ill-posed nature of the problem represents one of the main challenges. Additionally, the presence of objects belonging to the long-tail of training data used for this task -- such as transparent objects and mirrors -- further makes it more complex.
\end{itemize}

\textbf{Datasets.} We build our challenge around the Booster dataset \cite{booster,booster2}, composed of 419 high-resolution balanced and unbalanced stereo pairs, captured in 64 different scenes and respectively distributed into 228 and 191 pairs for training and testing purposes -- with 38 and 26 for the two sets respectively. A newer version of the dataset \cite{booster2} extended it with a second testing split, tailored for evaluating monocular depth estimation approaches over 187 single frames, captured in 21 different environments.

As in the previous editions \cite{ramirez2023ntire,ramirez2024ntire}, we use the original 228 training stereo pair as a shared \textit{training split}, common to both tracks. We then select two distinct \textit{validation splits}, by selecting frames with different illuminations from 3 scenes of the stereo and monocular testing splits -- i.e., \textit{Microwave, Mirror1, Pots} for the Stereo track, and \textit{Desk, Mirror3, Sanitaries} for the Mono track respectively, for a total of 15 validation samples for each track from total 26 and 28 available from the selected scenes. 
The remaining images in the two original testing splits become the official stereo and mono \textit{testing splits} for this challenge, for a total of 169 and 159 samples.

\textbf{Evaluation Protocol.} 
For each track, Stereo and Mono respectively, we adopt the official metrics reported on the Booster benchmark \cite{booster,booster2}. 
For Stereo track, we compute the percentage of pixels with disparity errors larger than a threshold $\tau$ (bad-$\tau$, with $\tau \in [2, 4, 6, 8]$), as well as the Mean Absolute Error (MAE) and Root Mean Squared Error (RMSE) in pixel. 
For Mono track, we compute the percentage of pixels having the maximum between the prediction/ground-truth and ground-truth/prediction ratios lower than a threshold ($\delta < i$, with $i$ being 1.05, 1.15, and 1.25) and the absolute error relative to the ground truth value (Abs Rel.), as well as the mean absolute error (MAE), and Root Mean Squared Error (RMSE).
Following the latest edition \cite{ramirez2024ntire}, we compute metrics on three different sets of pixels as in \cite{costanzino2023iccv}: \textit{ToM} regions -- i.e., those belonging to non-Lambertian surfaces -- \textit{All} pixels and \textit{Others} -- i.e., the difference between \textit{All} and \textit{ToM} sets.
To rank submissions and determine the winner, we use bad-2 and $\delta < 1.05$ -- respectively for Stereo and Mono tracks -- averaged over all pixels, highlighted in \textcolor{red}{\textbf{red}} in the tables. Specifically, we define two rankings based on performance on \textit{ToM} and \textit{All} regions, respectively\footnote{we will highlight that the two coincide on the Mono track}. 
Finally, as most state-of-the-art monocular networks estimate depth up to an unknown pair of scale and shift factors, 
before computing metrics we recover metric depth from predicted maps $\hat{d}$ as $\alpha\hat{d} + \beta$, with $\alpha,\beta$ being scale and shift factors.
Following \cite{Ranftl2022}, $\alpha,\beta$ are estimated with Least Square Estimation (LSE) regression over the ground truth depth map $d$:

\small{
\begin{equation}
\label{eq:rescaling}
    (\alpha,\beta) = \text{arg}\min_{\alpha,\beta} \sum_{p} \Big( \alpha \hat{d}(p) + \beta - d(p) \Big)^2
\end{equation}}
\normalsize where $p$ are the pixel locations having both predictions and ground truth depths available.

\begin{table*}[t]
    \setlength{\tabcolsep}{4pt}
    \centering
    \resizebox{\linewidth}{!}{
    \begin{tabular}{l||c|rrrrrr||c|rrrrrr||rrrrrr}
    & \multicolumn{7}{c||}{\cellcolor{blue!25}ToM} & \multicolumn{7}{c||}{\cellcolor{pink}All} & \multicolumn{6}{c}{\cellcolor{YellowOrange}Other} \\
    \hline
    Team & Rank & \textcolor{red}{$\boldsymbol{\delta} \boldsymbol{<} $\textbf{{1.05}}} & $\delta<1.15$ & $\delta<1.25$ & Abs Rel. & MAE & RMSE & Rank & \textcolor{red}{$\boldsymbol{\delta} \boldsymbol{<} $\textbf{{1.05}}} & $\delta<1.15$ & $\delta<1.25$ & Abs Rel. & MAE & RMSE & $\delta<1.05$ & $\delta<1.15$ & $\delta<1.25$ & Abs Rel. & MAE & RMSE \\
    \hline
    \textbf{Lavreniuk} & \gold{\#1} & \gold{87.67} & \bronze{99.15} & \silver{99.84} & \gold{2.54} & \silver{2.71} & \gold{3.54} & \gold{\#1} & \gold{84.19} & \silver{97.73} & \gold{99.50} & \silver{3.63} & \silver{3.12} & \gold{6.02} & \gold{82.13} & \silver{96.93} & \gold{99.27} & \silver{4.11} & \silver{3.39} & \gold{6.94} \\
    \textbf{colab} & \silver{\#2} & \silver{86.64} & \gold{99.60} & \bronze{99.83} & \silver{2.59} & \gold{2.58} & \bronze{3.76} & \silver{\#2} & \silver{82.93} & \gold{98.53} & \silver{99.46} & \gold{3.53} & \gold{2.97} & \silver{6.11} & \silver{80.22} & \gold{97.88} & \gold{99.27} & \gold{4.03} & \gold{3.33} & \silver{6.95} \\
    \textbf{PreRdw} & \bronze{\#3} & \bronze{84.58} & \silver{99.42} & \gold{99.86} & \bronze{2.70} & \bronze{2.79} & \silver{3.64} & \bronze{\#3} & \bronze{80.47} & \bronze{96.73} & \bronze{98.45} & \bronze{3.99} & \bronze{3.41} & \bronze{6.59} & \bronze{79.10} & \bronze{95.95} & \bronze{98.15} & \bronze{4.41} & \bronze{3.65} & \bronze{7.46} \\
    \textbf{IPCV} & \#4 & 62.66 & 95.43 & 99.19 & 4.60 & 4.70 & 6.01 & \#4 & 65.61 & 91.54 & 97.34 & 6.37 & 5.30 & 9.70 & 62.70 & 90.47 & 97.10 & 7.00 & 5.63 & 10.67 \\
    \hline
    \textbf{ZoeDepth [Baseline]} & \#5 & 45.21 & 82.27 & 93.06 & 8.04 & 8.71 & 9.57 & \#5 & 61.31 & 87.97 & 94.38 & 7.60 & 6.38 & 10.88 & 60.23 & 87.43 & 93.71 & 8.34 & 6.31 & 12.18 \\
    \hline
    \end{tabular}
    }
    \vspace{-0.3cm}
    \caption{\textbf{Mono Track: Evaluation on the Challenge Test Set.} Predictions evaluated at full resolution (4112$\times$3008) on All pixels and pixels belonging to ToM (Transparent or Mirror) or Other materials. In \gold{gold}, \silver{silver}, and \bronze{bronze}, we show first, second, and third-rank approaches, respectively. We rank methods on two metrics, \textcolor{red}{$\boldsymbol{\delta} \boldsymbol{<} $\textbf{{1.05}}} computed on either ToM or All pixels.
    }\vspace{-0.3cm}
    \label{tab:mono}
\end{table*}

\section{Challenge Results}\label{sec:results}

For each track, four teams participated in the final evaluation phase, with their outcomes detailed in Sections \ref{sec:results_stereo} and \ref{sec:results_mono}.
A brief explanation of each approach for both stereo and mono tracks is provided in Section \ref{sec:description_stereo} and Section \ref{sec:description_mono}, while the team composition is detailed in Sections \ref{sec:teams_stereo} and \ref{sec:teams_mono}.

\subsection{Track 1: Stereo}\label{sec:results_stereo}

Table \ref{tab:stereo} reports the results for this first track. 
At the bottom, we report the baseline -- i.e., CREStereo \cite{li2022practical}. 
From left to right, we report bad-$\tau$ metrics, MAE, and RMSE metrics for \textit{Tom}, \textit{All}, and \textit{Other} pixels respectively. On the right of the team's name, we report their overall rank, computed according to bad-2 errors -- the most restrictive metric -- on \textit{ToM} and \textit{All} regions.

All submitted methods outperformed the baseline on \textit{ToM} and \textit{All} pixels, with \textbf{SRC-B} achieving the lowest error rates on \textit{ToM} pixels and \textbf{weouibaguette} \textit{All} pixels, while the baseline still performs the best on \textit{Other} pixels.

Interestingly, unlike previous iterations of the challenge, while there is a somewhat clear trend on \textit{ToM} pixels, such as low bad-2 errors correlating with low MAE and RMSE scores, the results \textit{Other} and \textit{All} are consistently mixed up, making it hard to identify a jack-of-all-trades model.
Fig. \ref{fig:stereo} depicts some qualitative results from the stereo testing set. 
We can appreciate how the submitted methods tend to deal better with some specific challenges, such as the bottles in row 2, which CREStereo falters to infer without discontinuities, and the book in column 4, where the submitted methods produce much smoother results.

\begin{figure*}[t]
\setlength{\tabcolsep}{1pt}
    \centering
    \begin{tabular}{cccccccc}
\scriptsize\textit{\textbf{RGB}} & \scriptsize\textbf{\textit{GT}} & \scriptsize\textbf{\textit{ZoeDepth}}\cite{bhat2023zoedepth} & \scriptsize\textbf{\textit{Lavreniuk}} & \scriptsize\textbf{\textit{colab}} & \scriptsize\textbf{\textit{PreRdW}} & \scriptsize\textbf{\textit{IPCV}} \\    
 
\includegraphics[width=0.125\linewidth]{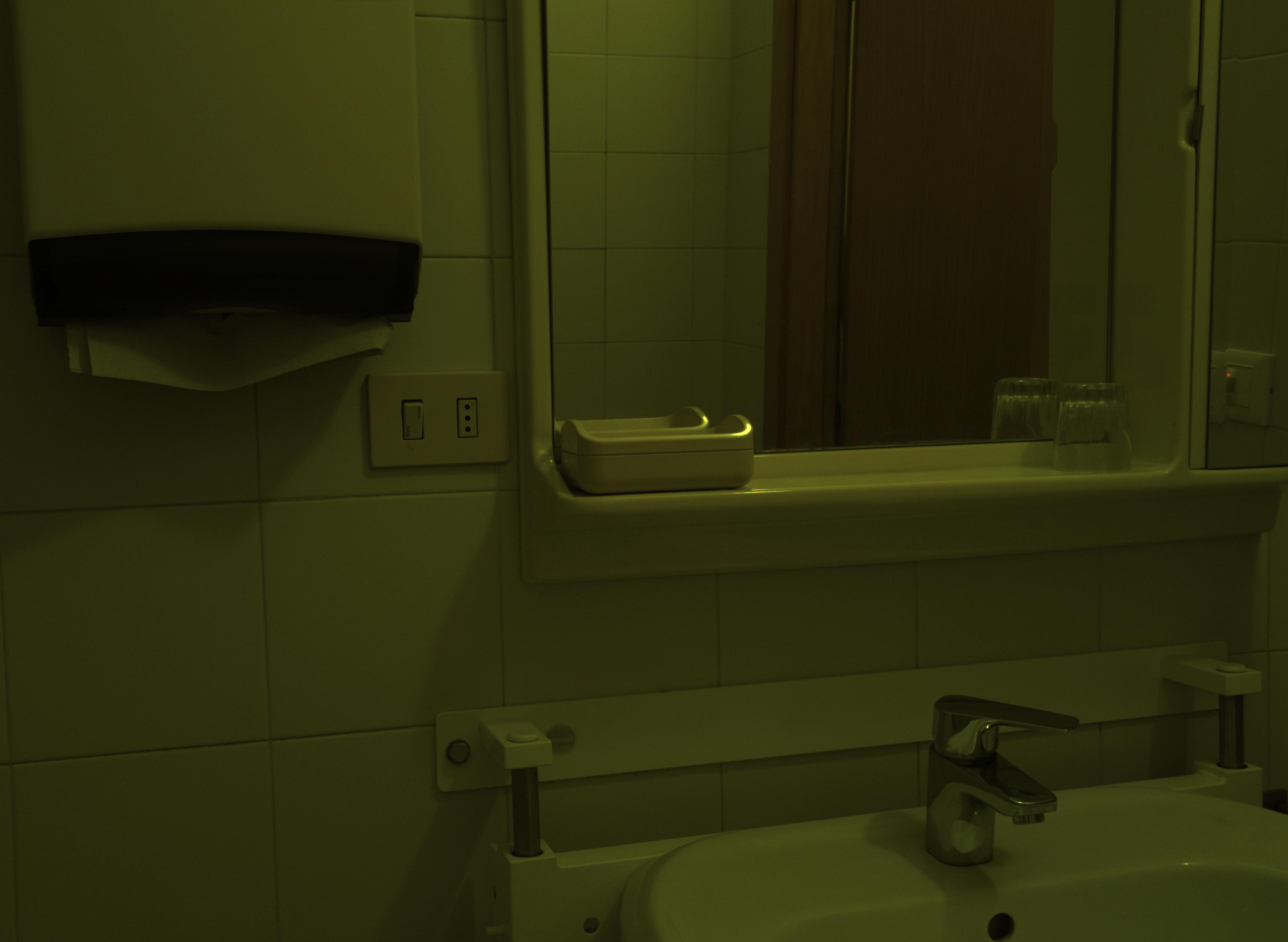 } & \includegraphics[width=0.125\linewidth]{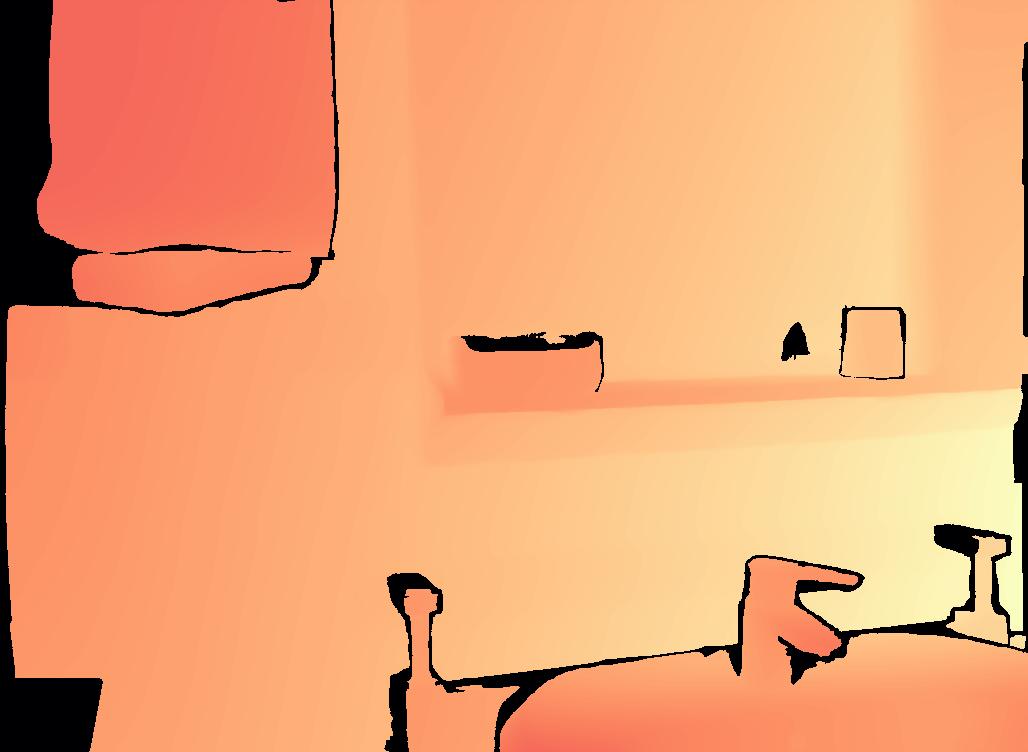 } & \includegraphics[width=0.125\linewidth]{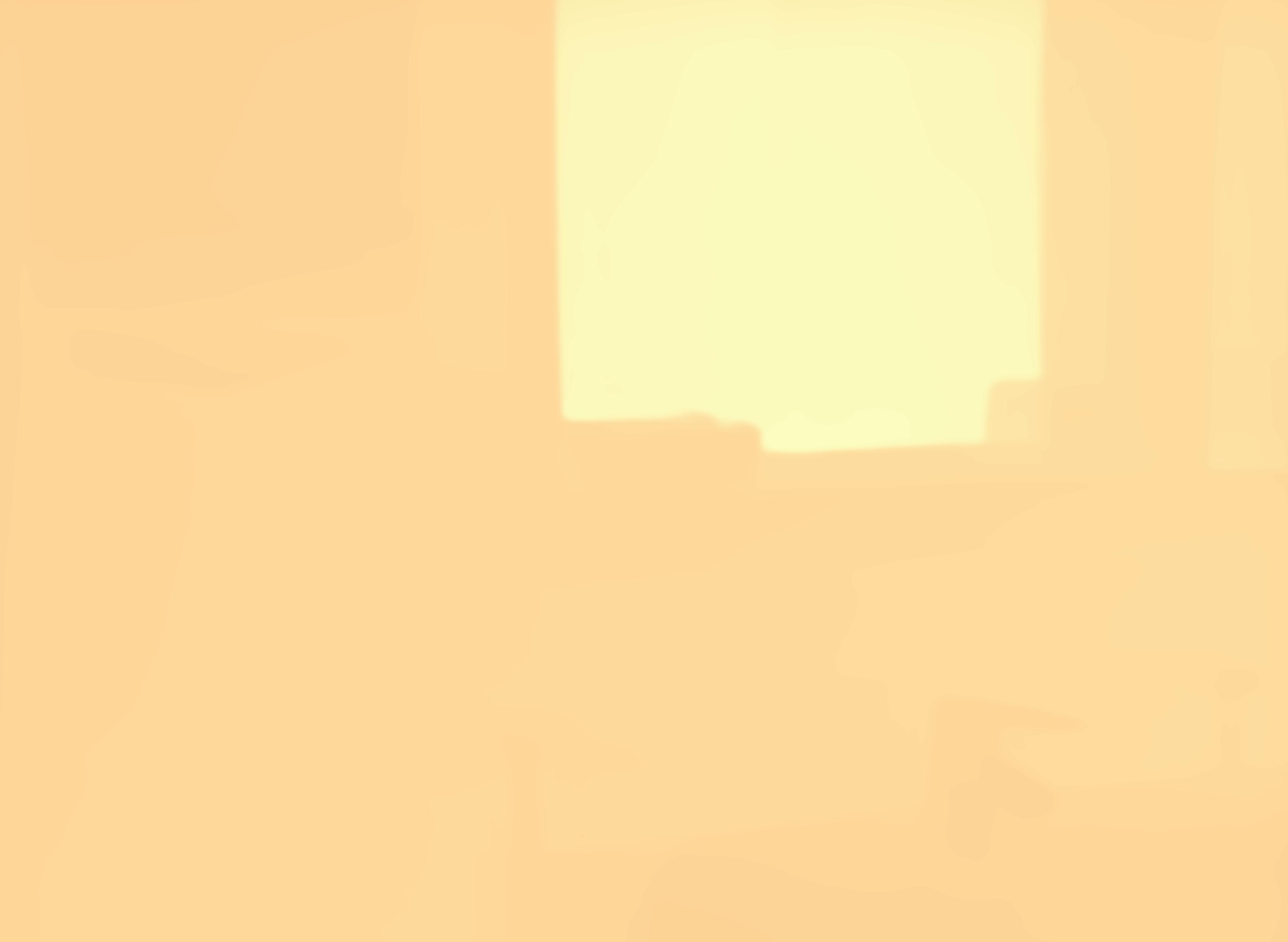 } & \includegraphics[width=0.125\linewidth]{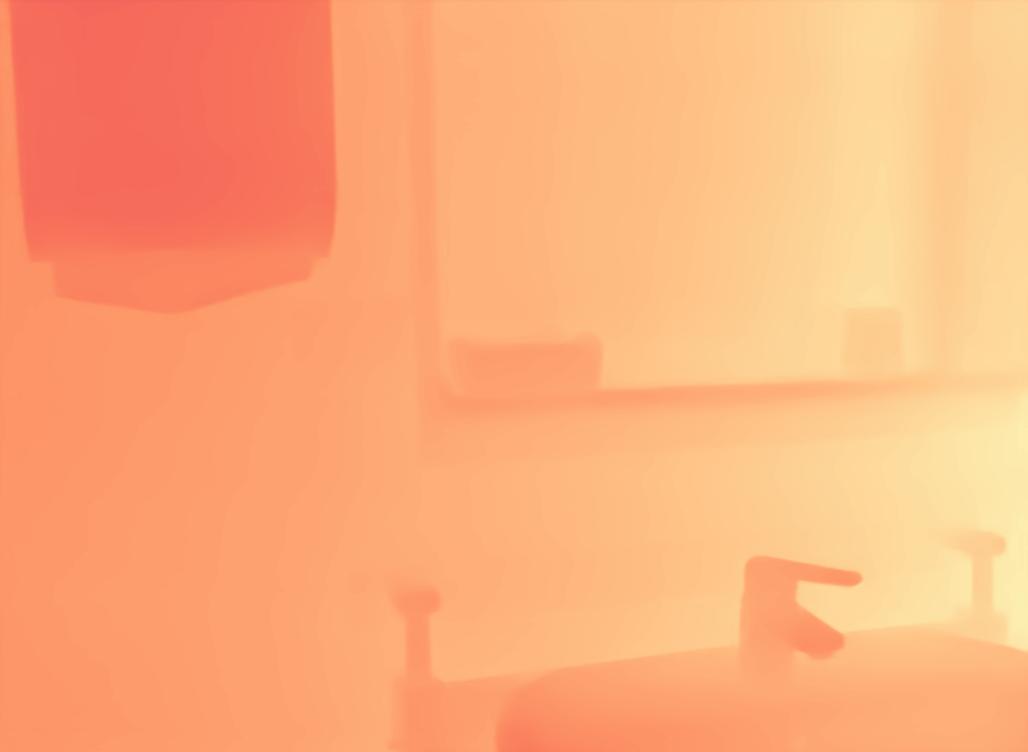} & \includegraphics[width=0.125\linewidth]{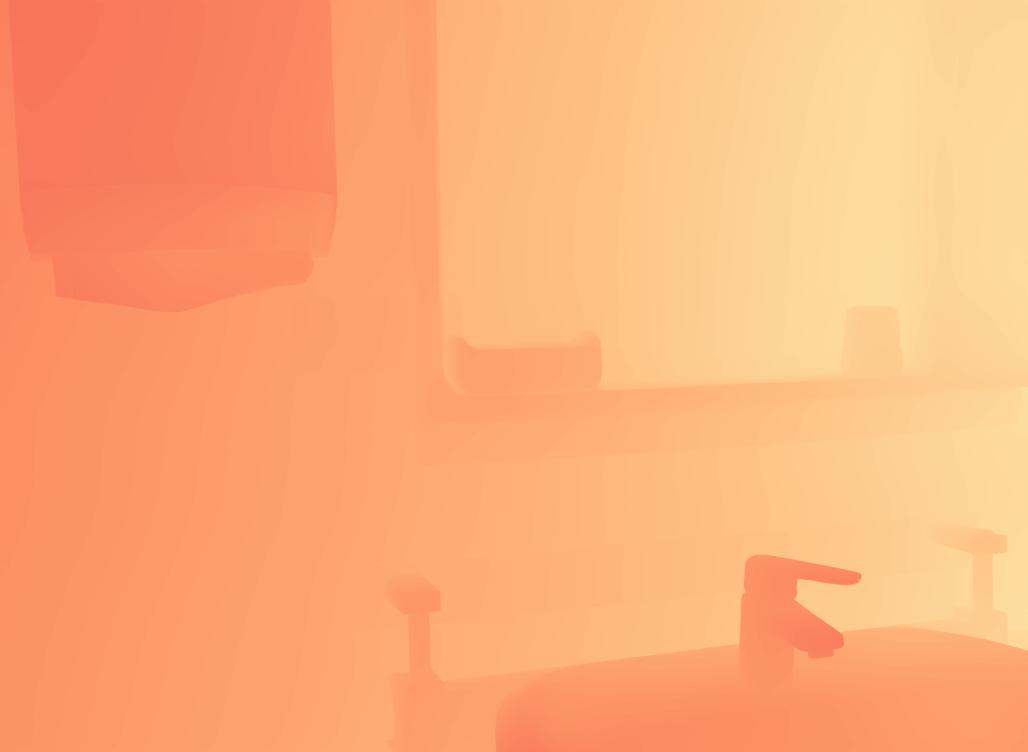} & \includegraphics[width=0.125\linewidth]{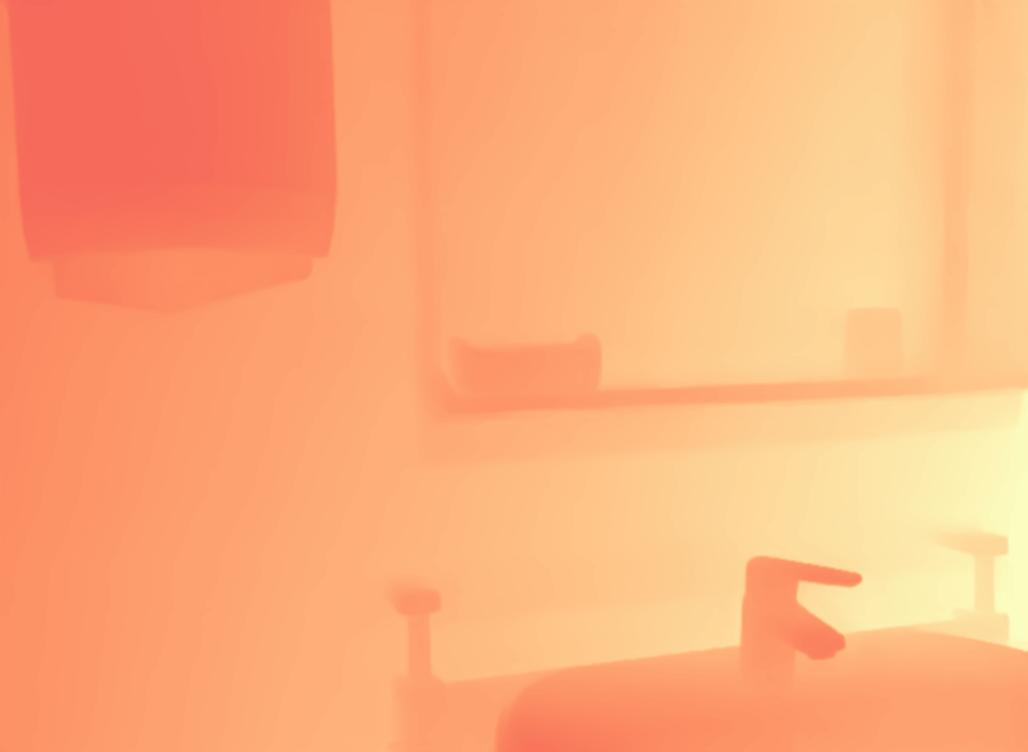} & \includegraphics[width=0.125\linewidth]{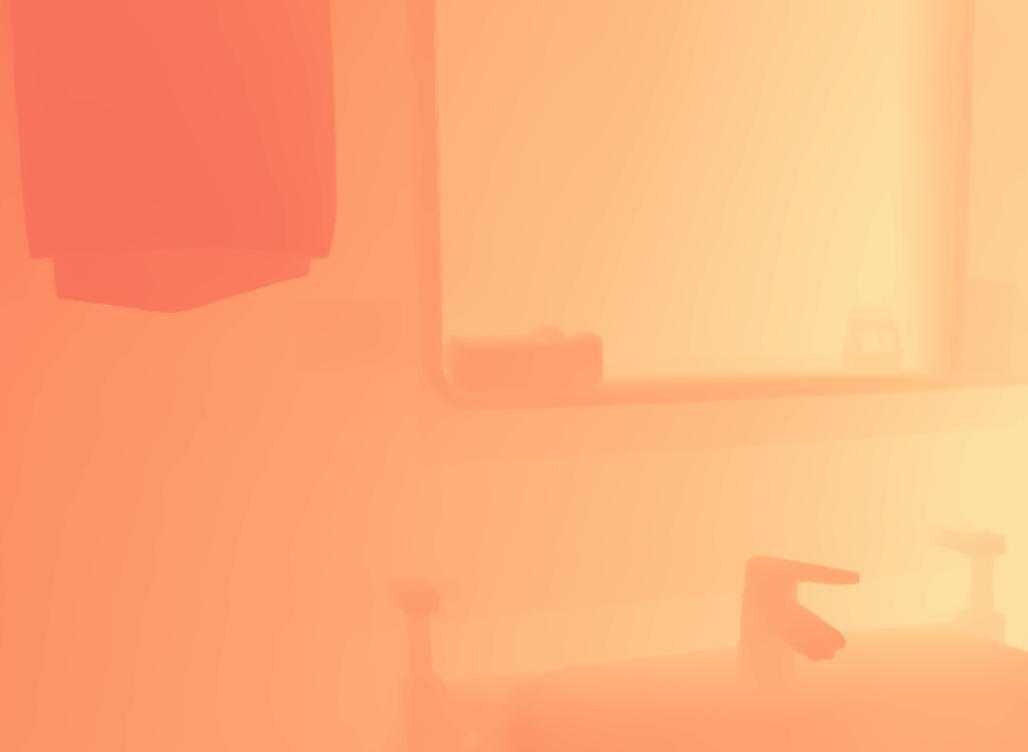} \\

\includegraphics[width=0.125\linewidth]{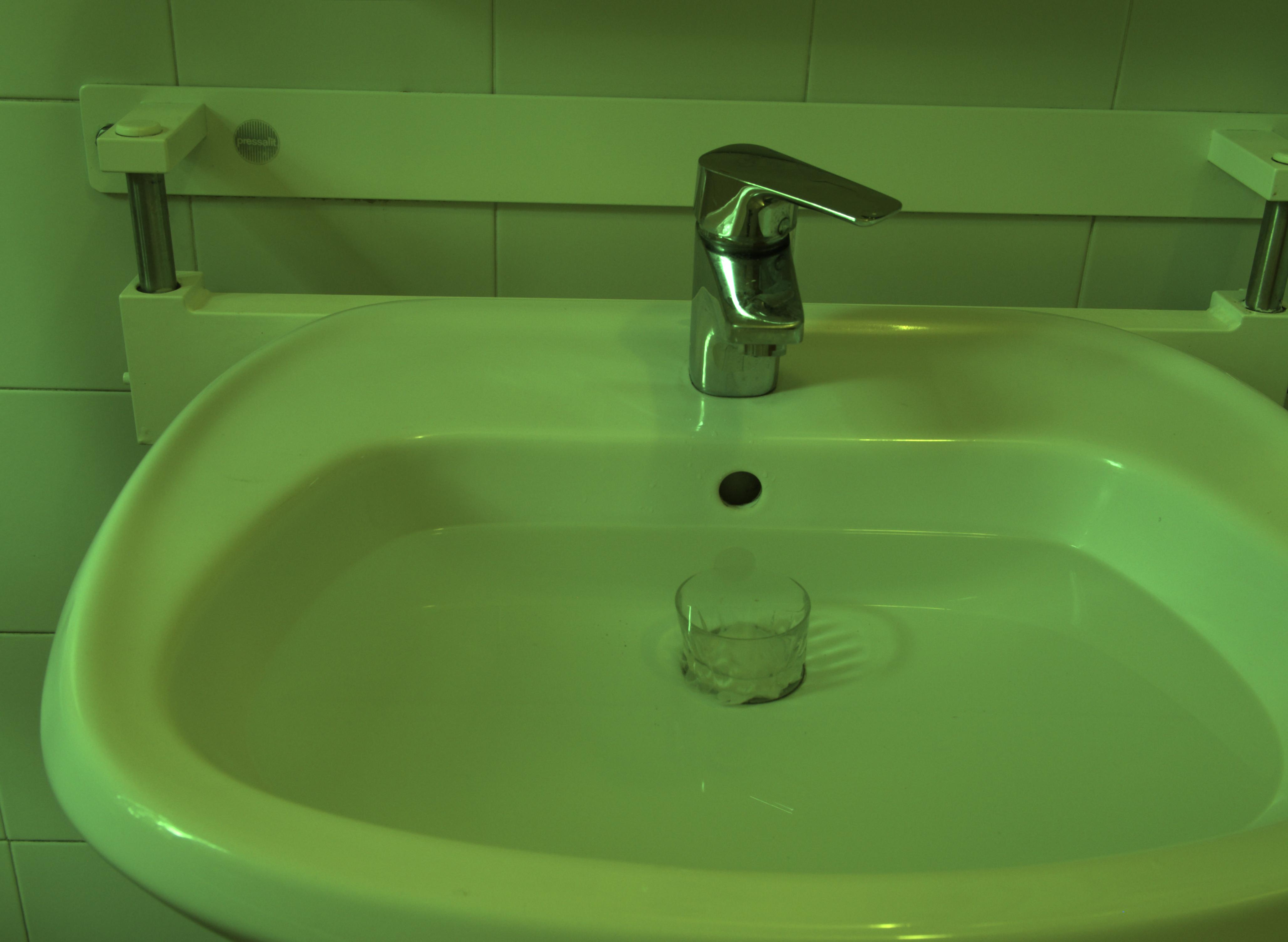 } & \includegraphics[width=0.125\linewidth]{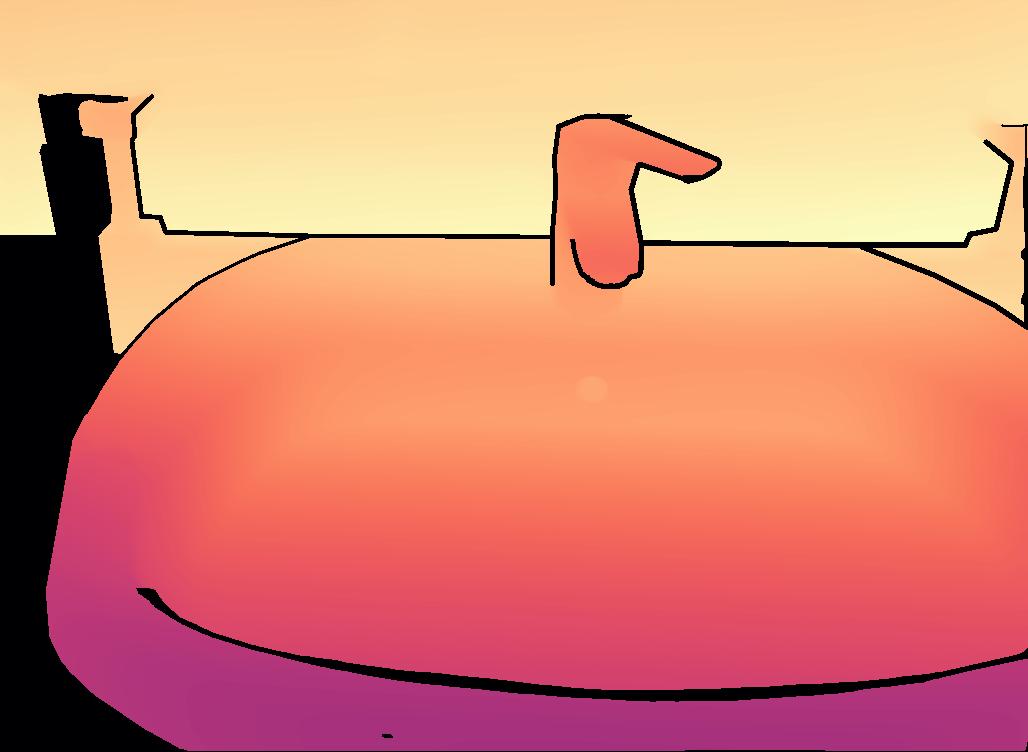 } & \includegraphics[width=0.125\linewidth]{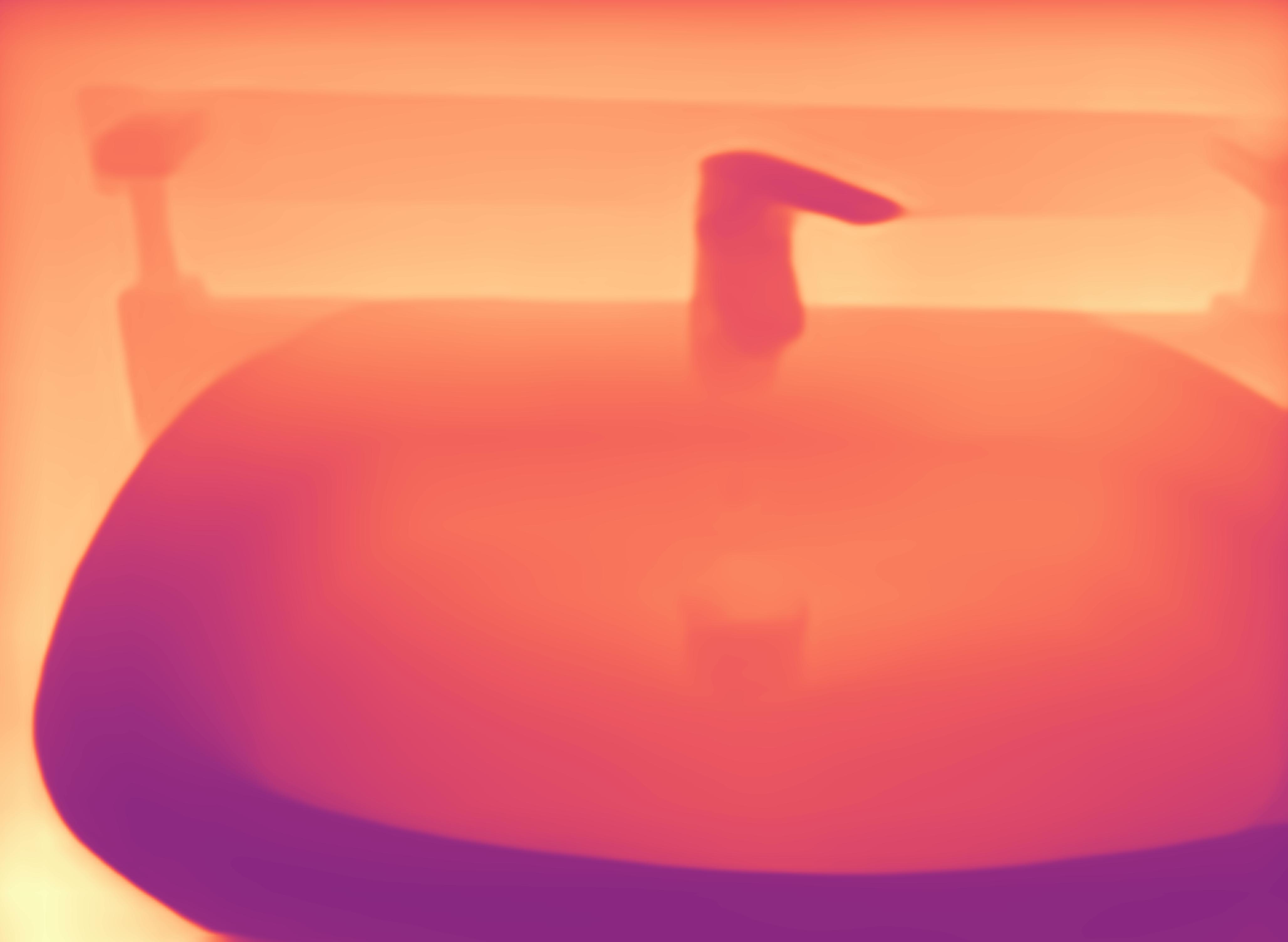 } & \includegraphics[width=0.125\linewidth]{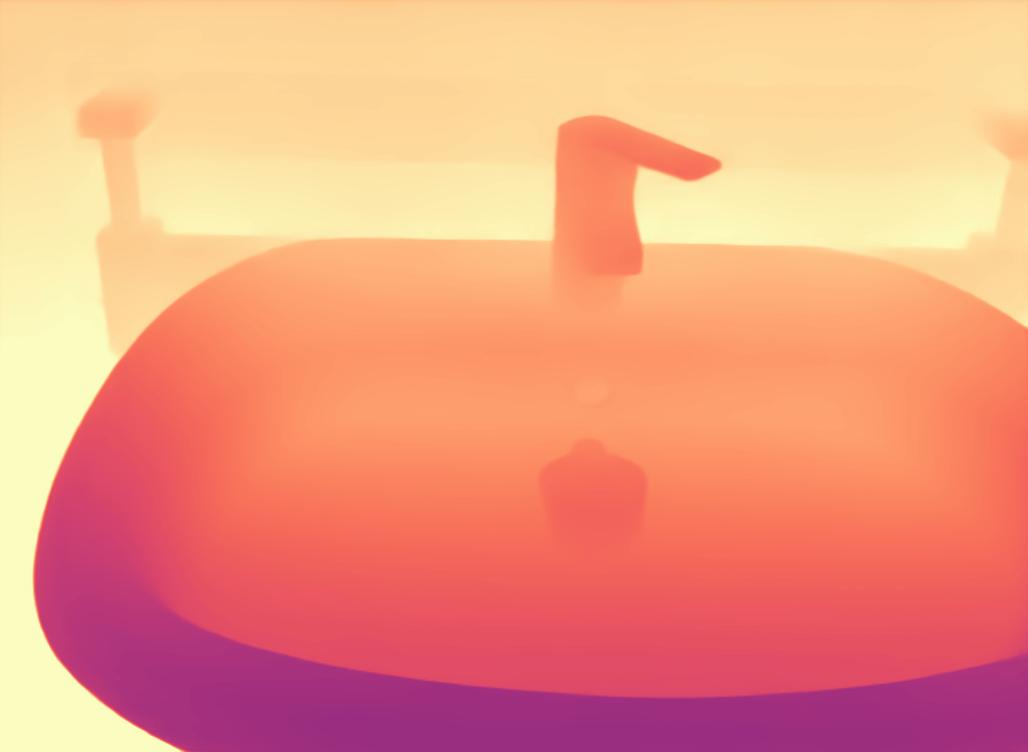} & \includegraphics[width=0.125\linewidth]{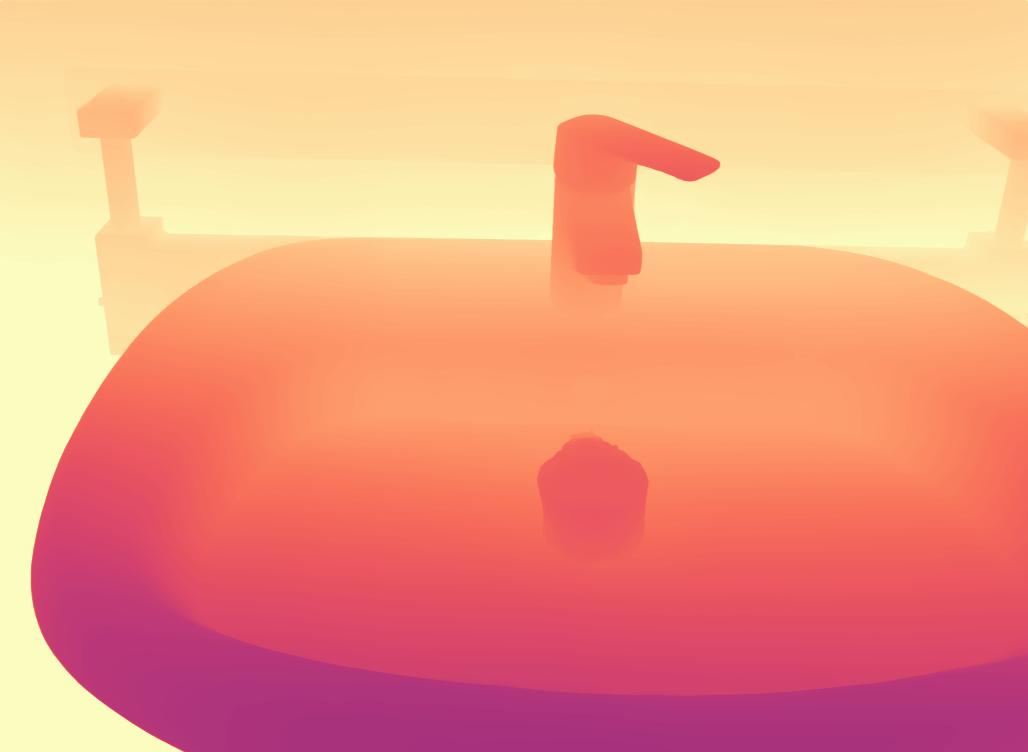} & \includegraphics[width=0.125\linewidth]{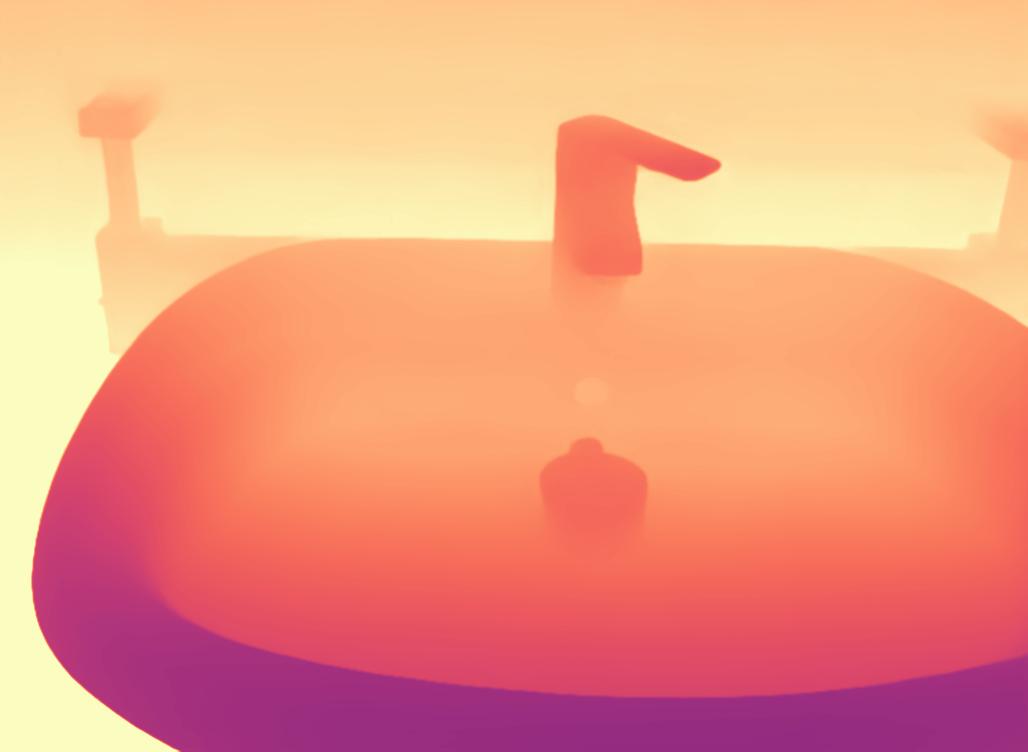} & \includegraphics[width=0.125\linewidth]{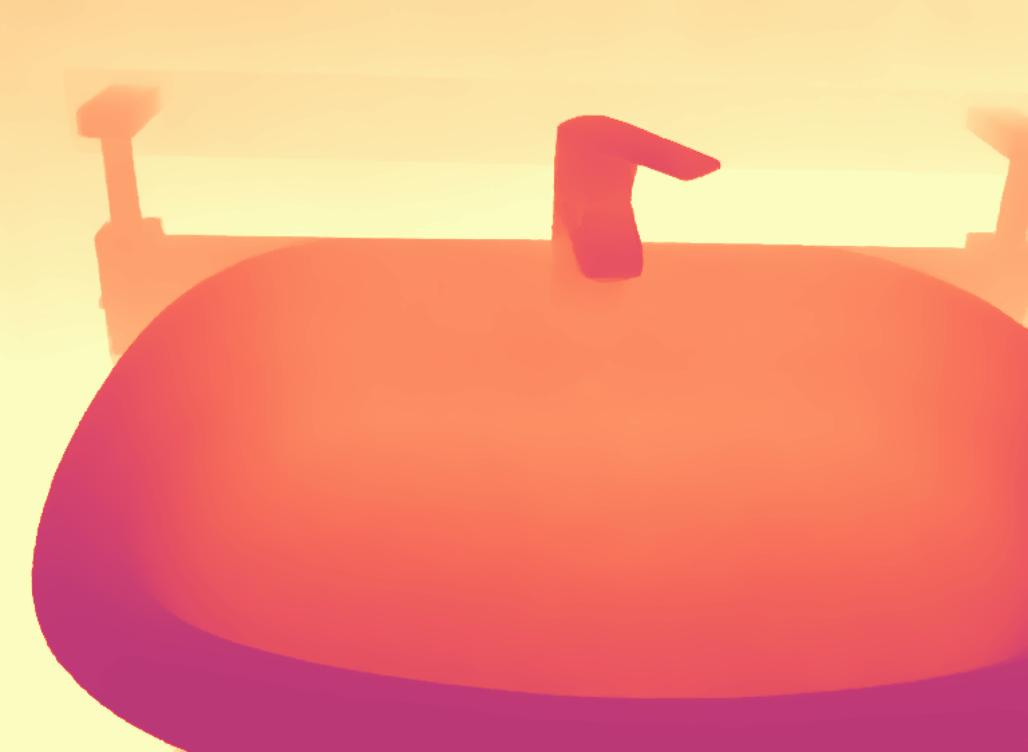}  \\

\includegraphics[width=0.125\linewidth]{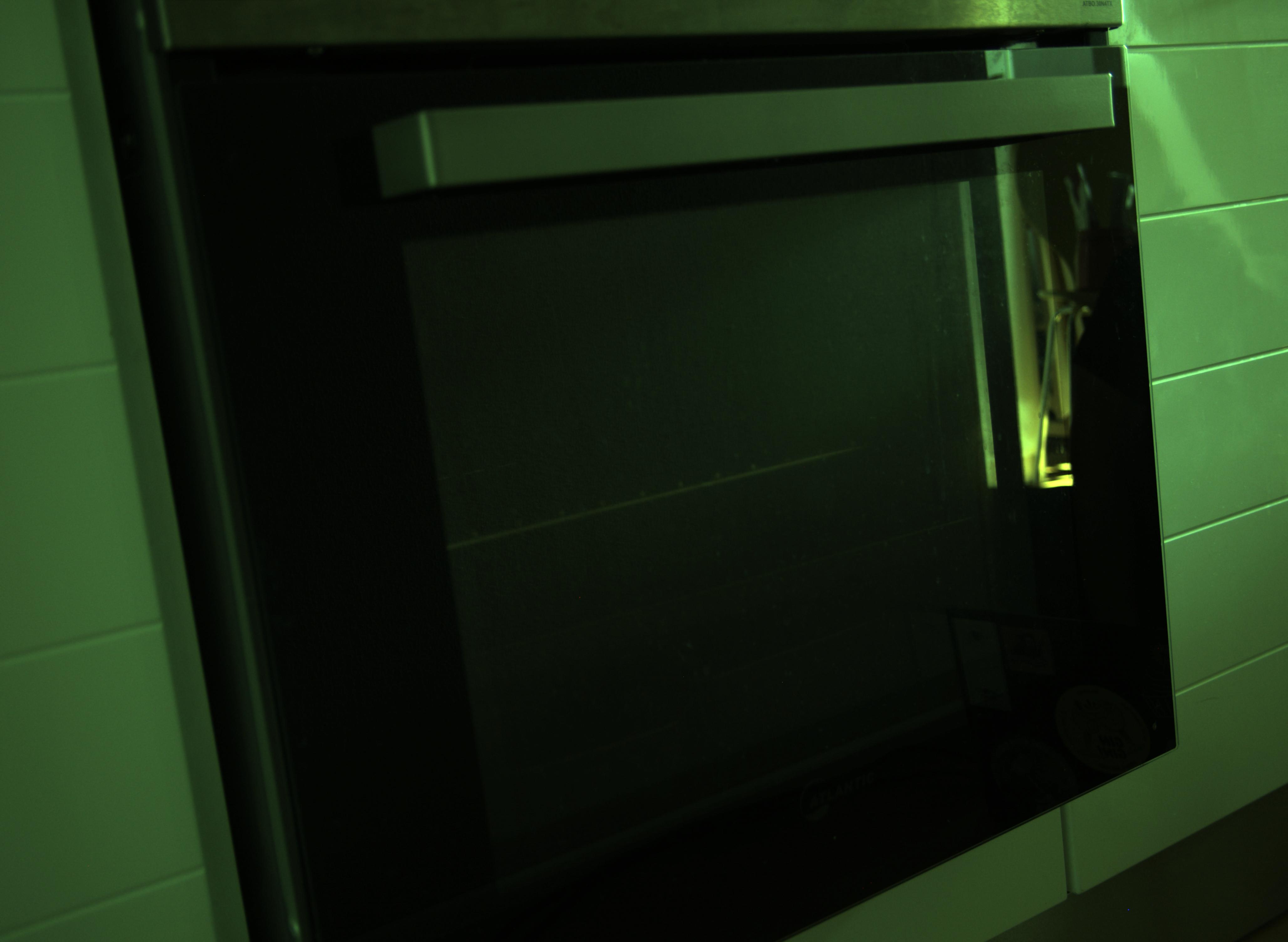 } & \includegraphics[width=0.125\linewidth]{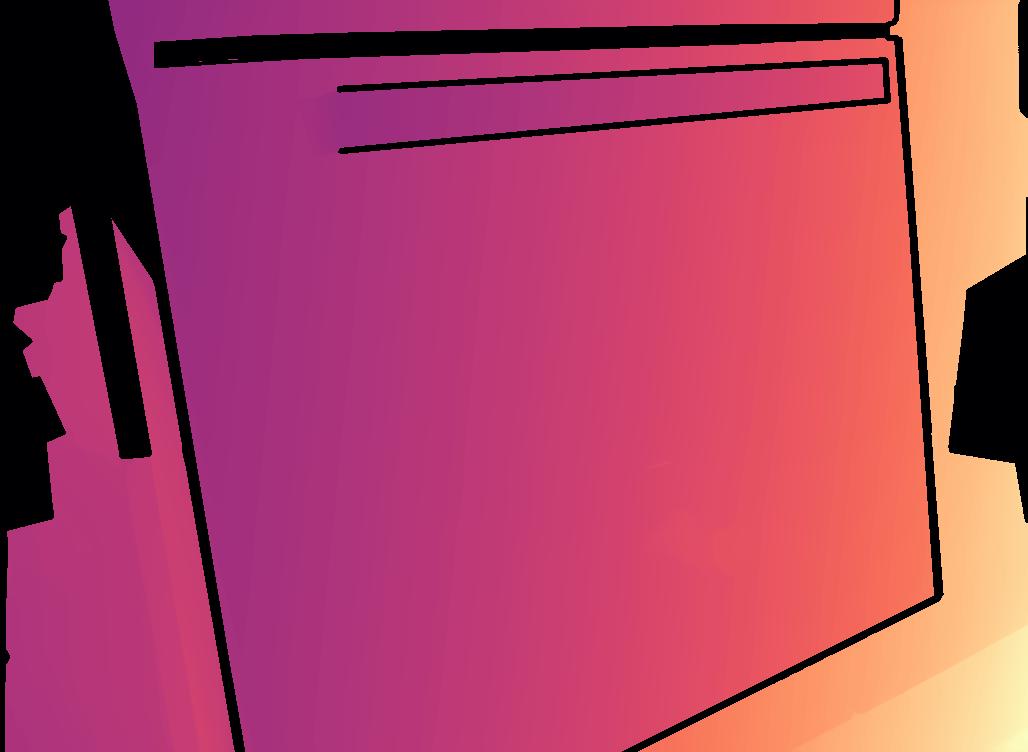 } & 
\includegraphics[width=0.125\linewidth]{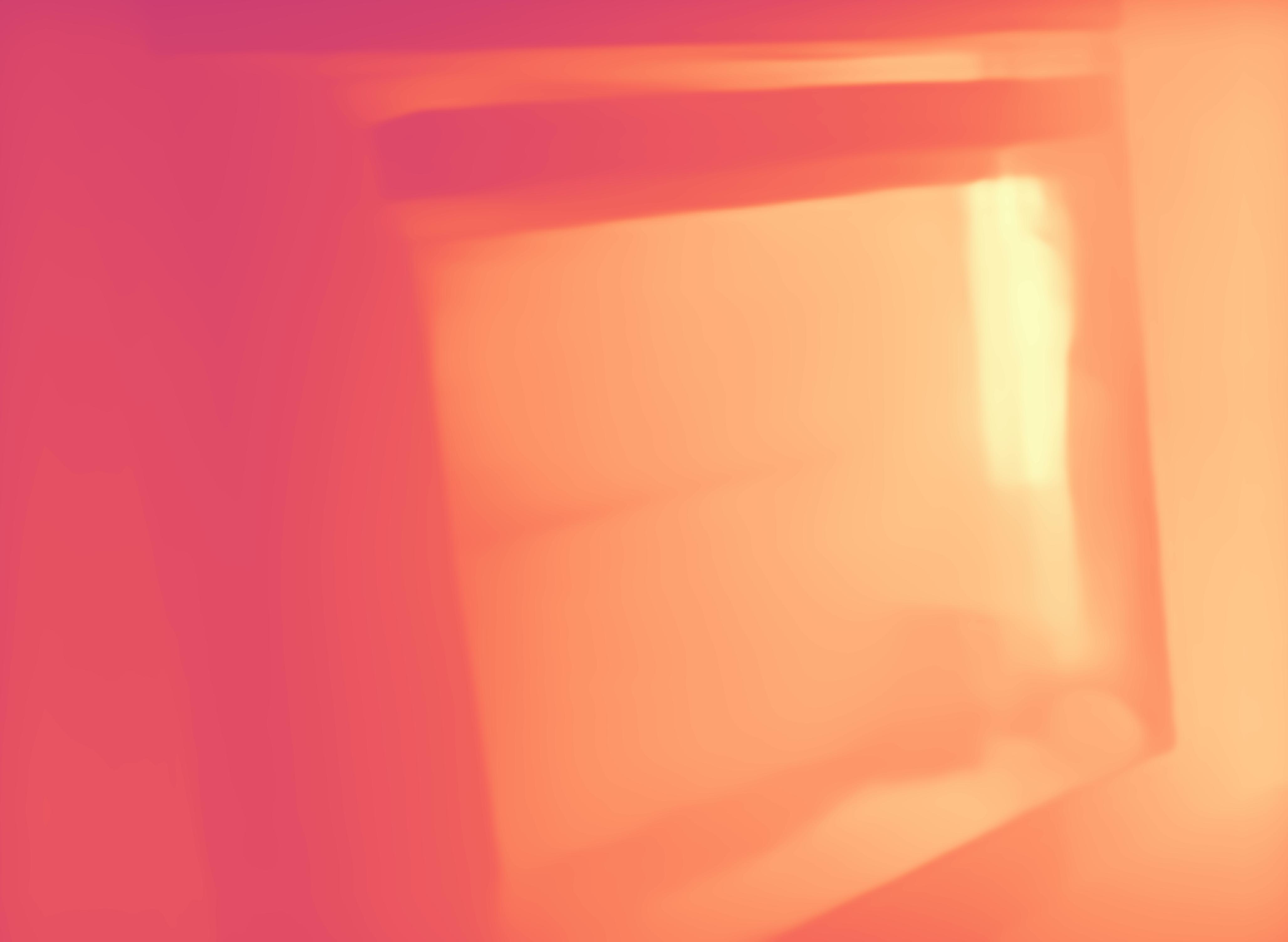 } & \includegraphics[width=0.125\linewidth]{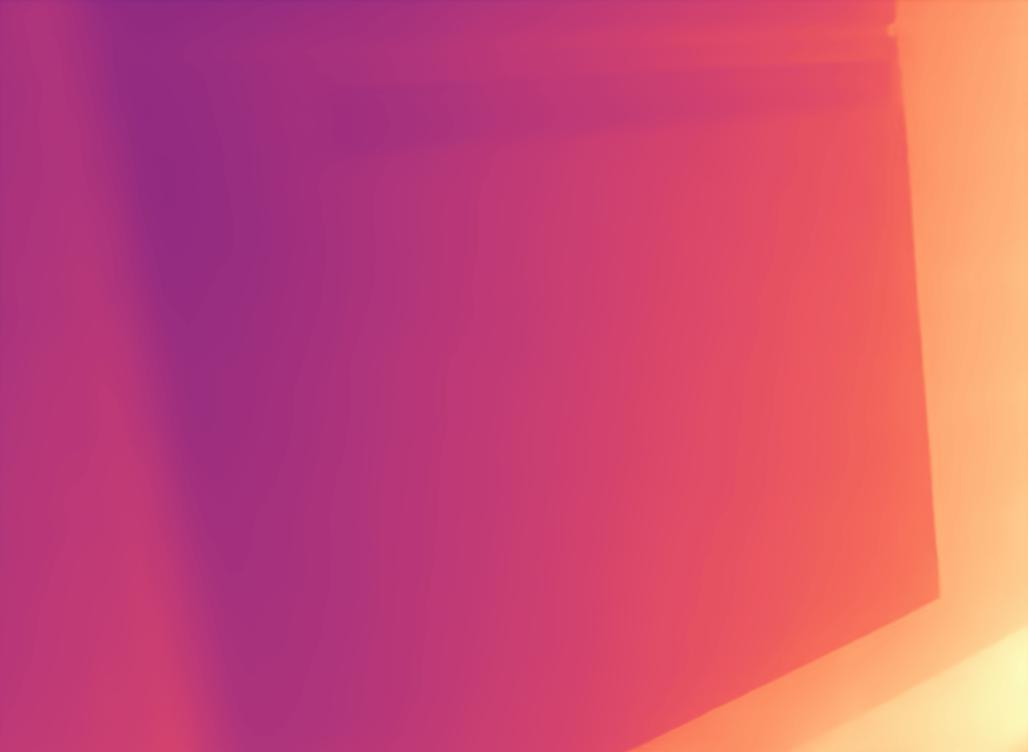} & \includegraphics[width=0.125\linewidth]{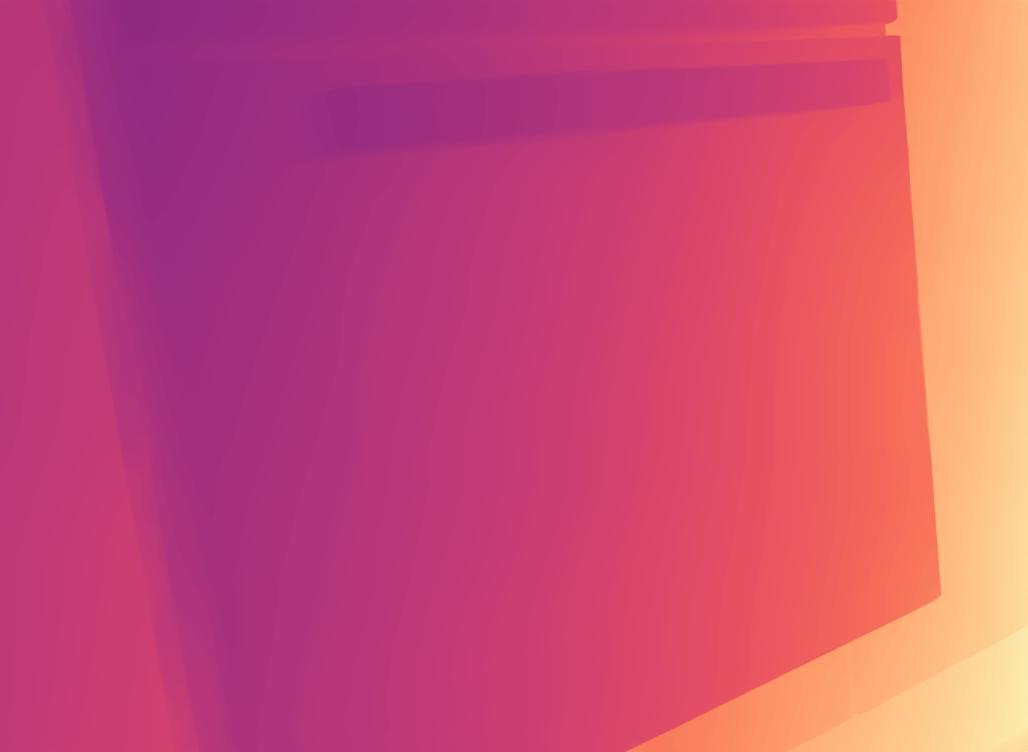} & \includegraphics[width=0.125\linewidth]{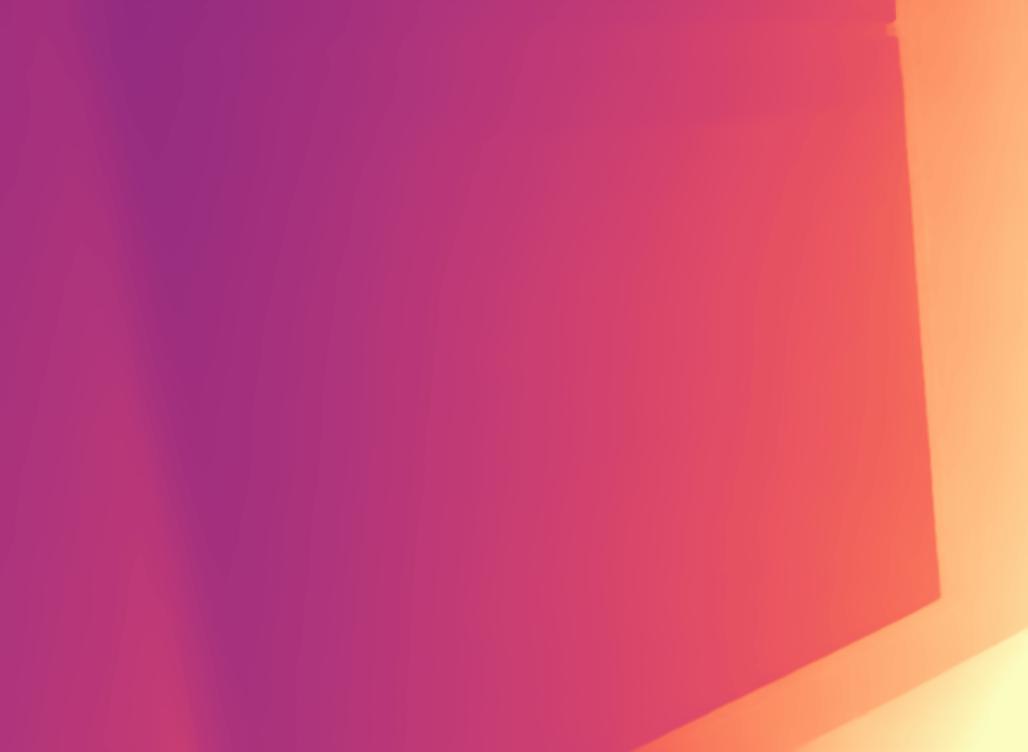} & \includegraphics[width=0.125\linewidth]{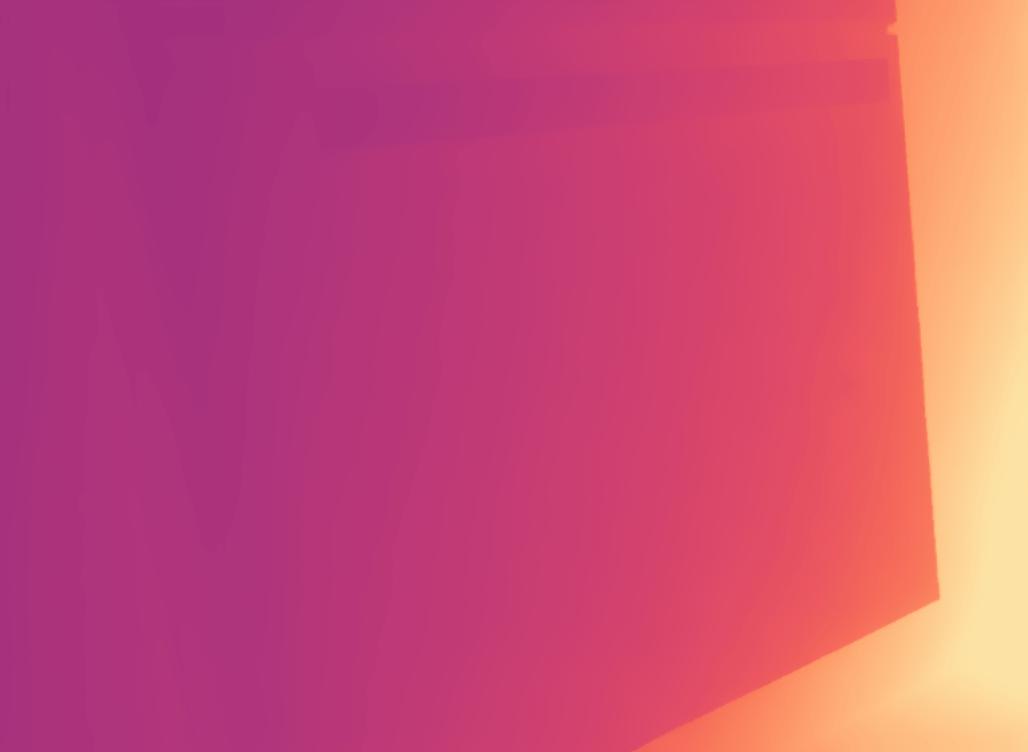}  \\

\includegraphics[width=0.125\linewidth]{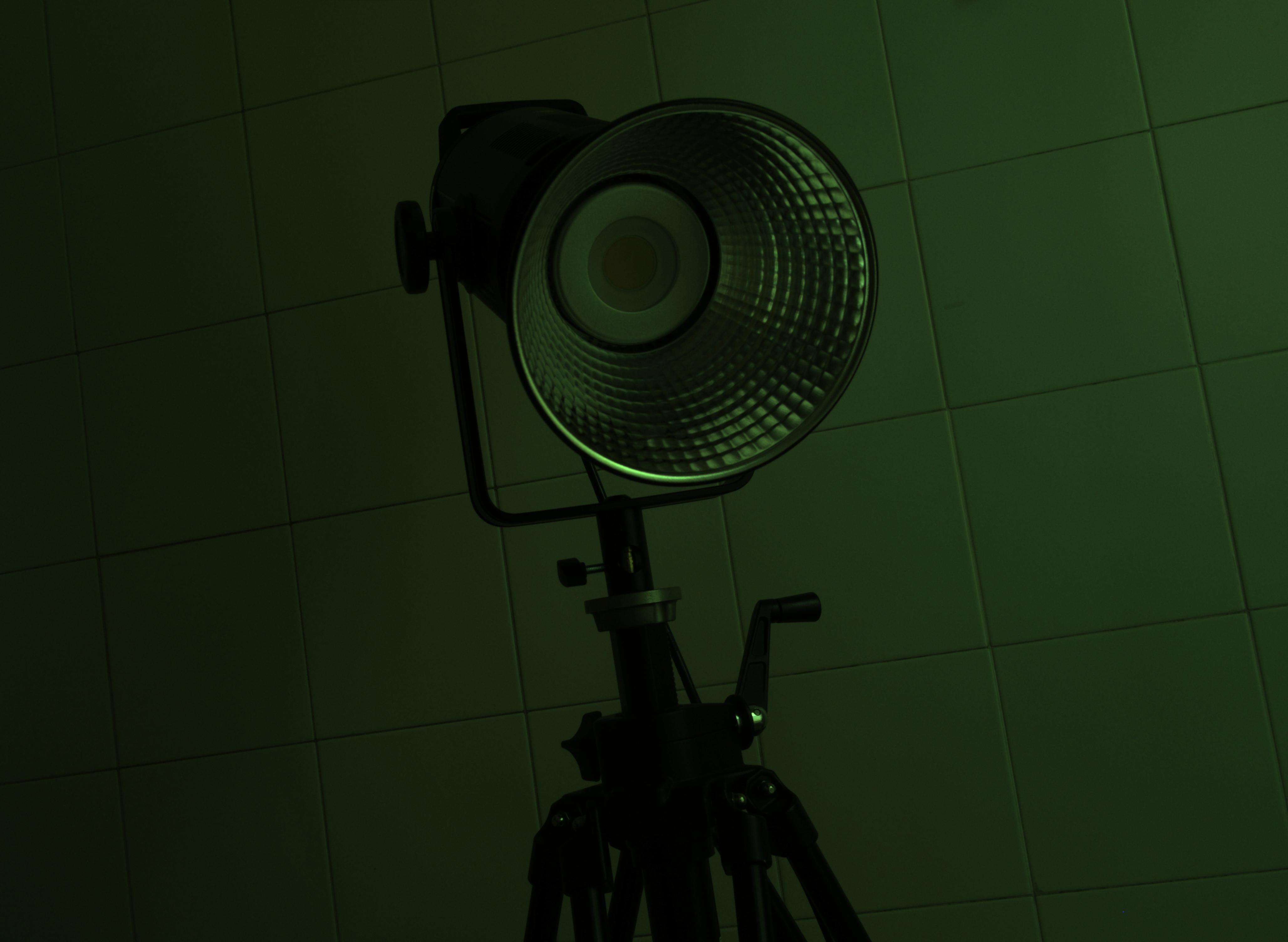 } & 
\includegraphics[width=0.125\linewidth]{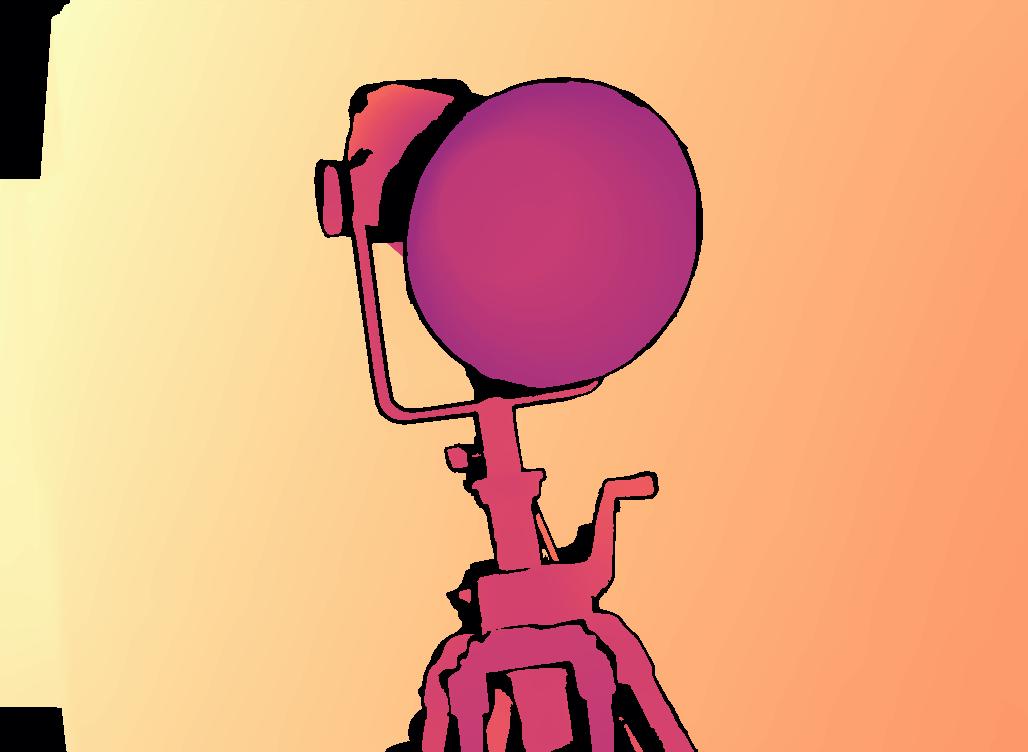 } & \includegraphics[width=0.125\linewidth]{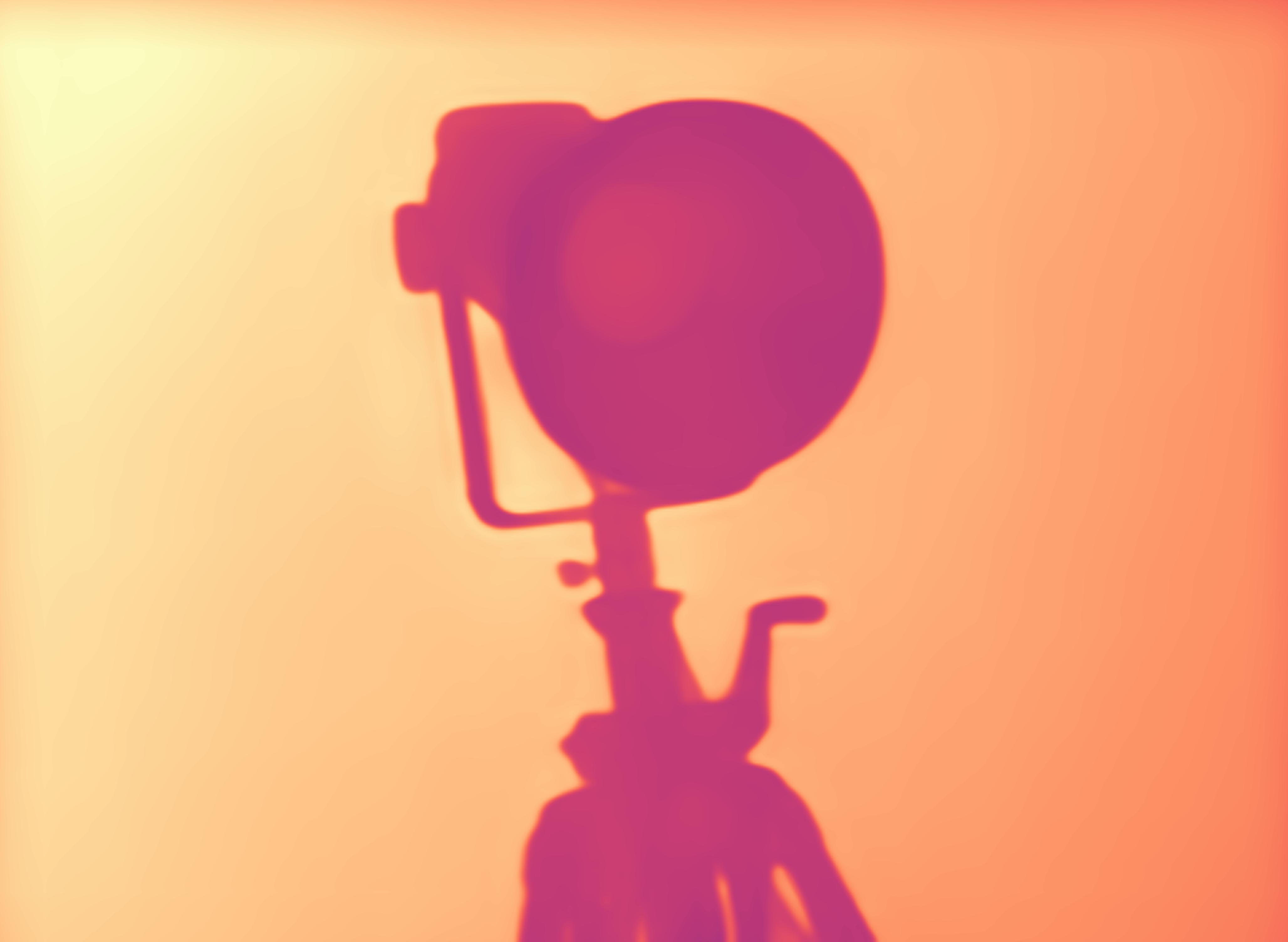 } & \includegraphics[width=0.125\linewidth]{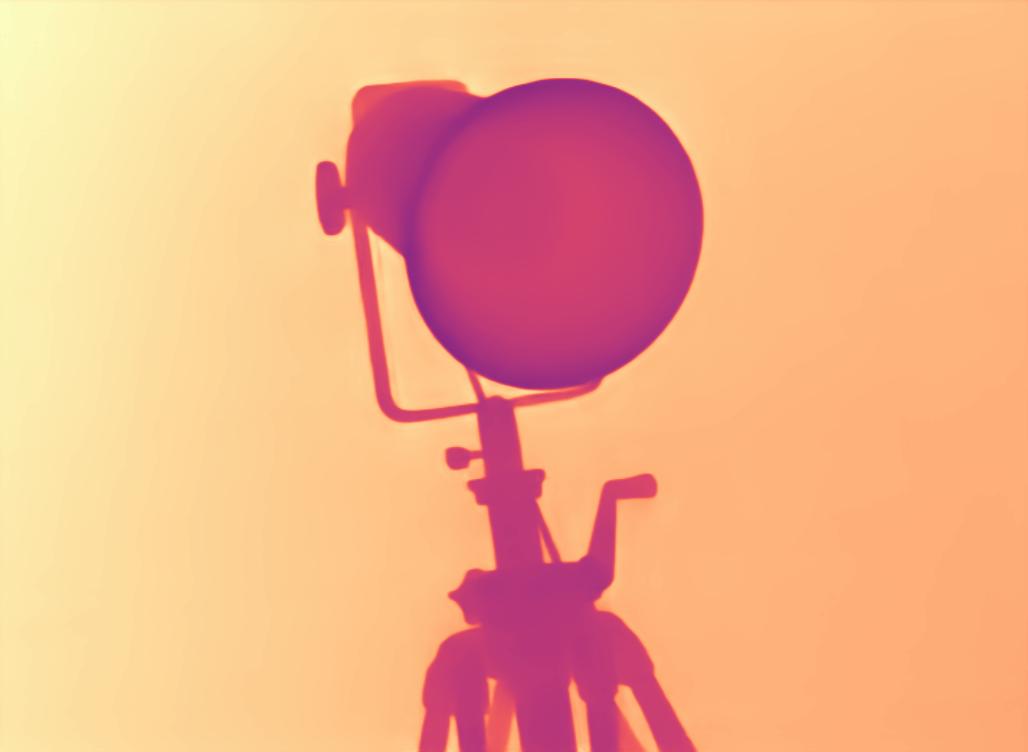} & \includegraphics[width=0.125\linewidth]{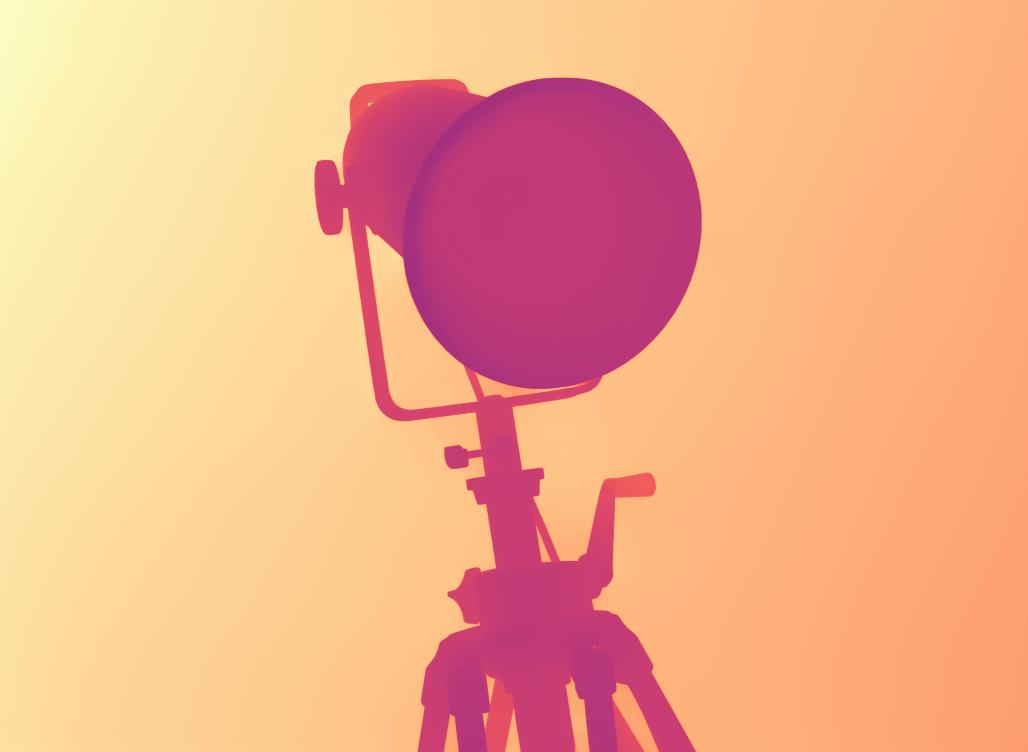} & \includegraphics[width=0.125\linewidth]{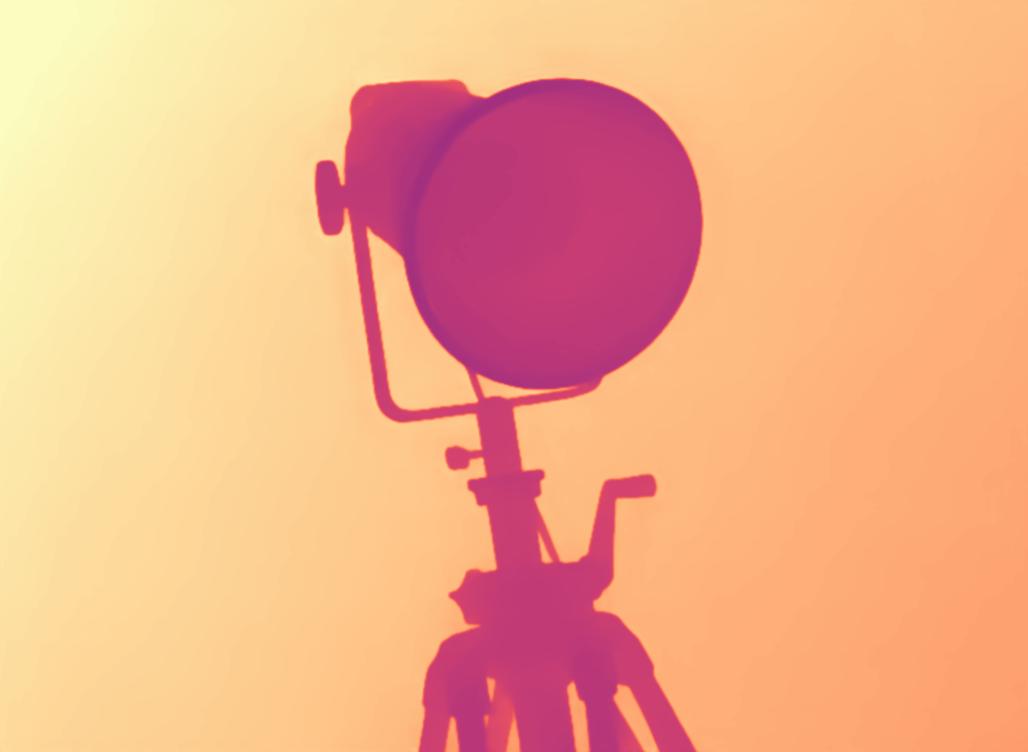} & \includegraphics[width=0.125\linewidth]{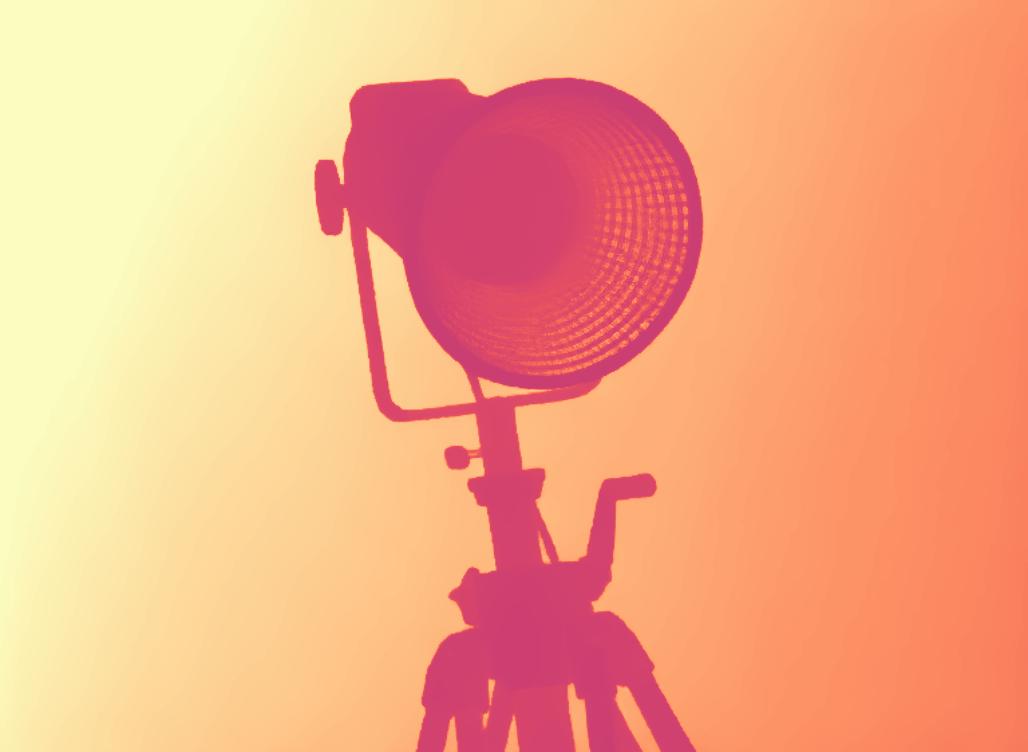}  \\

\end{tabular}\vspace{-0.3cm}
\caption{\textbf{Qualitative results -- Mono track.} From left to right: RGB reference image, ground-truth disparity, predictions by ZoeDepth \cite{bhat2023zoedepth}, Lavreniuk, colab, PreRdW, and IPCV.}\vspace{-0.3cm}
\label{fig:mono}
\end{figure*}

\subsection{Track 2: Mono}\label{sec:results_mono}

Table \ref{tab:mono} shows the results for the second track. At the very bottom, we report the results achieved by the baseline method -- i.e., the ZoeDepth \cite{bhat2023zoedepth} model using the weights provided by the authors. 
From left to right, we report deltas, Abs Rel., MAE, and RMSE metrics for \textit{Tom}, \textit{All}, and \textit{Other} pixels respectively. We report two different rankings, according to the performance observed on $\delta<1.05$ -- the most restrictive metric -- computed over \textit{ToM} and \textit{All} pixels.

Unlike the stereo track, all of the submitted methods consistently outperformed the ZoeDepth baseline, with \textbf{Lavreniuk} achieving the best accuracy values on \textit{ToM}, \textit{All} and \textit{Other} pixels. 
Indeed, conversely to what was observed in the stereo track, \textbf{Lavreniuk} represents the most versatile method being the top performer on all the pixel categories.
For what concerns \textit{ToM} regions, the top \#3 methods are able to push the strictest accuracy metric -- $\delta<1.05$ -- beyond 85\%, with a remarkable 15\% improvement with respect to last year, as well as to reduce the Abs Rel. below 3\%.
The improvements are reflected on \textit{All} and \textit{Other} pixels as well. Despite the minor gain with respect to the baseline, compared to what was observed on \textit{ToM} regions, the improvement is yet consistent.

Fig. \ref{fig:mono} shows some qualitative examples from the mono testing set. 
Similarly to last edition, any of the submitted models can properly handle \textit{ToM} regions, such as for the oven in the third row, while still struggling on mirrors or water surfaces, as in first and second rows. 

\section{Challenge Methods}

\subsection{Track 1: Stereo}\label{sec:description_stereo}

\subsubsection{Baseline - CREStereo \cite{li2022practical}}

For the first track, we set the state-of-the-art CREStereo architecture \cite{li2022practical} as our baseline. This model consists of a hierarchical network employing a recurrent refinement process, designed to update the predicted disparity map in coarse-to-fine manner. This process is implemented based on an adaptive group correlation layer (AGCL), where an alternate 2D-1D local search strategy with deformable windows is employed for robust matching even in the presence of imperfect rectification. The AGCL module computes correlations between pixels in local search windows, in contrast to what the all-pairs correlation module from RAFT-Stereo \cite{lipson2021raft} does, reducing the computational requirements.
To obtain the final predictions, we process images at quarter resolution using the original weights released by the authors, then we upsample predicted disparity maps to the original resolution through bilinear interpolation.

\subsubsection{Team 1 - NJUST-KMG}

The NJUST-KMG team (CodaLab: \textit{chenyin}) adapts DEFOM-Stereo \cite{jiang2025defom} by integrating Depth Anything V2. Their approach improves CNN encoders with a depth foundation model, introduces a scale update module before the delta update module, and leverages reflective data with multi-scale sampling for training.
The feature encoder fuses DPT and CNN feature maps at 1/4 resolution for matching, while the context encoder combines multi-level DPT and CNN features.

Disparity estimation is initialized with Depth Anything V2's depth, followed by scale correction and detail refinement using pyramid lookup. The system is pre-trained on KITTI \cite{Menze2015CVPR}, Middlebury \cite{scharstein2014high}, and ETH3D \cite{schops2017multi}, then fine-tuned on Booster \cite{booster} and CREStereo \cite{li2022practical} datasets.  Quarter-resolution inference ensures efficient processing of high-resolution inputs.

\begin{figure}[t]
    \centering
    \includegraphics[width=0.8\linewidth]{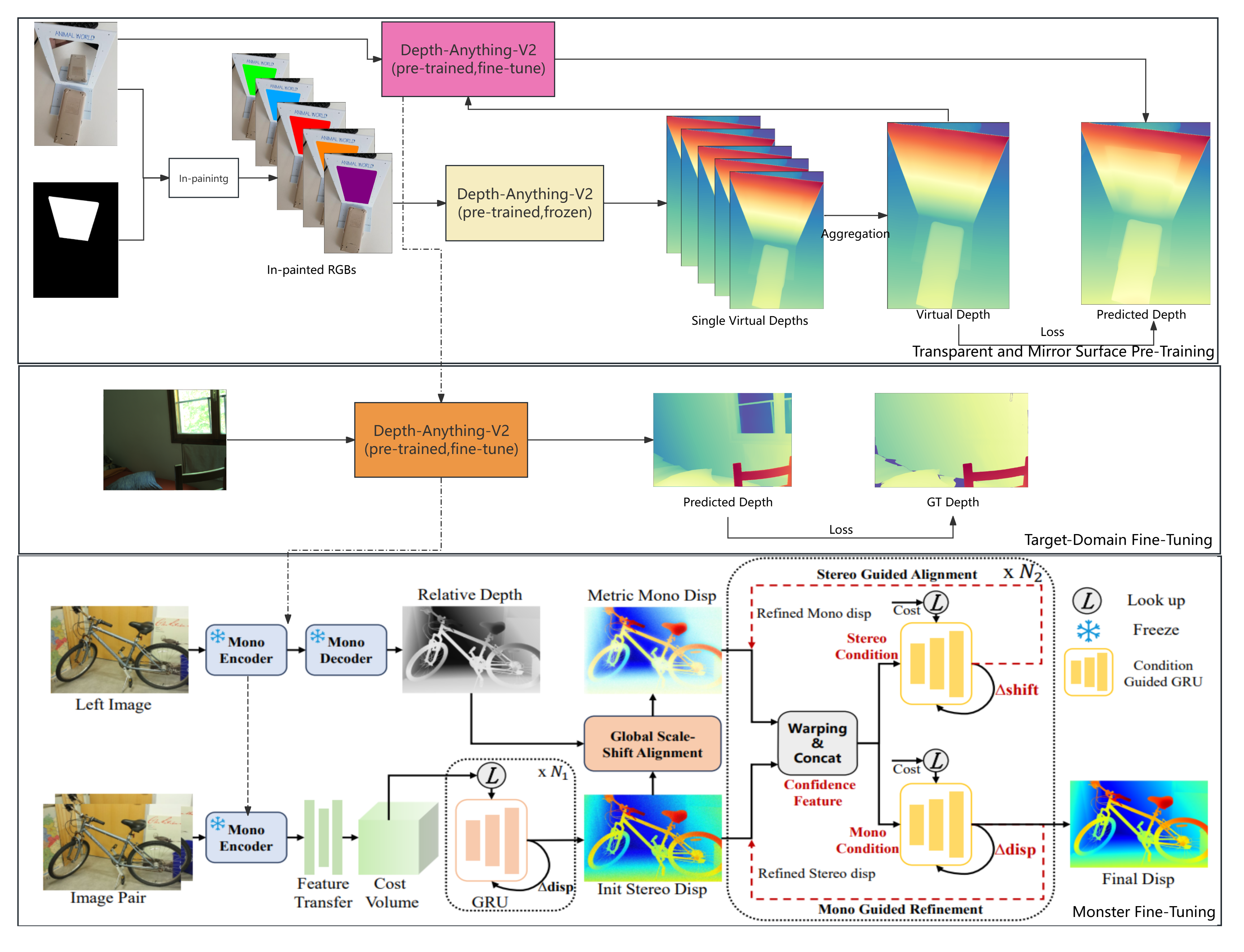}
    \vspace{-0.3cm}
    \caption{\textbf{Network Architecture -- Team \textit{Robot01-vRobotit}.}}\vspace{-0.4cm}
    \label{fig:robot01}
\end{figure}

\subsubsection{Team 2 - Robot01-vRobotit}

The Robot01-vRobotit team (CodaLab: \textit{Bupt-chenwu}) introduces an improved MonSter \cite{cheng2025monster} based on Depth Anything V2. Their approach leverages the complementary strengths of monocular depth estimation and stereo matching in a dual-branch architecture, where monocular depth provides global structural information while stereo matching refines pixel-level geometric details.

The training process is divided into two stages: first, the Depth Anything V2-Large model is fine-tuned to improve its accuracy in TOM regions of the Booster dataset; second, the MonSter network is fine-tuned using the improved monocular model from the first stage. During training, extensive data augmentation is applied, including resizing, random cropping, saturation adjustment, color jittering, and spatial scaling. The fine-tuning is performed on NVIDIA RTX 4090 GPUs for 1000 epochs, with a learning rate of $1 \times 10^{-4}$ and batch size of 12.

\subsubsection{Team 3 - SRC-B [Stereo]}

The SRC-B team (CodaLab: \textit{pixinsight}) presents "Multi-Scale-Mono-Stereo," a method that integrates monocular depth network features to improve depth estimation in high-resolution images with non-Lambertian surfaces. The approach adopts a data augmentation strategy similar to ASGrasp \cite{shi2024asgrasp}, leveraging Blender to generate additional stereo training samples from the AI2THOR \cite{kolve2017ai2} dataset.
To effectively use the pre-trained monocular model, they adopt the stereo branch structure from MonSter \cite{cheng2025monster}, incorporating the pretrained ViT encoder of Depth Anything V2 with frozen parameters. A feature transfer network is introduced to downsample and transform the ViT-extracted features into a multi-scale feature pyramid, which is then concatenated with features extracted by IGEV \cite{xu2023iterative}.
To enhance performance on high-resolution images, the team integrates a stacked cascaded architecture during training, enabling the network to adaptively propagate depth information across different scales. The network is implemented in PyTorch and trained on NVIDIA RTX 3090 GPUs, with the Depth Anything V2 module's weights kept fixed while fine-tuning the stereo module for an additional 100,000 steps.

\begin{figure}[t]
    \centering
    \includegraphics[width=0.8\linewidth]{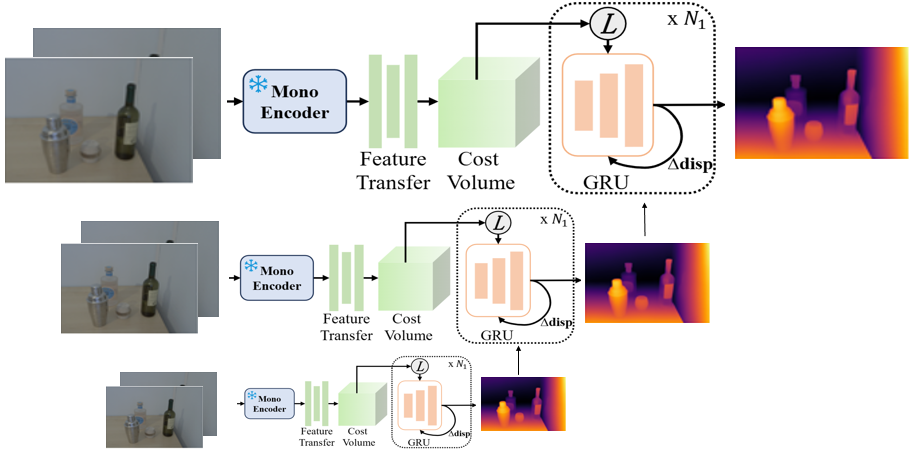}
    \vspace{-0.3cm}
    \caption{\textbf{Network Architecture -- Team \textit{SRC-B [Stereo]}.}}\vspace{-0.4cm}
    \label{fig:src_b_stereo}
\end{figure}

\subsubsection{Team 4 - weouibaguette}

The weouibaguette team employs FoundationStereo (FS) \cite{wen2025stereo}, finding that the original network outperforms all custom-trained models they developed. Using the Booster training set, they evaluated FS under various parameter configurations and observed, similar to last year's NTIRE winner (MiMcAlgo), that inference yielded better overall metrics on down-sampled images.

The team experimented with different down-sampling factors (0.5, 0.3, 0.25, and 0.1) and tested inference with and without hierarchical processing. The output of FS was resized using linear interpolation. Their results indicate that optimal performance was achieved with a resizing factor of 0.2 and hierarchical inference enabled.

\subsection{Track 2: Mono}\label{sec:description_mono}

\subsubsection{Baseline - ZoeDepth \cite{bhat2023zoedepth} }

For this second track, we set the ZoeDepth model as our baseline, a state-of-the-art framework for single-image depth estimation. It builds over the DPT backbone \cite{Ranftl_2021_ICCV}, enhanced through a metric bins module implemented for learning metric depth rather than an affine-invariant output.
As for the Stereo track, we obtain the predicted depth maps by using the original weights made available by the authors. 

\subsubsection{Team 1 - Lavreniuk}
The Lavreniuk team's ``DeepBlend" method uses the Depth Anything v2 (DAv2) model as the primary depth estimator, with additional modifications to improve accuracy in challenging scenarios. For training, images are categorized into two groups: transparent objects and mirrors. For both categories, pseudo-labeling depth masks are created using a refined blending technique that fuses the original image with a transparent object mask, improving upon \cite{costanzino2023iccv}. For mirror surfaces, an additional restoration step is applied, where inpainted images generated using a fast Fourier convolution-based model serve as an auxiliary input for depth estimation.
Experiments with various depth estimation models led to selecting Depth Anything \cite{depthanything} and Depth Anything v2 for pseudo-labeling. Interestingly, averaging their outputs yields better results than using DAv2 alone, which performs better for transparent and mirror surfaces. As blending and inpainting remove the transparent or mirror surfaces from images, DA effectively complements DAv2 there. 
The final DAv2 model is fine-tuned on a carefully selected subset of datasets, including TransCG \cite{fang2022transcg}, ClearGrasp \cite{sajjan2020clear}, MIDepth \cite{liang2024delving}, Hammer \cite{jung2023importance}, HouseCat6D \cite{jung2024housecat6d}, MSD \cite{yang2019my}, Trans10K \cite{xie2020segmenting}, and Booster \cite{booster}. During training, extensive data augmentations are applied to enhance robustness to varying lighting conditions. At inference, test-time augmentations (flipping and color jittering) further refine the final depth predictions.

\begin{figure}[t]
\centering
\includegraphics[width=0.65\linewidth]{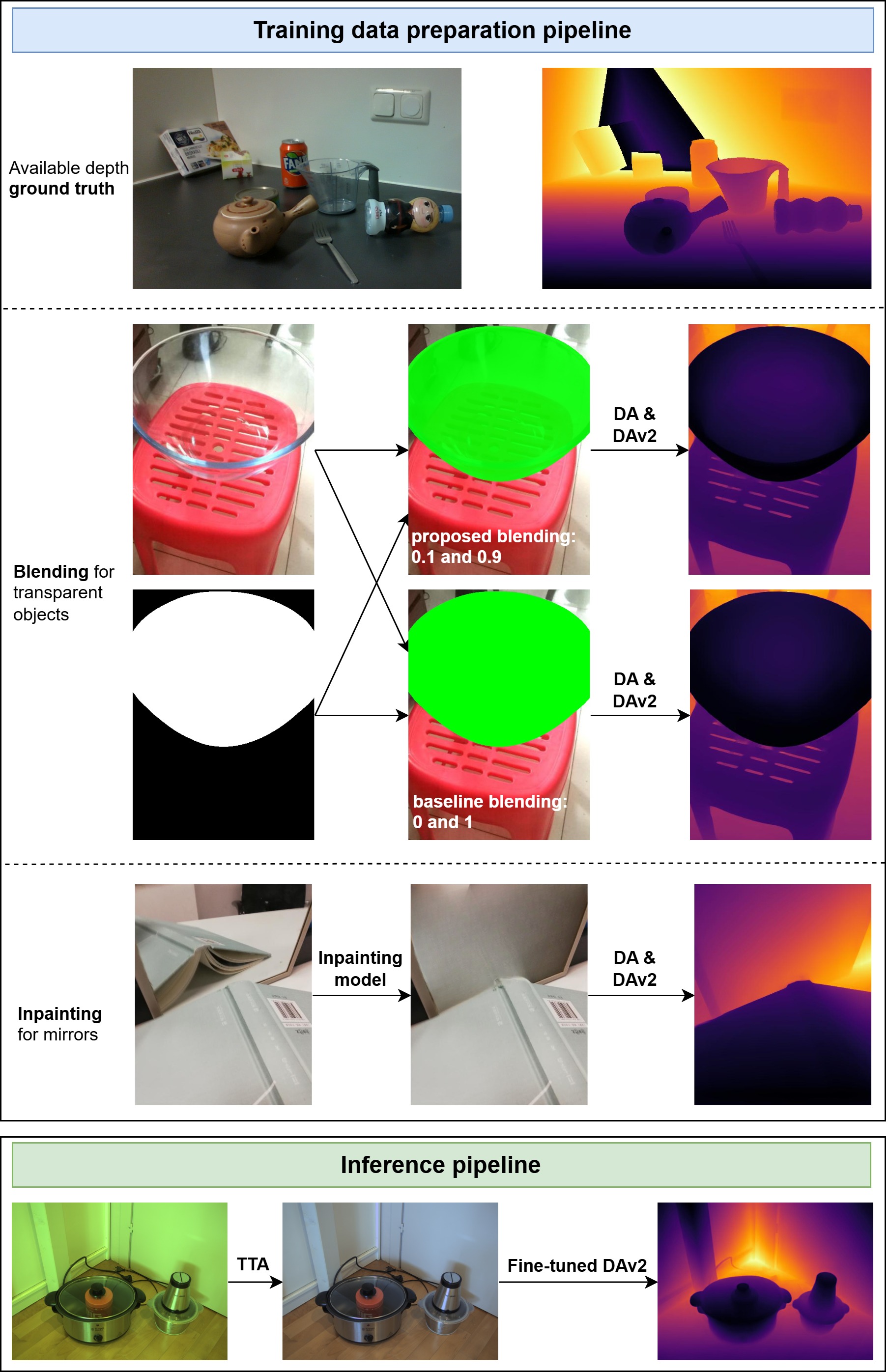}
\vspace{-0.2cm}
\caption{\textbf{Network Architecture -- Team \textit{Lavreniuk}.}}
\vspace{-0.5cm}
\label{fig:lavreniuk}
\end{figure}

\begin{figure}[t]
\centering
\includegraphics[width=0.65\linewidth]{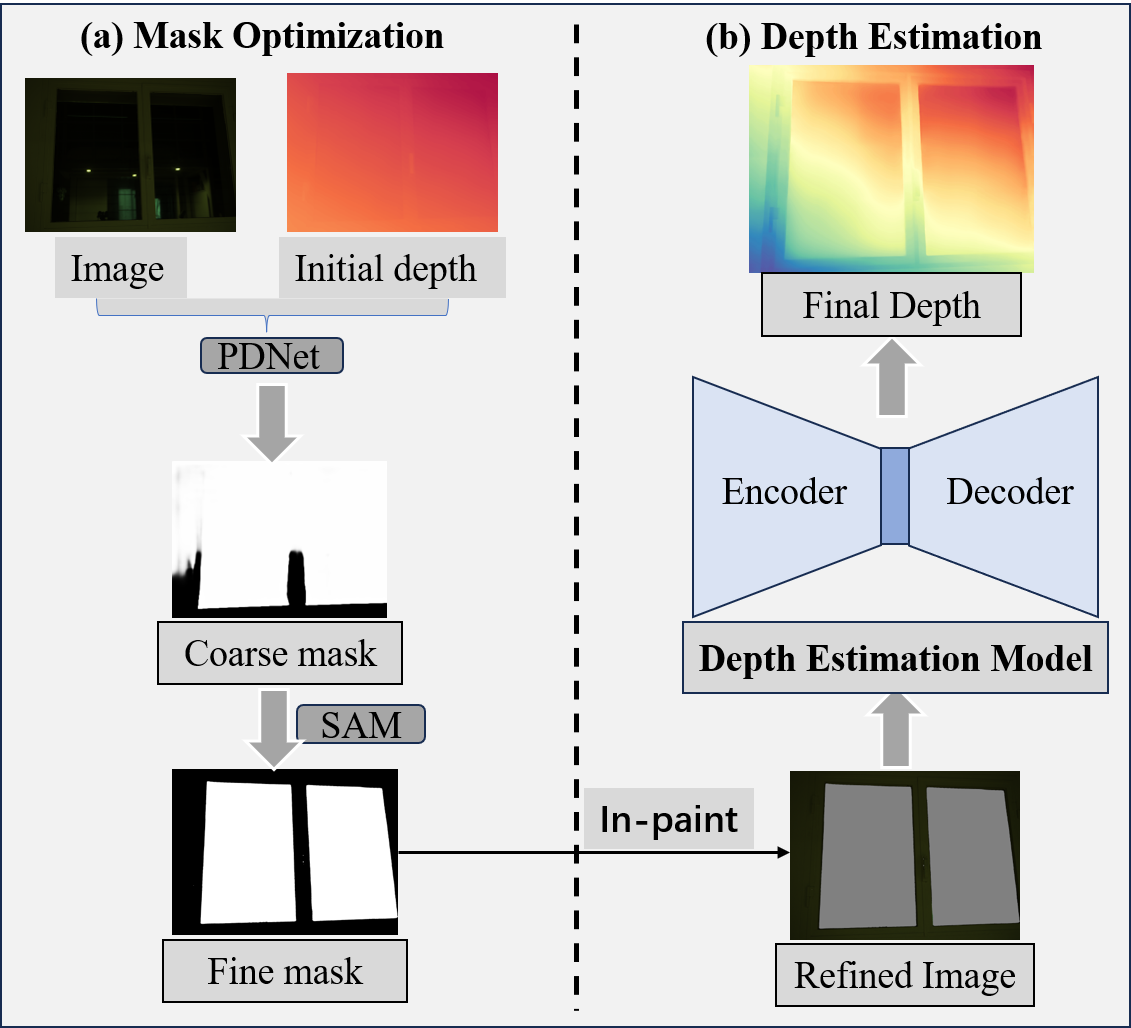}
\vspace{-0.3cm}
\caption{\textbf{Network Architecture -- Team \textit{colab}.}}
\vspace{-0.5cm}
\label{fig:colab}
\end{figure}

\subsubsection{Team 2 - colab}
The colab team (CodaLab: \textit{what}) presents ``DepthInpaint," an efficient depth estimation optimization framework for ToM objects combining two-stage ToM mask optimization and in-paint mechanism. Through this two-stage approach, it overcomes the physical limitations of traditional monocular depth perception and significantly improves the accuracy of depth estimation for ToM surfaces. 
The method first generated a rough ToM surfaces mask based on the PDNet model \cite{mei2021depth}, which combines RGB image and depth information for mirror segmentation. Then, DepthInpaint refines the rough mask with Segment Anything Model \cite{kirillov2023segment} (SAM) based on the image and initial mask. Following this, a physics-guided image inpaint strategy is applied to the masked regions, eliminating artifacts from specular highlights and medium refraction. Finally, Depth Pro \cite{bochkovskii2024depth} is used to generate depth metric from inpaint images. The key innovations of the approach are: 1) a second-stage mask optimization strategy for ToM surfaces, and 2) a mask-guided image-in-paint mechanism.
\vspace{-0.2cm}

\subsubsection{Team 3 - IPCV}
The IPCV team (CodaLab: \textit{JameerBabu}) employs Marigold \cite{ke2023repurposing} a monocular depth estimation model that leverages the visual knowledge embedded in diffusion-based image generators.
Specifically, Marigold is built upon the architecture of Stable Diffusion \cite{stable-diffusion-v1-5}, a latent diffusion model. The core component of this architecture is a denoising U-Net, which operates within the latent space of the model. To adapt Stable Diffusion for depth estimation, Marigold employs a fine-tuning protocol that focuses on this denoising U-Net while preserving the integrity of the latent space. This fine-tuning is performed using synthetic RGB-D datasets, such as Hypersim \cite{roberts2021hypersim} and Virtual KITTI \cite{cabon2020virtual}, and can be completed within a few days on a single GPU.
For implementation, the team directly used inference with the pretrained weights from Marigold repository.

\begin{figure}[t]
\centering
\includegraphics[width=0.75\linewidth]{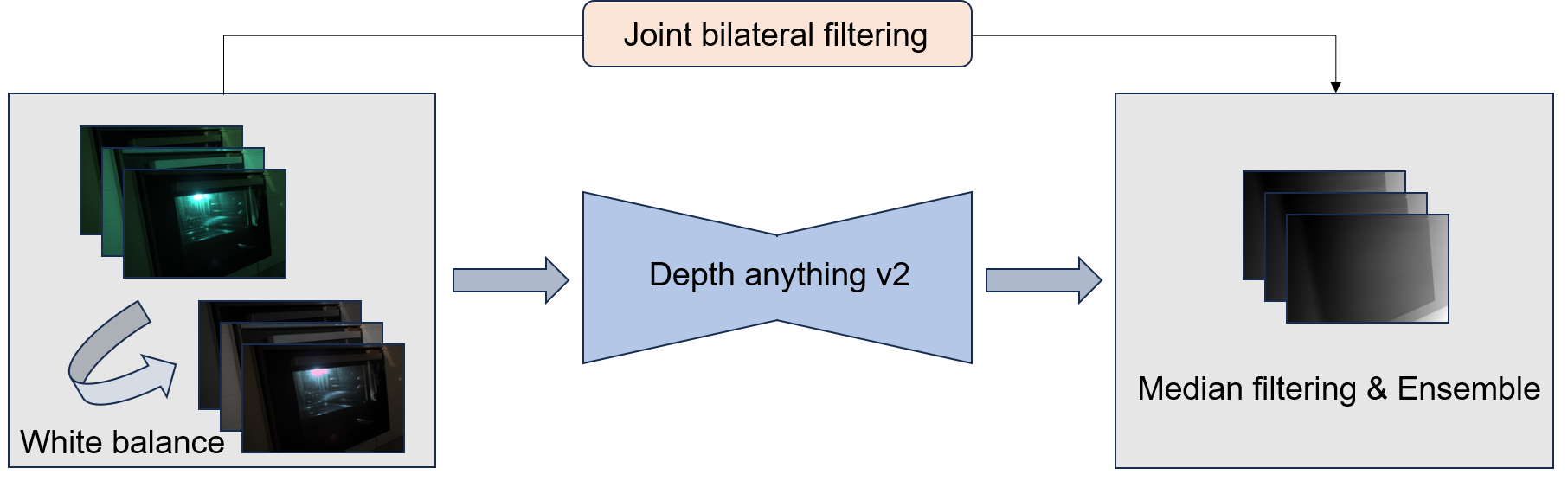}
\vspace{-0.3cm}
\caption{\textbf{Network Architecture -- Team \textit{PreRdw}.}}\vspace{-0.5cm}
\label{fig:prerdw}
\end{figure}
\subsubsection{Team 4 - PreRdw}
The PreRdw team (CodaLab: \textit{jingc}) presents ``Reflective Depth Wizard", a method that takes advantage of Depth Anything V2's inherent generalization capability. The team fine-tuned the model for accurately estimating depth on challenging reflective and transparent materials, initializing training using the pre-trained Hypersim model as their foundation.
To prepare the training data, they converted Booster depth annotations into usable depth maps and computed scale (s) and shift (t) parameters to properly align the model's predictions with the ground truth depths during training. All input images were resized to $518 \times 714$ pixels for processing. 
During inference, the team implemented several enhancements: (i) Color calibration: Applied the Gray World algorithm to normalize the input image color distributions; (ii) Depth refinement: Processed raw depth predictions with a bilateral filter to smooth surfaces while preserving edge details; (iii) Ensemble optimization: Combined results across varying lighting conditions and applied median filtering to reduce noise in final depth maps.

\section*{Acknowledgments}
This work was partially supported by the Humboldt Foundation. We thank the NTIRE 2025 sponsors: ByteDance, Meituan, Kuaishou, and University of Wurzburg (Computer Vision Lab).

\appendix
\section{NTIRE 2025 Organizers}

\noindent\textit{\textbf{Title}}:\\
NTIRE 2025 Challenge on HR Depth from Images of Specular and Transparent Surfaces\\
\textit{\textbf{Members}}:\\
Pierluigi Zama Ramirez$^1$ (pierluigi.zama@unibo.it), 
Alex Costanzino$^1$,
Fabio Tosi$^1$,
Matteo Poggi$^1$,
Samuele Salti$^1$,
Stefano Mattoccia$^1$,
Luigi Di Stefano$^1$,
Radu Timofte$^2$ \\
\textit{\textbf{Affiliations}}:\\
$^1$ University of Bologna, Italy\\
$^2$ Computer Vision Lab, University of W\"urzburg, Germany\\

\section{Track 1: Teams and Affiliations} \label{sec:teams_stereo}

\subsection*{NJUST-KMG}

\noindent\textit{\textbf{Members:}}\\
Zhe Zhang$^1$ (zhe.zhang@njust.edu.cn), Yang Yang (yyang@njust.edu.cn) $^1$\\
\textit{\textbf{Affiliations:}}\\
$^1$ Nanjing University of Science and Technology, China\\
\vspace{-0.3cm}

\subsection*{Robot01-vRobotit}

\noindent\textit{\textbf{Members:}}\\
Wu Chen$^1$ (chenw@bupt.edu.cn), Anlong Ming$^1$ (mal@bupt.edu.cn), Mingshuai Zhao$^1$ (mingshuai\_z@bupt.edu.cn), Mengying Yu$^1$ (yumengying@bupt.edu.cn), Shida Gao$^1$ (gaostar2024@bupt.edu.cn), Xiangfeng Wang$^1$ (xiangfeng\_w@foxmail.com), Feng Xue$^2$ (feng.xue@unitn.it)\\
\textit{\textbf{Affiliations:}}\\
$^1$ Beijing University of Posts and Telecommunications, China\\
$^2$ University of Trento, Italy\\
\vspace{-0.3cm}

\subsection*{Samsung R\&D Institute China-Beijing (SRC-B)}

\noindent\textit{\textbf{Members:}}\\
Jun Shi$^1$ (jun7.shi@samsung.com), Yong Yang$^1$, Yong A$^1$, Yixiang Jin$^1$, Dingzhe Li$^1$\\\
\textit{\textbf{Affiliations:}}\\
$^1$ Samsung R\&D Institute China-Beijing (SRC-B)\\
\vspace{-0.3cm}

\subsection*{weouibaguette}

\noindent\textit{\textbf{Members:}}\\
Aryan Shukla$^1$ (aryan.shukla.1@ens.etsmtl.ca), Liam Frija-Altarac$^1$ (liam.frija-altarac.1@ens.etsmtl.ca), Matthew Toews$^1$
 \\
\textit{\textbf{Affiliations:}}\\
$^1$ École de technologie supérieure (ÉTS), Montréal, Canada\\
\vspace{-0.3cm}

\section{Track 2: Teams and Affiliations} \label{sec:teams_mono}
\subsection*{colab}

\noindent\textit{\textbf{Members:}}\\
Hui Geng$^1$ (gengh666666@163.com), Tianjiao Wan$^1$, Zijian Gao$^1$, Qisheng Xu$^1$, Kele Xu$^1$, Zijian Zang$^2$\\
\textit{\textbf{Affiliations:}}\\
$^1$ National University of Defense Technology, Changsha, China\\
$^2$ Fudan University\\
\vspace{-0.3cm}

\subsection*{IPCV}

\noindent\textit{\textbf{Members:}}\\
Jameer Babu Pinjari$^1$ (jameer.jb@gmail.com), Kuldeep Purohit$^1$ (kuldeeppurohit3@gmail.com)\\
\textit{\textbf{Affiliations:}}\\
$^1$ Independent Researchers\\
\vspace{-0.3cm}

\subsection*{Lavreniuk}

\noindent\textit{\textbf{Members:}}\\
Mykola Lavreniuk$^1$ (nick\_93@ukr.net)\\
\textit{\textbf{Affiliations:}}\\
$^1$ Space Research Institute NASU-SSAU, Kyiv, Ukraine\\
\vspace{-0.3cm}

\subsection*{PreRdw}

\noindent\textit{\textbf{Members:}}\\
Jing Cao$^1$ (caojing@stu.hit.edu.cn), Shenyi Li$^1$ (504774430@qq.com), Kui Jiang$^1$ (jiangkui@hit.edu.cn), Junjun Jiang$^1$ (jiangjunjun@hit.edu.cn), Yong Huang$^1$ (huangyong@hit.edu.cn)\\
\textit{\textbf{Affiliations:}}\\
$^1$ Harbin Institute of Technology, China\\

{\small
\bibliographystyle{ieee_fullname}
\bibliography{egbib,depth}
}
 
\end{document}